\documentclass[a4paper,11pt]{article}

\usepackage{IJUQ_31935_Macros}

\usepackage{geometry}										% [showframe]
\usepackage{hyperref}
\hypersetup{
	pdftitle = {Machine learning applied to simulations of collisions between 
	rotating, differentiated planets},
	pdfauthor = {Miles L. Timpe, Maria Han Veiga, Mischa Knabenhans, Joachim 
	Stadel, Stefano Marelli},
	pdfsubject = {Manuscript submitted to XXX},
	pdfkeywords = {emulation, giant impacts, Gaussian processes, machine 
	learning, neural networks, planet formation, polynomial chaos expansion, 
	XGBoost},
	colorlinks = false,										% false: boxed 
}

\usepackage{bm}
\usepackage{subfigure}
\usepackage{longtable}

\newcommand{\eg}{\textit{e.g.}}
\newcommand{\ie}{\textit{i.e.}}

\usepackage[english]{babel}									% language
\usepackage[utf8]{inputenc}									% input text 
\usepackage[T1]{fontenc}									% output encoding

\usepackage{lmodern}										% Latin Modern

\usepackage{amsmath}										% improved 
\usepackage{amstext}										% text in 
\usepackage{amsfonts}										% extended set of 
\usepackage{amssymb}										% extended symbol 
\usepackage{bbm}
\usepackage{comment}
\usepackage{graphicx}
\usepackage{tabularx}
\usepackage{multirow}
\usepackage{multicol}
\usepackage{caption}
 
\newcommand{\matlab}{\textsc{Matlab}}

\def \bfal {\bm{\alpha}}

\def \bfC {\textbf{C}}
\def \bfK {\textbf{K}}

\def \bfX {\textbf{X}}

\def \bfZ {\textbf{Z}}

\def \bfc {\textbf{c}}

\def \bfv {\textbf{v}}
\def \bfw {\textbf{w}}
\def \bfwh {\widehat{\textbf{w}}}
\def \bfx {\textbf{x}}
\def \bfy {\cy}
\def \bfz {\textbf{z}}
\def \bfth {\bm{\theta}}
\def \bfthh {\widehat{\bm{\theta}}}

\newcommand{\vY}{\ensuremath{\ve{y}}}

\newcommand{\norm}[1]{\left\lVert#1\right\rVert}

\renewcommand{\eqref}[1]{Eq.~(\ref{#1})}
\newcommand{\secref}[1]{Section~\ref{#1}}

\usepackage[numbers,sort&compress]{natbib}							% number 
\begin{document}

\title{Extending classical surrogate modelling to high dimensions through 
supervised dimensionality reduction: a data-driven approach} 
%For at least  authors with different addresses, use instead the following 
%commands
\author{C. Lataniotis, S. Marelli, B. Sudret}

\maketitle

\begin{abstract}
	Thanks to their versatility, ease of deployment and high-performance, 
	surrogate models have become staple tools in the arsenal of uncertainty 
	quantification (UQ). 
	From local interpolants to global spectral decompositions, surrogates are 
	characterised by their ability to efficiently emulate complex computational 
	models based on a small set of model runs used for training.
	An inherent limitation of many surrogate models is their susceptibility to 
	the curse of dimensionality, which traditionally limits their applicability 
	to a maximum of $\co(10^2)$ input dimensions. 
	We present a novel approach at high-dimensional surrogate modelling that is 
	model-, dimensionality reduction- and surrogate model- agnostic (black 
	box), and can enable the solution of high dimensional (\ie{} up to 
	$\co(10^4)$) problems. 
	After introducing the general algorithm, we demonstrate its performance by 
	combining Kriging and polynomial chaos expansions surrogates and kernel 
	principal component analysis. 
	In particular, we compare the generalisation performance that the resulting 
	surrogates achieve to the classical sequential application of 
	dimensionality reduction followed by surrogate modelling on several 
	benchmark applications, comprising an analytical function and two 
	engineering applications of increasing dimensionality and complexity.
\end{abstract}

% #######################	
\section{Introduction}
% #######################
\label{sec:Introduction}

It is a common practice to study the behaviour of physical and engineering 
systems through computer simulation. 
In a real-world setting, such systems  are driven by input parameters, the 
values of which can be uncertain or even unknown.
Uncertainty quantification (UQ) aims at identifying and quantifying the sources 
of uncertainty in the input parameters to assess the uncertainty they cause in 
the model predictions. 
In the context of Monte Carlo simulation, such workflow typically entails the 
repeated evaluation of the computational model. 
However, it may become intractable when a single simulation is computationally 
demanding, as is often the case with modern computer codes. 
A remedy to this problem is to substitute the model with a surrogate that 
accurately mimics the model response within the chosen parameter bounds, but is 
computationally inexpensive.
An additional benefit of surrogate models is that they are often non-intrusive, 
i.e. their construction only depends on a training set of model evaluations, 
without access to the model itself. 
This includes the case when the model is not available, but only a pre-existent 
data set is, as is typical in machine learning applications. 
The latter setting is the focus of this paper.
Popular surrogate modelling techniques (SM) include Gaussian process modelling 
and regression \citep{Sacks1989,Rasmussen2006}, polynomial chaos expansions 
\citep{Ghanembook1991,Xiu2002,XiuBook2010}, low-rank tensor approximations 
\citep{Chevreuil2015,KonakliJCP2016}, and support vector regression 
\citep{Vapnik1995}.  
Parametrising and training a surrogate model, however, can become harder or 
even intractable as the number of input parameters increases, a well known 
problem often referred to as \emph{curse of dimensionality} (see \eg{} 
\citet{Verleysen05}). 
Similar challenges arise in the presence of high-dimensional model responses 
(see \eg{} \citet{Gu2016}), but this is beyond the scope of this paper. 

For the sake of clarity, in the following we will classify high-dimensional 
inputs in two broad categories, depending on their characteristics: 
\textit{unstructured} or \textit{structured}. 
Unstructured inputs are characterised by the lack of an intrinsic ordering, and 
they are commonly identified with the so-called ``model parameters'', \eg{} 
point loads on mechanical models, or resistance values in electrical circuit 
models.
Structured inputs (also known as functional data, see \eg{} 
\citet{Ramsay2004}), on the other hand, are characterised by the existence of a 
natural ordering and/or a distance function (\textit{i.e.} they show strong 
correlation across some physically meaningful set of coordinates), as it is 
typical for time-series or space-variant quantities represented by maps.
Boundary conditions in complex simulations that rely on discretisation grids, 
\textit{e.g.} time-dependent excitations at grid nodes, often belong to this 
second class. 
In most practical applications, unstructured inputs range in dimension in the 
order $\co(10^{0-2})$, while structured inputs tend to be in the order 
$\co(10^{2-6})$.

Several strategies have been explored in the literature to deal with high 
dimensional problems for surrogate modelling. 
A common approach in dealing with unstructured inputs is input variable 
selection, which consists in identifying the  ``most important'' inputs 
according to some importance measure, see \textit{e.g.} \citet{Saltelli2008}, 
\citet{Iooss2015}, and simply ignoring the others (\textit{e.g.} by setting 
them to their nominal value).

In the context of kernel-based emulators (\textit{e.g.} Gaussian process 
modelling or support vector machines), some attention has been devoted to the 
use of simple isotropic kernels \citep{djolonga2013}, or to the design of 
specific kernels for high-dimensional input vectors, sometimes including 
deep-learning techniques (\textit{e.g.}, 
\cite{Lawrence2005,durrande2012,wilson2016}).

In more complex scenarios, the more general concept of \textit{dimensionality 
reduction} (DR) is applied, which essentially consists in mapping the input 
space to a suitable lower dimensional space using an appropriate transformation 
prior to the surrogate modelling stage.
The latter approach is considered in this work due to its applicability to 
cases for which variable selection seems inadequate or insufficient (\eg{} in 
the presence of structured inputs).  

%\orangebf{Review existing methods for dealing with this problem}

In the current literature, a two-step approach is often followed for dealing 
with such problems: first, the input dimension is reduced; then, the surrogate 
model is constructed directly in the reduced (feature-) space. The 
dimensionality reduction step is based on an \emph{unsupervised} objective, 
\ie{} an objective that only takes into account the input observations. 
Examples of unsupervised objectives include the minimisation of the input 
reconstruction error \citep{vincent2008}, maximisation of the sample variance 
\citep{Pearson01}, maximisation of statistical independence 
\citep{hyvarinen1997one}, and preservation of the distances between the 
observations \citep{tenenbaum2000isomap,roweis2000nonlinear,hinton2003sne}. 
While in principle attractive due to their straightforward implementation, 
unsupervised approaches for dimensionality reduction may be suboptimal in this 
context, because the input-output map of the reduced representation may exhibit 
a complex topology unsuitable for surrogate modelling 
\citep{wahlstrom2015,calandra2016}.

To deal with this issue, various \emph{supervised} techniques have been 
proposed, in the sense that the objective of the input compression takes into 
account the model outputs.
One such approach that has received attention recently is based on the 
so-called \emph{active subspaces} concept \citep{Constantine2014}.
Various methods that belong to this category, provide a linear transformation 
of the high dimensional input space into a reduced space that is characterised 
by maximal variability w.r.t. the model output. 
However, active subspace methods often require the availability of the model 
gradient w.r.t. the input parameters, a limiting factor in data-driven 
scenarios where such information is not available and needs to be approximated 
\citep{Fornasier2012}. 
Moreover, the numerical computation of the gradient may be infeasible in 
problems that involve structured inputs such as time series or 2D maps with 
$\co(10^{2-6})$ components.

Other data-driven supervised DR techniques have been proposed in the 
literature, that are dependent on the properties of a specific combination of 
either DR or SM techniques.
In the context of polynomial chaos expansions, \citet{Tipireddy2014} propose a 
basis adaptation scheme based on an optimal transformation of the probabilistic 
input space to a suitable Gaussian space.
Further improvements have been proposed in \citet{Tsilifis2019}, where the 
efficiency of the method is improved by combining the basis adaptation with 
compressive sensing schemes. \citet{Papaioannou2019} propose instead a 
gradient-free algorithm based on partial least-squares that is applicable in 
settings with black-box numerical models. 
\citet{HintonSalakhutdinov2006b} employ multi-layer neural networks for both 
the DR and the SM steps. Specifically, an unsupervised objective based on the 
reconstruction error is followed by a generalisation performance objective that 
aims at fine tuning the network weights with respect to a measure of the 
surrogate modelling error. 
Similar approaches have been proposed with other combinations of methods. 
In \citet{damianou2013deep},  the same idea is extended by using stacked 
Gaussian processes instead of multilayer neural networks. 
In \citet{huang2015,calandra2016} this approach is extended by combining neural 
networks with Gaussian processes within a Bayesian framework.

All of these methods demonstrate that supervised methods yield a significant 
accuracy advantage over the unsupervised ones, as the final goal of the 
supervised learner (\textit{i.e.} surrogate model accuracy) matches the final 
goal of high-dimensional surrogate modelling in the first place. 
However, this increased accuracy comes at the cost of restricting the 
applicability of such methods to specific combinations of DR and SM techniques.

In this paper, we propose a novel method of performing dimensionality reduction 
for surrogate modelling in a data-driven setting, which we name  DRSM. 
The aim of this method is to capitalise on the performance gains of supervised 
DR, while maintaining maximum flexibility in terms of both DR and SM 
methodologies. 
Recognising that different communities, applications and researchers have in 
general access to one or two preferred techniques for either DR or SM, the 
proposed approach is fully non-intrusive, \textit{i.e.} both the DR and the SM 
stages are considered as \textit{black boxes} under very general conditions.
The novelty lies in the way the two stages are coupled into a single problem, 
for which dedicated solvers are proposed.

This paper is structured as follows: Section 2 introduces the main ingredients 
required by DRSM, namely dimensionality reduction and surrogate modelling. For 
the sake of clarity, some of the techniques that will be specifically used in 
the applications section are also introduced, \textit{i.e.} kernel principal 
component analysis (KPCA) for DR, Gaussian process modelling, a.k.a. Kriging, 
and polynomial chaos expansions (PCE) for SM. The core framework underlying 
DRSM is then introduced. 
Finally, the effectiveness of DRSM is analysed on several benchmark 
applications including both unstructured and structured inputs, ranging from 
low-dimensional analytical functions to a complex engineering 2-dimensional 
heat-transfer problem.

% #######################	
\section{Ingredients for surrogate modelling in high dimension} 
\label{sec:Methodology}
% #######################
As the name implies, DRSM consists in the combination of two families of 
computational tools: dimensionality reduction and surrogate modelling. This 
section aims at highlighting the main features of each, and how they can be 
exploited without resorting to intrusive, dedicated algorithms.

\subsection{Dimensionality reduction} \label{sec:Meth:DR}

Consider a set of high-dimensional samples $\cx = \acc{\bfx^{(i)}\in \Rr^M\, , 
\, i=1\enu N}$. 
In an abstract sense, dimensionality reduction (DR) refers to the parametric 
mapping $g: \cx \in \Rr^M \mapsto \cz \in \Rr^m$ of the form:

\begin{equation} \label{eq:DR_general_form}
\bfz = g(\bfx ; \bfw)
\end{equation}
where $\bfz \in \cz$, $\bfx \in \cx$, and $\bfw$ is the set of parameters 
associated with the mapping. 
Dimensionality reduction occurs if $m\ll M$, \ie{} if $m=\co\prt{10^{0-1}}$ 
whereas $M=\co\prt{10^{2-4}}$.
The nature and number of the parameters $\bfw$ depends on the specific DR 
method under consideration. 

Such transformations are motivated by the assumption that the samples in $\cx$ 
lie on some manifold with dimensionality $m$ that is embedded within the 
$M$-dimensional space. 
This specific value of $m$ is in some applications referred to as the 
``intrinsic dimension'' of $\cx$ \citep{Fukunaga2013}. 
From an information theory perspective, the intrinsic dimension refers to the 
minimum number of scalars that is required to represent $\cx$ without any loss 
w.r.t. an appropriate information measure. 
In practice it is a-priori unknown. 
In such cases DR is an ill-posed problem that can only be solved by assuming 
certain properties of $\cx$, such as its intrinsic dimension. 
Alternatively the latter may be approximated and/or inferred from the available 
data by various approaches (see \eg{} \citet{Camastra2003} for a comparative 
overview).

An important aspect of all parametric DR methods, regardless of their 
specificity, is that for each choice of dimension $m$ the remaining parameters 
$\bfw$ are estimated by minimising a suitable error measure (sometimes referred 
to as loss function):
\begin{equation}
\widehat{\bfw} =  \arg \underset{\cd_\bfw}{\min} \, J(\bfw;\cx), 
\end{equation}
where $ \widehat{\bfw}$ denotes the estimated parameters, $\cd_\bfw$ the 
feasible domain of $\bfw$, $J(\cdot)$ the error measure and $\cx$ the available 
data. 
The choice of the error measure depends on the specific application DR is used 
for. 
When the goal is direct compression of a high dimensional input without 
information loss (a common situation in telecommunication-related 
applications), a typical choice of $J(\cdot)$ is the so-called mean-squared 
reconstruction error, that reads:
\begin{equation}
J(\bfw;\cx) =  \frac{1}{N} \sum_{i=1}^{N} \norm{\bfx^{(i)} - 
\tilde{\bfx}^{(i)}}^2,
\end{equation}
where $\tilde{\bfx}=g^{-1}(\bfz,\bfw)$ denotes the reconstruction of the sample 
$\bfx$,  calculated through the inverse transform  $g^{-1}: \cz \in \Rr^m 
\mapsto \cx \in \Rr^M$. 
In the general case, additional parameters may be introduced in $g^{-1}$, or 
the inverse transform may not exist at all (see \eg{}  \citet{Kwok2003}).

For a detailed description of the specific DR methods used in this paper to 
showcase the proposed methodology, namely principal component analysis (PCA) 
and kernel PCA, the reader is referred to \secref{sec:Meth:Selected DR and SM}.

\subsection{Surrogate Modelling} \label{sec:Meth:SM:Intro}

In the context of UQ, the physical or computational model of a system can be 
seen as a black-box that performs the mapping:
\begin{equation} \label{eq:true_model}
\ve{Y} = \cm(\ve{X}),
\end{equation}
where $\ve{X}$ is a random vector that parametrises the variability of the 
input parameters (\textit{e.g.} through a joint probability density function) 
and $\ve{Y}$ is the corresponding random vector of model responses. 
One of the main applications of UQ is to propagate the uncertainties from 
$\ve{X}$ to $\ve{Y}$ through the model $\cm$. 
Direct methods based on Monte-Carlo simulation may require that the 
computational model is run several thousands of times for different 
realisations $\bfx$ of the input random vector $\ve{X}$. 
However, most models that are used in applied sciences and engineering  
(\textit{e.g.} high-resolution finite element models) can have high 
computational costs per model run. As a consequence, they cannot be used 
directly. 
To alleviate the associated computational burden, surrogate models have become 
a staple tool in all types of uncertainty quantification applications.

A surrogate model $\widehat{\cm}$ is a computationally inexpensive 
approximation of the true model of the form: 
\begin{equation}\label{eq:surrogate}
\cm(\ve{X}) = \widehat{\cm}(\ve{X}; \bm{\theta})  + \epsilon,
\end{equation}
where $\bm{\theta}$ is a set of parameters that characterise the surrogate 
model and $\epsilon$ refers to an error term. The parameters $\bm{\theta}$ are 
inferred (typically through some form of optimisation process) from a limited 
set of runs of the original model ${\cx = \acc{\bfx^{(1)} \enu \bfx^{(N)}}}$, 
called the \textit{experimental design}. 
As an example, $\bfth$ denotes the set of coefficients in the case of a 
truncated polynomial chaos expansion, or the set of parameters of both the 
trend and the covariance kernel in case of Gaussian process modelling. 
Throughout the rest of the paper, the output of the model $\cm$ is considered 
scalar, \ie{} $y = \cm(\bfx) \in \Rr$. 

Arguably the most well-known accuracy measure for most surrogates is the 
relative generalisation error $\varepsilon_{gen}$ that reads:
\begin{equation} \label{eq:epsilong_gen_ideal}
\varepsilon_{gen} = \Esp{\prt{Y - \widehat{\cm}(\ve{X};\bm{\theta})}^2}/\Var{Y}.
\end{equation}
This error measure (or, more precisely, one of its estimators) is also the 
ideal objective function for the optimisation process involved in the 
calibration of the surrogate parameters $\bm{\theta}$.  
In practical situations, however, it is not possible to calculate 
$\varepsilon_{gen}$ analytically. An estimator $\widehat{\varepsilon}_{gen}$ of 
this error can be computed by comparing the true and surrogate model responses 
evaluated at a sufficiently large \textit{validation set} $\cx_v = 
\acc{\bfx^{(1)} \enu \bfx^{(N_v)} }$ of size $N_v$:
\begin{equation} \label{eq:epsilong_gen_estim}
\widehat{\varepsilon}_{gen} = \frac{\sum_{i=1}^{N_v} \prt{ \cm(\bfx^{(i)}) - 
\widehat{\cm}(\bfx^{(i)}) }^2} 
{\sum_{i=1}^{N_v} \prt{\cm(\bfx^{(i)}) - \widehat{\mu}_y}^2}, 
\end{equation}
where $\widehat{\mu}_y=\frac{1}{N} \sum_{i=1}^{N_v} \cm(\bfx^{(i)})$ is the 
sample mean of the validation set responses and $\widehat{\cm}(\bfx^{(i)})$ is 
used in place of $\widehat{\cm}(\bfx^{(i)};\bm{\theta})$ to simplify the 
notation. 

In data-driven applications, or when the computational model is expensive to 
evaluate,  only a single set $\cs \overset{\text{def}}{=}  \acc{\cx, \bfy}$ is  
available. 
The entire set is therefore used for calculating the surrogate parameters. 
Estimating the generalisation error by means of \eqref{eq:epsilong_gen_estim} 
on the same set, however, corresponds to computing the so-called 
\emph{empirical error}, which is prone to underestimate drastically the true 
generalisation error, due to the overfitting phenomenon.
In such cases, a fair approximation of $\widehat{\varepsilon}_{gen}$ can be 
obtained by means of cross-validation (CV) techniques (see \eg{} 
\citet{Hastie2001}). 
In $k$-fold CV, $\cs$ is randomly partitioned into $k$ mutually exclusive and 
collectively exhaustive sets $\cs_i$ of approximately equal size: 

\begin{equation} \label{eq:CV_sets}
\cs_i \cap  \cs_j = \emptyset ~, ~ \forall (i,j) \in \lbrace 1,\ldots,k \rbrace 
^2 ~ \text{ and } \bigcup_{i=1}^{k} \cs_i = \cs .
\end{equation} 

The $k$-fold cross-validation error $\varepsilon_{CV}$ reads:
\begin{equation} \label{eq:epsilon_cv}
\varepsilon_{CV} = \frac{\sum_{i=1}^{k} \sum_{\bfx \in \cs_i}\prt{ \cm(\bfx) - 
\widehat{\cm}^{\cs\setminus\cs_i}( \bfx) }^2} 
{\sum_{\bfx \in \cs} \prt{\cm(\bfx) - \widehat{\mu}_y}^2},
\end{equation}
where $\widehat{\cm}^{\cs\setminus\cs_i}$ denotes the surrogate model that is 
calculated using $\cs$ excluding $\cs_i$. The bias of the generalisation error 
estimator is expected to be minimal in the extreme case of \emph{leave-one-out 
(LOO) cross-validation} \citep{Arlot2010}, which corresponds to $N-$fold cross 
validation. The LOO error $\varepsilon_{LOO}$ is calculated as in 
\eqref{eq:epsilon_cv} after substituting the set $S_i$ by the singleton 
$\acc{\bfx^{(i)}}$ (\ie{} $k=N$): 
\begin{equation} \label{eq:epsilon_LOO}
\varepsilon_{LOO} = \frac{\sum_{i=1}^{N} \prt{ \cm(\bfx^{(i)}) - 
\widehat{\cm}^{\backslash i}( \bfx^{(i)}) }^2} 
{\sum_{i=1}^{N} \prt{\cm(\bfx^{(i)}) - \widehat{\mu}_y}^2},
\end{equation}
where the term $\cm^{\backslash i}( \bfx^{(i)})$, denotes the surrogate built 
from the set $S \backslash \acc{\bfx^{(i)}}$, evaluated at $\bfx^{(i)}$. 
The calculation of $\varepsilon_{LOO}$ can be computationally expensive, 
because it requires the evaluation of $N$ surrogates, but it does not require 
any additional run of the full  computational model. 
For Gaussian process modelling and polynomial chaos expansions, computational 
shortcuts are available to alleviate such costs (\textit{e.g.} 
\citet{Dubrule1983,BlatmanJCP2011}), in the sense that  $\varepsilon_{LOO}$ in 
\eqref{eq:epsilon_LOO} is evaluated from a single surrogate model 
$\widehat{\cm}$ calculated from the full data set $\cs$.

As a final step in the surrogate modelling procedure, the set of parameters 
$\bm{\theta}$ of the surrogate model are optimised \textit{w.r.t.}  one of the 
generalisation error measures in \eqref{eq:epsilon_cv} or 
\eqref{eq:epsilon_LOO} directly, based on the available samples in $\cs$, \ie{}:
\begin{equation}
\label{eq:theta_optim_general}
\widehat{\bfth{}} = \arg \underset{\cd_{\bfth}}{\min}\,  
\widehat{\epsilon}_{gen}(\bfth; \cs),
\end{equation}
where $\widehat{\bfth{}} $ denotes the optimal set of parameters, $\cd_{\bfth}$ 
the feasible domain of parameters and $\widehat{\epsilon}_{gen}$ refers to the 
chosen estimator of $\epsilon_{gen}$. 
An important aspect of this optimisation step for many types of recent 
surrogates is that the number of parameters $\bm{\theta}$ scales with the 
number of input variables. Therefore, surrogates tend to suffer from the curse 
of dimensionality in two distinct ways: higher dimensional optimisation and 
underdetermination. Higher dimensional optimisation is linked to a complex 
objective-function topology, and is therefore prone to convergence to 
low-performing local minima. In general it requires global optimisation 
algorithms, such as genetic algorithms, covariance matrix adaptation, or 
differential evolution \citep{Goldberg1989,Hansen2003,Yang2007}. 
Underdetermination leads the solutions to the minimisation problem to be 
non-unique due to the lack of constraining data. In other words, surrogate 
models with more parameters require in general a larger experimental design or 
sparse minimisation techniques to avoid overfitting.

%%%%%%%%%%%%%%%%%%%%%%%%%%%%%%%%%%%%%%%%%%%%%%%%%%%%%%%%%
%                      DRSM                             %
%%%%%%%%%%%%%%%%%%%%%%%%%%%%%%%%%%%%%%%%%%%%%%%%%%%%%%%%%

\section{The proposed DRSM approach} \label{sec:Meth:DRSM}

\subsection{Introduction} \label{sec:Meth:DRSM:intro}

Consider now the experimental design $\cs = \acc{\cx, \bfy}$ introduced above, 
and assume that it is the only available information about the problem under 
investigation. Moreover, the dimensionality of the input space is high, \ie{} 
$\bfx^{(i)} \in \Rr^M\, ,\, i=1\enu N$ where $M$ is large, say 
$\co\prt{10^{2-4}}$. The goal is to calculate a surrogate model that serves as 
an approximation of the real model solely based on the available samples. This 
is a key ingredient for subsequent analyses in the context of uncertainty 
quantification.

To distinguish between various computational schemes, we denote from now on by 
$\widehat{\cm}|\cx,\bfy$ a surrogate model whose parameters $\bfth$ are 
calculated from the experimental design $\cx$ and associated model response 
$\bfy$. 
Due to the high input dimensionality, a surrogate $\widehat{\cm}|\cx,\bfy$ may 
lead to poor generalisation performance or it may not even be computationally 
tractable. 
To reduce the dimensionality, the class of DR methods was introduced in 
\secref{sec:Meth:DR}. 
A DR transformation, expressed by $\cz = g(\cx; \bfw)$, can provide a 
compressed experimental design, \ie{} $\bfz^{(i)} \in \Rr^m\, ,\, i=1\enu N$ 
with $m\ll M$. The surrogate $\widehat{\cm}|\cz,\bfy$ becomes tractable if $m$ 
is sufficiently small. 
The potential of $\widehat{\cm}|\cz,\bfy$ to achieve satisfactory 
generalisation performance depends on (i) the learning capacity of the 
surrogate itself and (ii) the assumption that the input-output map $\bfx 
\mapsto y$ can be sufficiently well approximated by a smaller set of features 
via the transformation $g(\cdot)$. 
This discussion focuses on the latter and assumes that the learning capacity of 
the surrogate is adequate. 
In case of unstructured inputs, the importance of each input variable may vary 
depending on the output of interest. 
In case of structured inputs, there is typically high correlation between the 
input components. 
Hence, in both families of problems a low-dimensional representation may often 
approximate well the input-output map.

Traditional DR approaches are focused on the discovery of the input manifold 
and not the input-output manifold. Performing an input compression without 
taking into account the associated output values may lead to a highly complex 
input-output map that is difficult to surrogate. 
In the DRSM (dimensionality reduction for surrogate modelling) approach 
proposed in this paper, we capitalise on this claim to try and find an optimal 
input compression scheme w.r.t. the generalisation performance of  
$\widehat{\cm}|\cz,\bfy$.

\subsection{A nested optimisation problem} \label{sec:Meth:DRSM:1_nested_optim}
The goal of DRSM is to optimise the parameters $\bfw$ of the compression scheme 
so that the auxiliary variables $\bfz = g(\bfx;\bfw)$ are suitable to achieve 
an overall accurate surrogate. The general formulation of this problem reads: 
\begin{equation}
\label{eq:DRSM_general}
\acc{\bfwh,\bfthh} = \underset{\bfw \in \cd_{\bfw},\,\bfth \in \cd_{\bfth} 
}{\arg \min} \ell \prt{\cm(\cdot),\widehat{\cm} \prt{g(\cdot;\bfw), \bfth} },
\end{equation}
where $\ell$ denotes the objective function (a.k.a. loss function) that 
quantifies the generalisation performance of the surrogate. 
In practice, if a validation set is available, $\ell$ corresponds to a 
generalisation error estimator like the one in \eqref{eq:epsilong_gen_estim}. 
In the absence of a validation set, then either the LOO estimator in 
\eqref{eq:epsilon_LOO} or its  $k$-fold CV counterpart in \eqref{eq:epsilon_cv} 
are used instead. 
In the following, it is assumed that a validation set is not available and the 
generalisation error is estimated by the LOO error, hence $\ell$ is substituted 
by the  $\varepsilon_{LOO}$ expression in \eqref{eq:epsilon_LOO}.

The proposed approach for solving \eqref{eq:DRSM_general}, is related to the 
concept of \textit{block-coordinate descent} \citep{Bertsekas1999}. 
During optimisation, the parameters $\bfw$ and $\bfth$ are updated in an 
alternating fashion. 
One of the main reasons for this choice is that the optimisation steps of both 
DR and SM techniques are often tuned ad-hoc to optimise their performance. 
Examples include sparse linear regression for polynomial chaos expansions 
\citep{BlatmanJCP2011}, or quadratic programming for support vector machines 
for regression \citep{Vapnik1995}. A single joint optimisation, albeit 
potentially yielding accurate results, would require the definition of complex 
constraints on the different sets of parameters $\bfw$ and $\bm{\theta}$. 
Therefore, the problem in \eqref{eq:DRSM_general} is expressed as a 
nested-optimisation problem. 
The outer loop optimisation reads:
\begin{equation}
\label{eq:DRSM_outer_loop}
\bfwh = \underset{\bfw \in \cd_{\bfw}}{\arg \min} ~ \varepsilon_{LOO}(\bfw; 
\bfthh(\bfw),\cx,\bfy),
\end{equation}
where $\varepsilon_{LOO}$ denotes the LOO error (\eqref{eq:epsilon_LOO}) of the 
surrogate $\widehat{\cm}(\bfz;\bfw,\cx,\bfy)$ evaluated at $\acc{\cx,\bfy}$ and 
$\bfthh(\bfw)$ denotes the optimal parameters of $\widehat{\cm}$ for that 
particular $\bfw$ value. 
The term $\bfthh(\bfw)$ is calculated by solving the inner loop optimisation 
problem: 

\begin{equation}
\label{eq:DRSM_inner_loop}
\bfthh = \underset{\bfth \in \cd_{\bfth} }{\arg \min} ~ 
\varepsilon_{LOO}(\bfth;\bfw, \cx,\bfy).
\end{equation}

The nested optimisation approach to DRSM comes with costs and benefits.
On the one hand, each objective function evaluation of the outer-loop 
optimisation becomes increasingly costly w.r.t. the number of samples in the 
experimental design and the complexity of the surrogate model. 
On the other hand, the search space in each optimisation step can be 
significantly smaller, compared to the joint approach, due to the reduced 
number of optimisation variables. 
Moreover, this nested optimisation approach enables DRSM to be entirely 
non-intrusive. 
Off-the-shelf well-known surrogate modelling methods can be used to solve 
\eqref{eq:DRSM_inner_loop}. 
% For example in case of polynomial chaos expansions the inner loop 
%optimisation may correspond to ordinary least-squares- or least angle 
%regression- based calculation of $\bfthh$. Such optimisation schemes cannot be 
%extended to also optimise $\bfw$ jointly, without explicitly deriving the 
%joint-optimisation formulas. 

\subsection{Proxy surrogate models for the inner optimisation} 
\label{sec:Meth:DRSM:proxy}

Albeit non-intrusive and having a relatively low dimension, the inner 
optimisation in \eqref{eq:DRSM_inner_loop} is in general the driving cost of 
DRSM. 
Indeed, calculating the parameters of a single high-resolution modern surrogate 
may require anywhere between a few seconds and several minutes.
To reduce the related computational cost, it is often possible to solve 
\textit{proxy surrogate} problems, \textit{i.e.} using simplified surrogates 
that, while not being as accurate as their full counterparts, are easier to 
parametrise.
A simple example would be to prematurely stop the optimisation in the inner 
loop in \eqref{eq:DRSM_inner_loop}, or to use isotropic kernels for 
kernel-based surrogates such as Kriging or support vector machines instead of 
their more accurate, but costly to train, anisotropic counterparts.
Once the outer loop optimisation completes on the proxy surrogate, thus 
identifying the quasi-optimal DR parameters $\bfwh$, a single high-accuracy 
surrogate is then computed on the compressed experimental design $\acc{\cz= 
g(\cx;\bfwh), \bfy}$. Further discussion on this topic can be found in Sections 
\ref{sec:Meth:SM:Kriging} and \ref{sec:Meth:SM:PCE}.

\section{Selected compression and surrogate modelling techniques used in this 
paper}
\label{sec:Meth:Selected DR and SM}
Due to the non-intrusiveness in the design of the DRSM method proposed in 
\secref{sec:Meth:DRSM}, no specific dimensionality reduction or surrogate 
modelling technique has been introduced yet. In the following section, two 
well-known dimensionality reduction (namely principal component analysis and 
kernel-principal component analysis) and two surrogate modelling techniques 
(Kriging and polynomial chaos expansions) are introduced to showcase the DRSM 
methodology on several example applications in \secref{sec:Applications}. Only 
the main concept and notation is reminded so that the paper is self-consistent. 

\subsection{Principal component analysis} \label{sec:Meth:DR:PCA}

Principal Component Analysis (PCA) is a dimensionality reduction technique that 
aims at calculating a linear basis of $\ve{X}$ with reduced dimensionality that 
preserves the sample variance \citep{Pearson01}. Given a sample of the input 
random vector $\cx = \acc{\bfx^{(1)},\dots,\bfx^{(N)}}$, the PCA algorithm is 
based on the eigen-decomposition of the sample covariance matrix $\ve{C}$:
\begin{equation} \label{eq:PCA_Covariance}
\ve{C} = \frac{1}{N} \bar{\cx}^\top \bar{\cx},
\end{equation}
of the form:
\begin{equation} \label{eq:PCA_eigendecomp}
\ve{C} \ve{v}^{(i)} = \lambda^{(i)} \ve{v}^{(i)} \, , \, i=1 \enu M
\end{equation}
where $\bar{\cx}$ denotes the centred (zero mean) experimental design, 
$\lambda^{(i)}$ denotes each eigenvalue of $\ve{C}$ and $\ve{v}^{(i)}$ the 
corresponding eigenvector. The dimensionality reduction transformation reads:
\begin{equation} \label{eq:PCA_fwtrans}
\cz = \bar{\cx} \, \mat{V}
\end{equation}
where $\mat{V}$ is the $M\times m$ collection of the $m$ eigenvectors of 
$\mat{C}$ with maximal eigenvalues. Those eigenvectors are called the 
\textit{principal components} because they correspond to the reduced basis of 
$\cx$ with maximal variance.
Based on the general DR perspective that was presented in \secref{sec:Meth:DR}, 
PCA is a linear transformation of the form $\cz = g(\cx;w)$, where the only 
parameter to be selected is the dimension $m$ of the reduced space, \ie{} $w=m$.

\subsection{Kernel principal component analysis}\label{sec:Meth:DR:KPCA}

Kernel PCA (KPCA) is the reformulation of PCA in a high-dimensional space that 
is constructed using a kernel function \citep{Scholkopf1998_KPCA}. 
KPCA has been primarily used for feature extraction purposes in pattern 
recognition problems \citep{Scholkopf1998_KPCA,Ince2007,Wang2011}, as well as 
image de-noising \citep{Mika1999,Bakir2004}. 

A kernel function applied on two elements $\bfx^{(i)}, \bfx^{(j)} \in 
\cd_{\bfx}$ has the following form:
\begin{equation} \label{eq:kappa_definition}
\kappa\prt{\bfx^{(i)}, \bfx^{(j)}} = \Phi\prt{\bfx^{(i)}} \cdot 
\Phi\prt{\bfx^{(j)}} 
\end{equation}
where $\Phi(\cdot)$ is a function that performs the mapping $ \Phi: \cd_{\bfx} 
\rightarrow \ch $ and $\ch$ is known as the feature space. 
Based on \eqref{eq:kappa_definition}, the so-called \textit{kernel trick} is 
applied, which refers to the observation that, if the access to $\ch$ only 
takes place through inner products, then there is no need to  explicitly define 
$\Phi(\cdot)$. The result of the inner product can be directly calculated using 
$\kappa(\cdot,\cdot)$. 
Kernel PCA is a non-linear extension of PCA where the kernel trick is used to 
perform PCA in $\ch$. The principal components in $\ch$ are obtained from the 
eigen-decomposition of the sample covariance matrix $\bfC_\ch$, analogously to 
the PCA case in \eqref{eq:PCA_Covariance}.

However, in KPCA the eigen-decomposition problem:
\begin{equation} \label{eq:KPCA_eigenprob}
\bfC_\ch \ve{v}^{(i)} = \lambda_i \ve{v}^{(i)} \, , \, i = 1 \enu N
\end{equation}
is intractable, since $\bfC_\ch$ cannot in general be computed ($\ch$ might 
even be infinitely dimensional). 
This problem is by-passed by observing that each eigenvector belongs to the 
span of the samples $\Phi\prt{\bfx^{(1)}}, \ldots, \Phi\prt{\bfx^{(N)}}$,  
therefore scalar coefficients $\alpha_k^{(i)}$ exist, such that each 
eigenvector $\ve{v}^{(i)}$ can be expressed as the following linear combination 
\citep{Scholkopf1998_KPCA}:
\begin{equation} \label{eq:KPCA_v}
\ve{v}^{(i)} = \sum_{k=1}^N \alpha_k^{(i)} \Phi\prt{\bfx^{(k)}} \, , \, i = 1 
\enu N .
\end{equation}
Based on \eqref{eq:KPCA_v} it can be shown that the eigen-decomposition problem 
in \eqref{eq:KPCA_eigenprob} can be cast as:
\begin{equation} \label{eq:KPCA_alphas}
\bfK  \bfal^{(i)} = \lambda^{(i)}  \bfal^{(i)}  \, , \, i = 1 \enu N 
\end{equation}
where $\bfK$ is the kernel matrix with elements:
\begin{equation}
K_{ij} = \kappa\prt{\bfx^{(i)}, \bfx^{(j)}}.
\end{equation}
As for the case of PCA, $\cz$ is calculated by projecting $\cx$ on the $m$ 
principal axes $\acc{\ve{v}^{(i)}\, , \, \allowbreak i = 1  \enu m}$ 
corresponding to the $m$ largest eigenvalues.
\citet{Scholkopf1998_KPCA} showed that $\cz$ can be directly computed based 
only on the values of the eigenvector expansion coefficients $\alpha_k^{(i)}$ 
and the kernel matrix $\mat{K}$. The $k$-th component of the $i$-th sample of 
$\cz$, denoted by $z_k^{(i)} $ is given by; 

\begin{equation}\label{eq:KPCA_FWtrans} 
z_k^{(i)} = \Phi\prt{\bfx^{(i)}}^\mathsf{T} \ve{v}^{(k)} = \sum_{j=1}^N 
\alpha_k^{(j)} \kappa\prt{\bfx^{(i)}, \bfx^{(j)}} 
\end{equation}

The key ingredient of KPCA is arguably the kernel function $\kappa$. 
In this paper two kernels are considered, namely the \textit{polynomial} 
kernel: 
\begin{equation}
\label{eq:KPCA_kernel_poly}
\kappa(\bfx, \bfx'; \bfw) = \prt{ w_1 \bfx^\mathsf{T}\bfx' + w_2}^{w_3}\, , 
\,   w_1>0, w_2 \geq 0, w_3 \in \Nn,
\end{equation}
and the \textit{Gaussian} kernel:
\begin{equation}
\label{eq:KPCA_kernel_gauss_aniso}
\kappa(\bfx, \bfx'; \bfw) = \exp \prt{ - 
\frac{1}{2}\sum_{k=1}^{M}\frac{1}{w_k^2} \prt{x_k- x_k' }^2} \, , \,  w_k > 0 
\, , \, k = 1 \enu M.
\end{equation}
A special case of the Gaussian kernel is the \textit{isotropic} Gaussian kernel 
(also known as \textit{radial basis function}) that simply assumes the same 
parameter value $w_k$ for all components of $\bfx$. 
Note that KPCA using a polynomial kernel with parameters $w_1=1$, $w_2=0$ and 
$w_3=1$ is identical to PCA, since  $\Phi(\bfx) = \bfx$. 
A discussion on the equivalence between PCA and KPCA with linear kernel 
($w_3=1$) for arbitrary values of $w_1,w_2$ can be found in 
\ref{sec:Appendix:PCA_vs_linKPCA}. 
From \eqref{eq:KPCA_FWtrans} it follows that $\cz$ can be expressed as $\cz = 
g(\cx; \bfw)$ where  $\bfw$ encompasses both the kernel parameters and the 
reduced space dimension $m$.

In the context of unsupervised learning, two methods to infer the values of 
$\bfw$ from $\cx$ are considered. 
The \textit{distance preservation} method aims at optimising $\bfw$ in such a 
way that the Euclidean distances between the samples are preserved between the 
original and the feature space \citep{Weinberger2004}. This is expressed by the 
following objective function: 
\begin{equation} \label{eq:KPCA_Jdist}
J_{dist} (\bfw ; \cx) = \sum_{i=1}^{N}\sum_{j=1}^{N} \prt{d_{ij} - 
\delta_{ij}}^2  
\end{equation} 
where 
\begin{equation} \label{eq:KPCA_Jdist_d}
d_{ij} = \norm{\bfx^{(i)} - \bfx^{(j)}}
\end{equation} 
and
\begin{equation} \label{eq:KPCA_Jdist_delta_1}
\delta_{ij} = \norm{\Phi(\bfx^{(i)},\bfw) - \Phi(\bfx^{(j)},\bfw) }.
\end{equation} 
By expanding the norm expression in \eqref{eq:KPCA_Jdist_delta_1} it is 
straightforward to show that: %
\begin{equation} \label{eq:KPCA_Jdist_delta_2}
\delta_{ij} = \sqrt{K_{ii} + K_{jj} - 2 K_{ij}},
\end{equation} 
hence the value of  $\delta_{ij}$ is readily available from the kernel matrix 
$\mat{K}$. 

The \textit{reconstruction error}-based method  aims at  optimising $\bfw$ in 
such a way that the so-called pre-image, $\tilde{\bfx} = g^{-1}(\bfz,\bfw')$, 
of $\bfz = g(\bfx, \bfw)$ approximates $\bfx$ as close as possible 
\citep{Alam2014}. This is expressed by the following objective function: 
\begin{equation} \label{eq:KPCA_Jrecon}
J_{recon} (\bfw ; \cx) = \frac{1}{N} \sum_{i=1}^{N} \norm{\bfx^{(i)} - 
\tilde{\bfx}^{(i)}}^2
\end{equation}
In contrast to PCA, calculating $\tilde{\bfx}$ is non-trivial, an issue that is 
known as the \textit{pre-image problem} (see \eg{}\citet{Kwok2003}). The 
approach for dealing with this problem is the one adopted by the popular 
\textsc{python} package \textsc{scikit-learn} \citep{Pedregosa2011}, which is 
based on \citet{Bakir2004}. After performing the KPCA transform $\cx \mapsto 
\cz$, the (non-unique) pre-image of a new point $\bfz$ is computed by 
kernel-ridge regression using a new kernel function $\kappa_{pre}$:
\begin{equation} \label{eq:KPCA_preimage_main}
\tilde{\bfx} = \bm{\beta}^\mathsf{T} \bm{l}(\bfz) ,
\end{equation}
where:  
\begin{equation} \label{eq:KPCA_preimage_l}
\bs{\ell}(\bfz) =\acc{\kappa_{pre}(\bfz, \, \bfz^{(j)}), \; j=1 \enu N},
\end{equation}
and $\bm{\beta}$ are the kernel-ridge regression coefficients. They are 
calculated as follows: 
\begin{equation} \label{eq:KPCA_preimage_beta}
\bm{\beta} = \prt{\bm{L} + r \bm{I}_N }^{-1} \cx \qquad L_{ij} =
\acc{\kappa_{pre}\prt{\bfz^{(i)},\bfz^{(j)}},\; i,j=1\enu N}
\end{equation}
where $r$ is a regularisation parameter and $\bm{I}_N$ is the $N$-dimensional 
identity matrix. In \citet{Pedregosa2011} 
and in this paper, we use for simplicity the same kernel for the pre-image 
problem as for KPCA, \ie{} $\kappa_{pre}\prt{\cdot, \cdot}$ is chosen equal to 
$\kappa \prt{\cdot, \cdot}$. 

Note that, in the unsupervised learning literature, the reduced space 
dimension, $m$, is typically not part of $\bfw$, \ie{} only the kernel 
parameters are considered when minimising the objective function in 
\eqref{eq:KPCA_Jdist} or \eqref{eq:KPCA_Jrecon}.

\subsection{Kriging} \label{sec:Meth:SM:Kriging}
Kriging, a.k.a. Gaussian process modelling, is a surrogate modelling technique 
which assumes that the true model response is a realisation of a Gaussian 
process. 
This technique is appealing because it enables the efficient approximation of 
highly non-linear models, it interpolates the experimental design, and it 
provides native point-wise error estimation.
A Kriging surrogate is described by the following equation 
\citep{santner_design_2003}:
\begin{equation} \label{eq:KrigingGeneral}
\widehat{\cm}(\bfx) = \ve{\beta}^\top \ve{f}(\ve{x}) + \sigma^2 Z(\bfx)
\end{equation}
where $\ve{\beta}^\top \ve{f}(\ve{x})$ is the mean value of the Gaussian 
process, also called \emph{trend},  $\sigma^2$ is the Gaussian process 
variance  and $Z(\bfx)$ is a zero-mean, unit-variance Gaussian process. This 
process is fully characterised by the auto-correlation function between two 
sample points $R(\bfx,\bfx';\bfth)$. 
The hyperparameters $\bfth$ associated with the correlation function 
$R(\cdot;\bfth)$ are typically unknown and need to be estimated from the 
available observations. 
Various correlation functions can be found in the literature 
\citep{Rasmussen2006,santner_design_2003}, including the \textit{linear}, 
\textit{exponential}, \textit{Gaussian} (a.k.a. \textit{squared exponential}) 
and \textit{Mat\'ern} functions. 
In this paper the separable Mat\'ern correlation family is chosen:
\begin{equation}\label{eq:Kriging_Matern_general}
R\prt{\abs{\bfx - \bfx'}; \bm{l}, \nu} = \prod_{i=1}^{M} \frac{1}{2^{\nu -1} 
\Gamma(\nu)} \prt{\sqrt{2\nu}  \frac{\abs{x_i - x'_i}}{l_i} }^\nu \kappa_\nu 
\prt{\sqrt{2\nu}  \frac{\abs{x_i - x'_i}}{l_i}}, 
\end{equation}
where $\bfx$, $\bfx'$ are two samples in the input space $\cd_x$, $\bm{l} = 
\acc{l_i>0,\, i=1 \enu M}$ are the scale parameters (also called 
\textit{correlation lengths}), $\nu \geq 1/2$ is the shape parameter, 
$\Gamma(\cdot)$ is the Euler Gamma function and $\kappa_\nu(\cdot)$ is the 
modified Bessel function of the second kind (a.k.a. Bessel function of the 
third kind). The values $\nu=3/2$ and $\nu=5/2$ of the shape parameter are 
commonly used in the literature. The \textit{isotropic} variant of the Mat\'ern 
correlation family assumes a fixed correlation length value $l$ in 
\eqref{eq:Kriging_Matern_general} over all $M$ input variables. 

Regarding the trend part $\ve{\beta}^\top \ve{f}(\ve{x})$ in 
\eqref{eq:KrigingGeneral}, the general formulation of \textit{universal 
Kriging} is adopted, which assumes that the trend is composed of a linear 
combination of $P$ pre-selected functions $\acc{f_i(\bfx),\, i=1 \enu P}$, 
\ie{}:
\begin{equation} \label{eq:Kriging_trend}
\ve{\beta}^\top \ve{f}(\ve{x}) = \sum_{i=1}^P \beta_i f_i(\bfx),
\end{equation}
where $\beta_i$ is the trend coefficient of each function. 

The Gaussian assumption states that the vector formed by the true model 
responses, $\vY$ and the prediction, $\widehat{Y}(\bfx)$, at a new point 
$\bfx$, has a joint Gaussian distribution defined by:
\begin{equation} 
\bra{  
	\begin{matrix}
	\widehat{Y}(\bfx) \\ \vY
	\end{matrix}
}
\sim 
\mathcal{N}_{N+1} \prt{  
	\bra{ \begin{matrix}
		\bm{f}^\top(\bfx) \bm{\beta} \\ \mat{F} \bm{\beta}
		\end{matrix}
	}
	, 
	\sigma^2 
	\bra{
		\begin{matrix}
		1 & \bm{r}^\top(\bfx) \\ 
		\bm{r}(\bfx) & \mat{R}
		\end{matrix}
	}
}
\end{equation} 
where $\bm{F}$ is the information matrix of generic terms:
\begin{equation}
F_{ij}  = f_j(\ve{x}^{(i)})~,~i=1 \enu N,~j=1 \enu P,
\end{equation}
$\bm{r}(\bfx)$ is the vector of cross-correlations between the prediction point 
$\bfx$ and each one of the observations whose terms read:
\begin{equation}
r_{i}(\bfx) = R(\bfx,\bfx^{(i)};\bm{\theta}), ~i=1 \enu N.
\label{eq:r0}
\end{equation}
$\bm{R}$ is the correlation matrix given by:
\begin{equation}
R_{ij} = R(\bfx^{(i)},\bfx^{(j)};\bm{\theta}), ~i,j=1 \enu N.
\end{equation}
The mean and variance of the Gaussian random variate $\widehat{Y}(\bfx)$ 
(a.k.a. mean and variance of the Kriging predictor) can be calculated based on 
the best linear unbiased predictor (BLUP) from \citet{santner_design_2003}:
\begin{equation} \label{eq:TheoryPredicorMean}
\mu_{\widehat{Y}}(\ve{x})  =\ve{f}(\ve{x})^\top \ve{\beta} + 
\ve{r}(\ve{x})^\top \mat{R}^{-1}\left (\vY-\mat{F}\ve{\beta} \right )\, ,
\end{equation}
\begin{equation} \label{eq:TheoryPredicorVariance}
\sigma_{\widehat{Y}}^2(\ve{x})  = \sigma^2 \left(  
1-\ve{r}^\top(\ve{x})\mat{R}^{-1}\ve{r}(\ve{x}) + \ve{u}^\top(\ve{x}) 
(\mat{F}^\top\mat{R}^{-1}\mat{F})^{-1}\ve{u}(\ve{x})  \right)
\end{equation}
where:
\begin{equation} \label{eq:AKG:TheoryCalcBeta}
\ve{\beta}  = \left( \mat{F}^\top \mat{R}^{-1} \mat{F}  
\right)^{-1}\mat{F}^\top\mat{R}^{-1} \vY
\end{equation}
is the generalised least-squares estimate of the underlying regression problem 
and 
\begin{equation}
\label{eq:AKG:TheoryPredicorU}
\ve{u}(\ve{x}) = \mat{F}^\top \mat{R}^{-1}\ve{r}(\ve{x}) - \ve{f}(\ve{x}).
\end{equation}
The mean response in \eqref{eq:TheoryPredicorMean} is considered as the output 
of a Kriging surrogate, \ie{} $\widehat{\cm}(\bfx) = \mu_{\widehat{Y}}(\bfx)$. 
It is important to note  that the Kriging model interpolates the data, \ie{}:
\begin{equation}\label{eq:Kriging_interpolates}
\mu_{\widehat{Y}}(\bfx) = \cm(\bfx), \quad \sigma_{\widehat{Y}}^2(\bfx) = 0, 
\quad \forall\,  \bfx \in \cx  
\end{equation}
The equations that were derived for the best linear unbiased Kriging predictor  
assumed that the covariance function $\sigma^2 R(\cdot;\bfth)$ is known. In 
practice however, the family and other properties of the correlation function 
need to be selected \textit{a priori}. The hyperparameters $\bm{\theta}$, the 
regression coefficients $\bm{\beta}$  and the variance $\sigma^2$ need to be 
estimated based on the available experimental design. 

The optimal estimates of the correlation parameters $\widehat{\bfth}$ are 
determined by minimising the generalisation error of the Kriging surrogate, 
based on the leave-one-out cross-validation error 
\citep{santner_design_2003,Bachoc2013b}:
\begin{equation} \label{eq:Kriging_thetaCV}
\ve{\theta}_{CV} = \underset{\cd_{\ve{\theta}}}{\arg \min} \sum_{i=1}^{K} 
\left( \cm(\ve{x}^{(i)}) - \mu_{\widehat{Y}, (-i)}(\ve{x}^{(i)})  \right) ^2  , 
\end{equation}
where $\mu_{\widehat{Y}, (-i)}(\ve{x}^{(i)})$ corresponds to the mean value of 
a Kriging predictor that was built from the samples $\cx \,\backslash 
\acc{\bfx^{(i)}, y^{(i)}}$, evaluated at $\ve{x}^{(i)}$. 
The computational cost for calculating the terms $\mu_{\widehat{Y}, 
(-i)}(\ve{x}^{(i)})$ can be significantly reduced as shown in 
\citet{Dubrule1983}. First, the following matrix inversion is performed: 
\begin{equation} \label{eq:Dubrule_B}
\mat{B} = \bra{ \begin{matrix}
	\sigma^2 \mat{R} & \mat{F} \\
	\mat{F}^\mathsf{T} & \mat{0}
	\end{matrix}}^{-1}.
\end{equation}
Then $\mu_{\widehat{Y},(-i)}(\ve{x}^{(i)})$ is calculated as follows:
\begin{equation} \label{eq:Dubrule_mean}
\mu_{\widehat{Y},(-i)}(\ve{x}^{(i)}) = - \sum_{j=1,j\neq i}^N 
\frac{\mat{B}_{ij}}{\mat{B}_{ii}} \, y^{(j)}.
\end{equation}
In this work we use cross-validation for estimating the correlation parameters 
instead of the maximum likelihood method \citep{santner_design_2003}).
This is motivated by the comparative study in \citet{Bachoc2013b}  between 
maximum likelihood (ML) and CV estimation methods. 
The CV method is expected to perform better in cases that the correlation 
family of the Kriging surrogate is not identical to the one of the true model. 
This is typically the case in practice and in the application examples in 
\secref{sec:Applications}. 

Determining the optimal parameters  $\ve{\theta}_{CV} $ in  
\eqref{eq:Kriging_thetaCV} leads to a complex multi-dimensional optimisation 
problem. 
Common optimisation algorithms employed to solve \eqref{eq:Kriging_thetaCV} can 
be cast into two categories: local and global. 
Local methods are usually gradient-based, such as the BFGS algorithm 
\citep{bazaraa2013bfgs}, and search locally in the vicinity of the starting 
point. 
This makes them prone to get stuck at local minima, although they can be 
computationally efficient due to the use of gradients. 
Global methods such as genetic algorithms \citep{Goldberg1989} do not rely on 
local information such as the gradient. They seek the global minimum by various 
adaptive resampling strategies within a bounded domain. 
This often leads to considerably more objective function evaluations compared 
to local methods.  

As mentioned in \secref{sec:Meth:DRSM:proxy}, to alleviate the computational 
costs in the inner loop optimisation in \eqref{eq:DRSM_inner_loop}, an 
inexpensive-to-calibrate Kriging surrogate is built.  
To this end, the isotropic version of the Mat\'ern correlation family is used, 
combined with low computational budget optimisation of the correlation 
parameters. 
For calculating the final, high-accuracy, Kriging surrogate an optimisation 
with high-computational budget is performed instead, combined with the use of 
an anisotropic correlation family. 
The introduction of  anisotropy is expected to improve the generalisation 
performance the metamodel, as shown for instance in the study by 
\citet{MoustaphaJRUES2018}.

\subsubsection{Polynomial chaos expansions} \label{sec:Meth:SM:PCE}

Polynomial chaos expansions represent a different class of surrogate models 
that has seen widespread use in the context of uncertainty quantification due 
to their flexibility and efficiency.
Due to the focus on data-driven problems, in this work we focus on 
non-intrusive polynomial chaos expansions \citep{TorreJCP2018}.
Consider that $\ve{X} \in \Rr^M$ is a random vector with independent components 
described by the joint PDF $f_{\ve{X}}$ and that the model output ${Y}$ in 
\eqref{eq:true_model} has finite variance. Then the polynomial chaos expansion 
of $\cm(\ve{X})$ is given by:
\begin{equation}
\label{eqn:PCE:PCE}
Y = \cm(\Ve{X})  = \sum\limits_{\ua\in\mathbb{N}^M} c_{\ua} 
{\Psi}_{\ua}(\Ve{X})  
\end{equation}
where the $\Psi_{\ua} (\Ve{X})$ are multivariate polynomials orthonormal  with 
respect to $f_{\Ve{X}}$, $\ua \in \mathbb{N}^M$ is a multi-index that 
identifies the components of the multivariate polynomials ${\Psi}_{\ua}$ and 
the 
$c_{\ua} \in \mathbb{R}$ are the corresponding coefficients.

In practice, the series in \eqref{eqn:PCE:PCE} is truncated to a finite sum, by 
introducing the truncated polynomial chaos expansion:
\begin{equation}
\label{eq:PCE_truncatedPCE}
\cm(\Ve{X}) \approx \widehat{\cm}(\Ve{X}) = \sum\limits_{\ua\in\Ca} c_{\ua} 
{\Psi}_{\ua}(\Ve{X})  
\equiv
\bm{c}^\top \ve{\Psi}(\bfx)
\end{equation}
where $\Ca \subset \mathbb{N}^M$ is the set of selected multi-indices of 
multivariate polynomials.
A typical truncation scheme consists in selecting multivariate polynomials up 
to a total degree $p$, \ie{} $\Ca = \acc{\ua \in \Nn^M \, : \, \norm{\ua}_1 
\leq p }$, with $\norm{\ua}_1 = \sum_{i=1}^{M}\alpha_i$. The corresponding 
number of terms in the truncated series rapidly increases with $M$, giving rise 
to the ``curse of dimensionality''. Other truncation strategies effective in 
higher dimension are discussed, \eg{}, in \citet{BlatmanPEM2010,Jakeman2015}.

The polynomial basis $\Psi_{\ua}(\Ve{X})$ in \eqref{eq:PCE_truncatedPCE} is 
traditionally built  starting from a set of \textit{univariate orthonormal 
polynomials} 
$\phi^{(i)}_k(x_i)$ which satisfy:
\begin{equation}
\label{eqn:PCE:Theory:UnivariateOrthonormalPoly}
\left< \phi^{(i)}_j(x_i),\phi^{(i)}_k(x_i) \right> \eqdef \int_{\cd_{X_i}} 
\phi^{(i)}_j(x_i)\phi^{(i)}_k(x_i) 
f_{X_i}(x_i)\di x_i = 	\delta_{jk}
\end{equation}
where $i$ identifies the input variable w.r.t. which they are orthogonal, as 
well as the
corresponding polynomial family, $j$ and $k$ the corresponding polynomial 
degree, $f_{X_i}(x_i)$ is 
the $i^{th}$-input marginal distribution and $\delta_{jk}$ is the Kronecker 
symbol. 
Note that this definition of inner product can be interpreted as the 
expectation value of the 
product of the multiplicands. 
The multivariate polynomials $\Psi_{\ua}(\Ve{X})$ are then assembled as the 
tensor product of 
their  univariate counterparts:
\begin{equation}
\label{eqn:PCE: multivariate polynomials}
\Psi_{\ua}(\ve{x}) \eqdef \prod_{i=1}^M \phi^{(i)}_{\alpha_i} (x_i)
\end{equation}
For standard distributions, such as uniform, Gaussian, gamma, beta, the 
associated families of orthogonal polynomials are well-known \citep{Xiu2002}.  
Orthogonal polynomials can be constructed numerically w.r.t. any distribution 
(including non-parametric ones like those obtained by kernel density smoothing) 
by means of Gram-Schmidt orthonormalisation (a.k.a. Stieltjes procedure for 
polynomials \citep{Gautschi2004}).

The expansion coefficients $\bm{c} = \acc{c_{\bm{\alpha}},\, \bm{\alpha} \in 
\Ca \subset \Nn^M }  $ in \eqref{eq:PCE_truncatedPCE} are calculated by 
minimising the expectation of least-squares residual \citep{Berveiller2006}:
\begin{equation} \label{eq:PCE_leastSquares_min}
\widehat{\bm{c}} = \arg \min \Esp{ \left(\bm{c}\tr
	{\Psi}(\Ve{X}) - \cm(\Ve{X}) \right)^2}.
\end{equation}
In the context of DRSM, the set of input parameters $\bfw$ for a PCE surrogate 
consists in $\bm{\theta} = \acc{p, \bm{c}}$, \ie{} the maximal degree of the 
truncated expansion and the associated coefficients. 
Due to the quadratic programming nature of the minimisation in 
\eqref{eq:PCE_leastSquares_min} and the linearity of PCE (see 
\eqref{eq:PCE_truncatedPCE}), we adopt the adaptive sparse-linear regression 
based on least angle regression first introduced by \citet{BlatmanJCP2011}.

As for the case of Kriging, the LOO error (see \eqref{eq:epsilon_LOO}) is 
analytically available from the expansion coefficients \citep{BlatmanJCP2011}:
\begin{equation}
\label{eq:PCE_LOO}
\varepsilon_{LOO} = {\sum\limits_{i = 1}^N \left( 
	\frac{\cm(\Ve{x}^{(i)}) - 
		\widehat{\cm}^{PC}(\Ve{x}^{(i)})}{1-h_i}\right)^2}\bigg/{\sum\limits_{i 
		= 1}^N 
	\left(\cm(\ve{x}^{(i)}) - \widehat{\mu}_Y\right)^2},
\end{equation}
where $h_i$ is the $i^{th}$ component of the vector given by: 
\begin{equation}
\label{eq:PCE_LOO_h}
\Ve{h} = \text{diag}\left(\mat{A}(\mat{A}\tr\mat{A})^{-1} \mat{A}\tr\right),
\end{equation}
and $\mat{A}$ is the experimental matrix with entries $A_{ij} = 
\Psi_j\prt{\bfx^{(i)})}$.

To calculate the \textit{proxy} PCE surrogates used during the DRSM 
optimisation phase (see \secref{sec:Meth:DRSM:proxy}), the input variables in 
$\bfz$, are modelled as uniformly distributed (with bounds inferred from the 
experimental design) and independent. 
The PCE coefficients are computed by solving \eqref{eq:PCE_leastSquares_min} 
using the ordinary least squares method \citep{Berveiller2006}. 
To calculate the PCE coefficients of the final, high-accuracy, surrogate 
$\widehat{\cm}(g(\bfx;\widehat{\bfw}))$,
the distributions of the input variables  are fitted using kernel-smoothing, 
while retaining the independence assumption, motivated by the results in 
\citet{TorreJCP2018}.  
In the latter it is shown that ignoring the input dependencies results in PCE 
surrogates with improved point-wise accuracy, because it avoids highly 
non-linear input transforms that introduce additional complexity. 
In addition, a sparse solution is obtained by solving the optimisation problem 
in \eqref{eq:PCE_leastSquares_min}
using least angle regression \citep{BlatmanJCP2011} instead of ordinary least 
squares.

% #######################	
\section{Applications} \label{sec:Applications}
% #######################

The performance of DRSM is evaluated on the following applications: (i) an 
artificial analytic function with $20$ unstructured inputs and approximately 
known intrinsic dimension, (ii) a realistic electrical engineering model with 
$80$ unstructured inputs and unknown intrinsic dimension and, (iii) a heat 
diffusion model with $16,000$ structured inputs and unknown intrinsic 
dimension.

For each example, DRSM is applied using KPCA for compression together with 
Kriging or polynomial chaos expansions for surrogate modelling. 
The surrogate performance is then compared, in terms of generalisation error, 
to the sequential application of unsupervised dimensionality reduction followed 
by surrogate modelling.
To improve readability, various details regarding the implementation of the 
optimisation algorithms and the surrogate models calibration are omitted from 
the main text and given in \ref{sec:Appendix:details} instead. 
All the surrogate modelling techniques were deployed with the 
\textsc{Matlab}-based uncertainty quantification software \textsc{UQLab} 
\citep{Marelli2014,UQLabPCE,UQLabKriging}.

%%%%%
\subsection{Sobol' function} \label{sec:App:sobol}
%%%%%
The Sobol' function (also known as $g$-function) is a commonly used benchmark 
function in the context of uncertainty quantification. It reads:
\begin{equation} \label{eq:app_Sobol}
Y = \prod_{i=1}^{M} \frac{\abs{4X_i - 2} + c_i}{1 + c_i} \,,
\end{equation}
where $\bfX = \lbrace X_1 \enu X_M \rbrace$ are independent random variables 
uniformly distributed in the interval $[0,1]$ and $ \bfc = \lbrace c_1 \enu 
c_M  \rbrace^\mathsf{T}$ are non-negative constants. In this application, we 
chose $M = 20$ and the constants $\bfc$ given by 
\citet{KonakliRESS2016,Kersaudy2015}:

\begin{equation} \label{eq:app_Sobol_constants}
\bfc = \lbrace 1, 2, 5, 10, 20, 50, 100, 500, 500 \enu 500\rbrace^\mathsf{T}.
\end{equation}  

It is straightforward to see that the effect of each input variable $X_i$ to 
the output $Y$ is inversely proportional to the value of $c_i$. In other words, 
a small (resp. large) value of $c_i$ results in a high (resp. low) contribution 
of $X_i$ to the value of $Y_i$. 
For the given values of the constants $\bfc$, one would expect that, roughly, 
the first $4$ to $6$ variables can provide a compressed representation of 
$\bfX$ with minimal information loss regarding the input-output relationship. 

%that contains enough information to predict $Y$ with minimal error.

To showcase the performance of DRSM, an experimental design $\cx$, consisting 
of $800$ samples, is generated by Latin Hypercube sampling of the input 
distribution \citep{McKay1979}. Based on the samples in $\cx$ and the 
corresponding model responses $\bfy$, several combinations of KPCA, Kriging and 
PCE are tested within the DRSM framework. An additional set of $10^5$ 
validation samples $\acc{\cx_v, \bfy_v}$ is generated for evaluating the 
performance of the final surrogates.

\begin{figure}[!t]
	\centering
	
	\subfigure[Kriging - LOO error]{\label{fig:res_sobol_drsm_m_vs_error-kg-loo}
		\includegraphics[width=.47\textwidth,trim={0.2cm 0 2.0cm 
		0},clip]{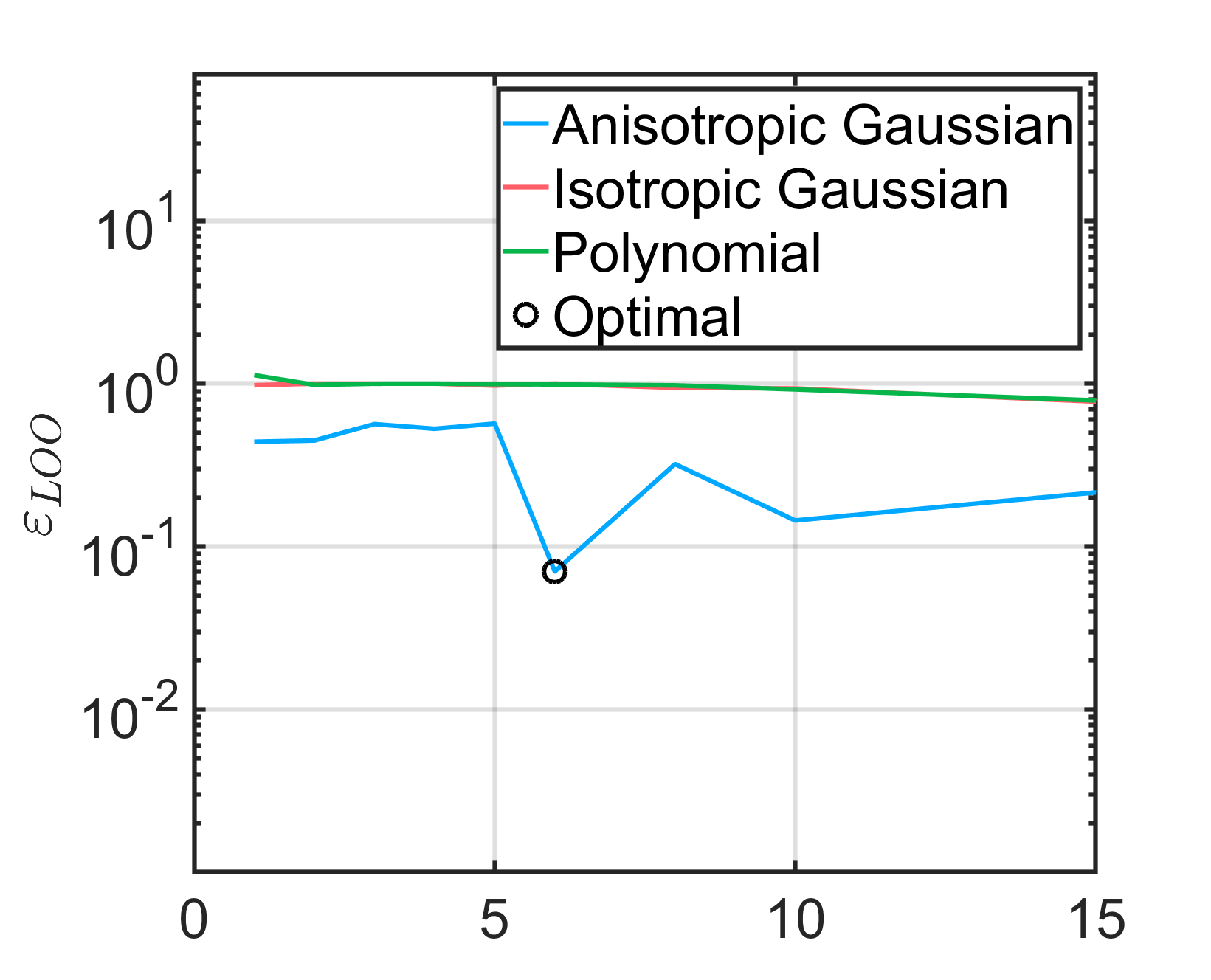}
	}
	\subfigure[PCE - LOO error]{\label{fig:res_sobol_drsm_m_vs_error-pce-loo}
		\includegraphics[width=.47\textwidth,trim={0.2cm 0 2.0cm 
		0},clip]{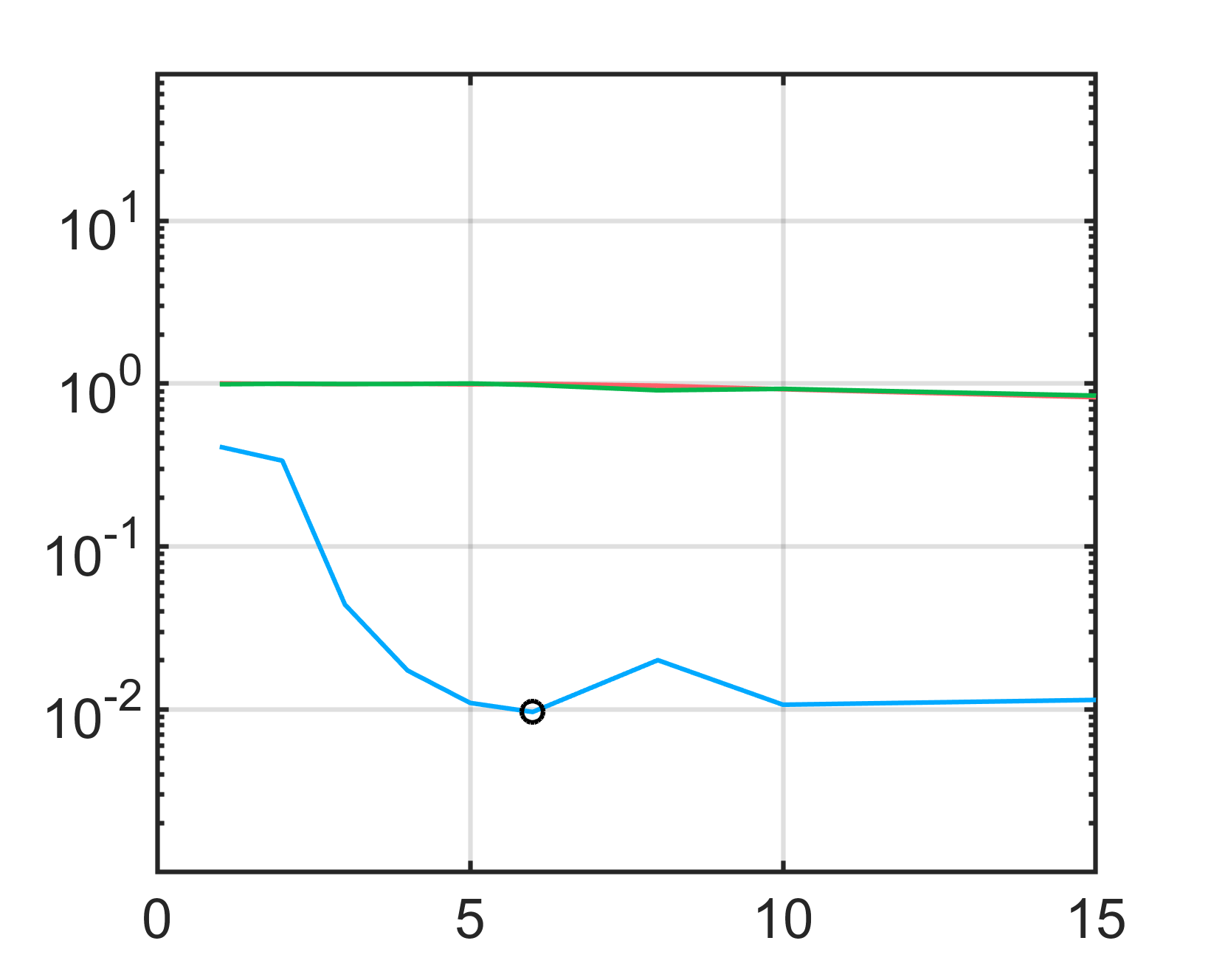}
	}
	
	\subfigure[Kriging - Validation 
	error]{\label{fig:res_sobol_drsm_m_vs_error-kg-rmse}
		\includegraphics[width=.47\textwidth,trim={0.2cm 0 2.0cm 
		0},clip]{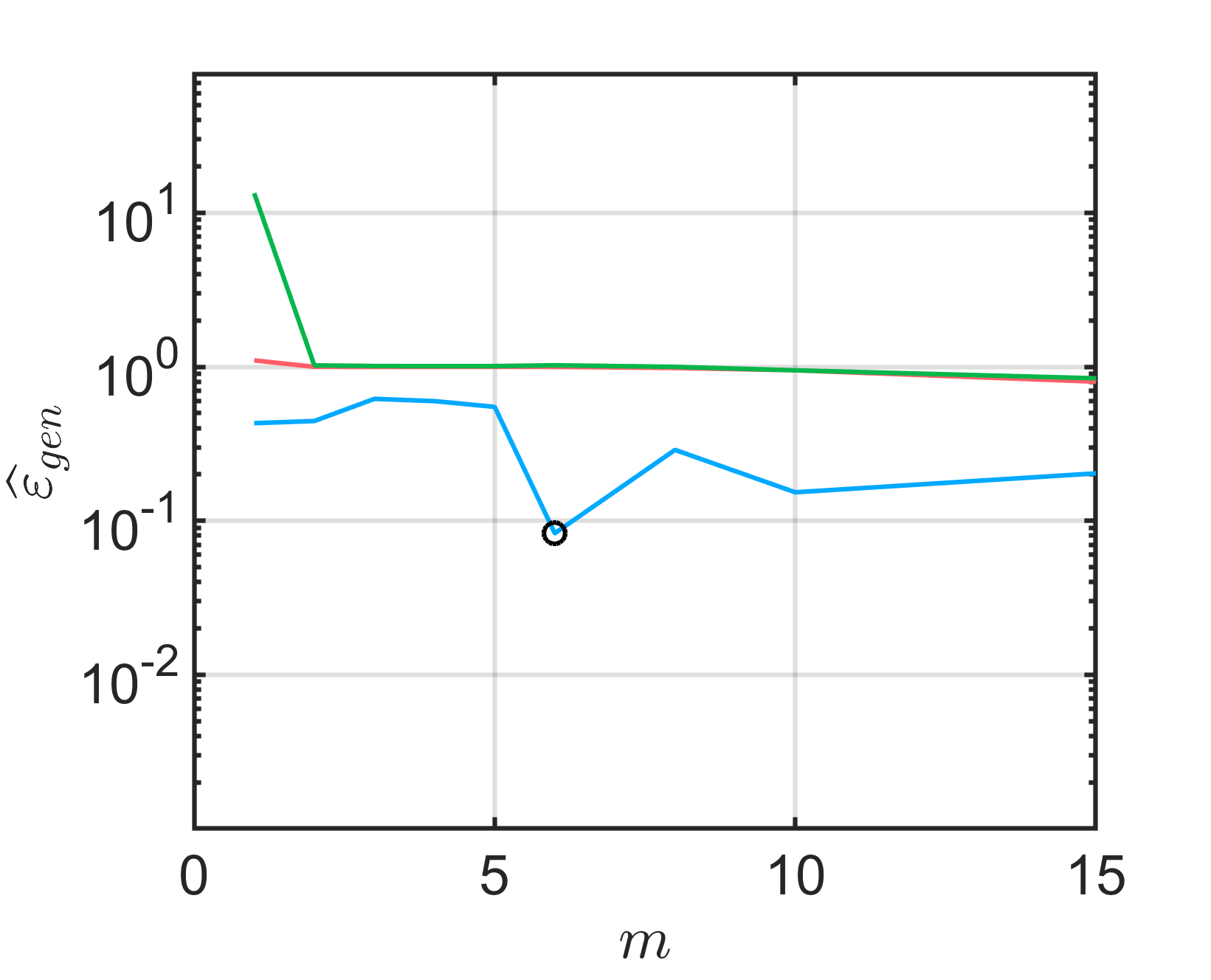}
	}
	\subfigure[PCE - Validation 
	error]{\label{fig:res_sobol_drsm_m_vs_error-pce-rmse}
		\includegraphics[width=.47\textwidth,trim={0.2cm 0 2.0cm 
		0},clip]{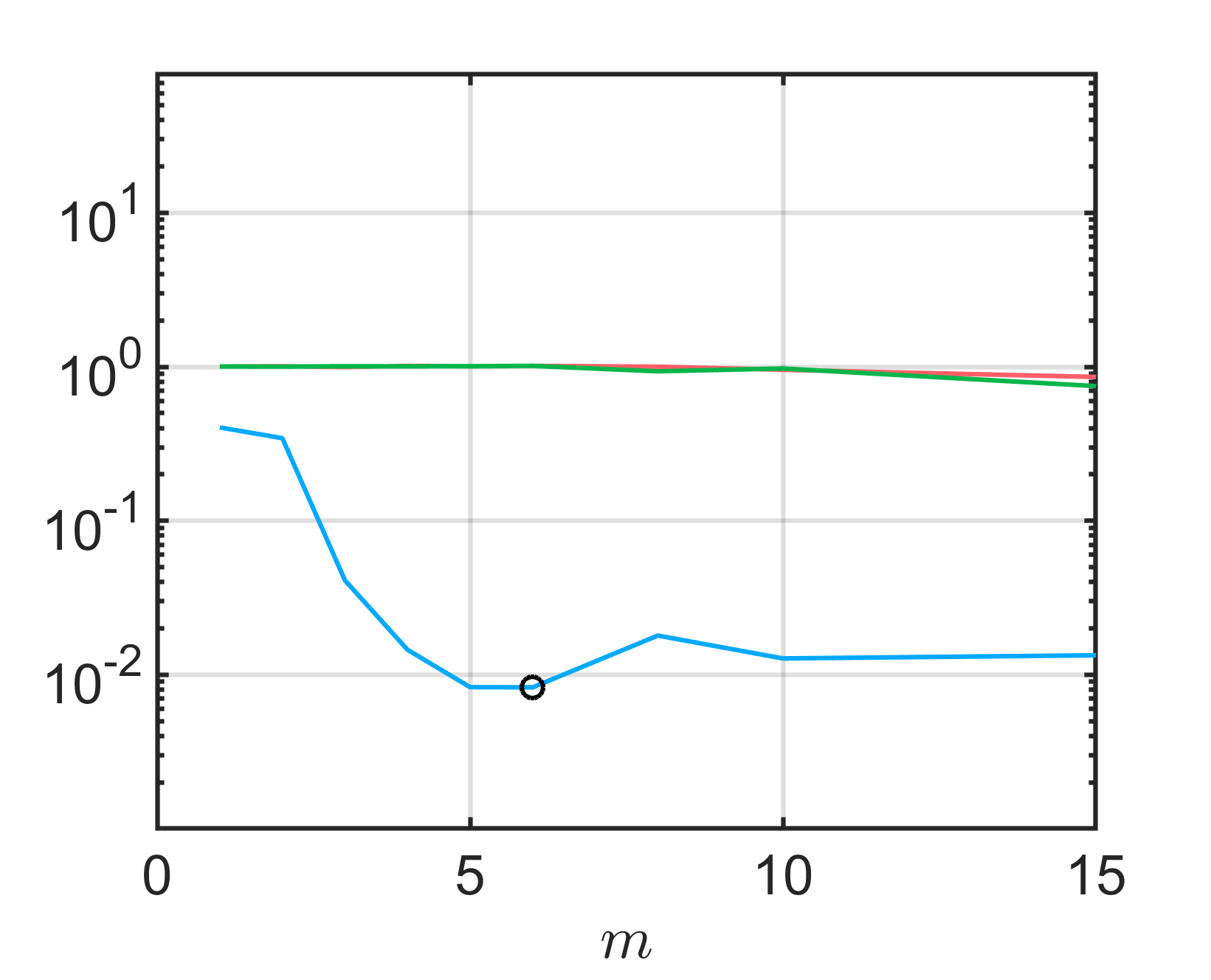}
	}
	
	\caption{Sobol' function: Error estimates of the DRSM surrogate as a 
	function of the reduced space dimension. Kernel PCA is used with isotropic 
	(resp.anisotropic) Gaussian as well as polynomial kernels.}
	\label{fig:res_sobol_drsm_m_vs_error}
\end{figure}

The first analysis consists in comparing the generalisation performance as a 
function of the compressed input dimension $m$ for Kriging and PCE models 
combined with KPCA with different kernels. 
Because of the availability of a validation set, the performance of the LOO 
error estimator in \eqref{eq:epsilon_LOO} is also assessed by comparing it with 
the true validation error in \eqref{eq:epsilong_gen_ideal}.
Figures~\ref{fig:res_sobol_drsm_m_vs_error-kg-loo} and 
\ref{fig:res_sobol_drsm_m_vs_error-pce-loo} show the LOO error estimator of the 
final surrogate model when using Kriging and PCE, respectively. 
In each panel the different curves correspond to different KPCA kernels, namely 
polynomial kernel (\eqref{eq:KPCA_kernel_poly}) and isotropic (resp. 
anisotropic) Gaussian (\eqref{eq:KPCA_kernel_gauss_aniso}).
Figures~\ref{fig:res_sobol_drsm_m_vs_error-kg-rmse} and 
\ref{fig:res_sobol_drsm_m_vs_error-pce-rmse} show the corresponding validation 
error on the validation set for the same scenarios.
At a first glance, it is clear that the top and bottom figures are remarkably 
similar, both in their trends and in absolute value. 
Therefore, it is concluded that on this example $\epsilon_{LOO}$ is a good 
measure of the generalisation error $\epsilon_{gen}$. 
This is an important observation, because in the general case a validation set 
is not available, while $\epsilon_{LOO}$ can always be calculated. 
Moreover, the intrinsic dimension identified by all the best DR-SM combinations 
is  equal to $\widehat{m} = 6$, which is a reasonable estimate based on the 
values of the constants $c_i$ in \eqref{eq:app_Sobol_constants}.

The DRSM algorithm identifies the anisotropic Gaussian kernel as the best KPCA 
kernel to be used in conjunction with both Kriging and PCE. 
However, the performance of PCE is significantly better in terms of 
generalisation error.
The optimal parameters for each case (Kriging and PCE) are highlighted by a 
black dot in \figref{fig:res_sobol_drsm_m_vs_error}, and their numerical values 
are reported in \tabref{tab:res_sobol_drsm_mstar}.
\begin{table}[]
	\centering
	\caption{Sobol' function: optimal DRSM configurations for Kriging- and 
	PCE-based surrogate models}
	\label{tab:res_sobol_drsm_mstar}
	\begin{tabular}{l c c c  r}
		\hline
		SM method & KPCA kernel & $\widehat{m}$ & $\varepsilon_{LOO}$ & 
		$\widehat{\varepsilon}_{gen}$ \\ \hline
		Kriging &
		Anisotropic Gaussian &
		$6$ &
		0.0704 &
		0.0830 \\
		PCE &
		Anisotropic Gaussian &
		$6$ &
		0.0096 &
		0.0083 \\
		\hline
	\end{tabular}
	
\end{table}

\begin{figure}[!ht]
	\centering
	\subfigure[Kriging]{\label{fig:res_sobol_drsm_vs_others-kg}
		\includegraphics[width=.47\textwidth,trim={0.2cm 0 2.0cm 
		0},clip]{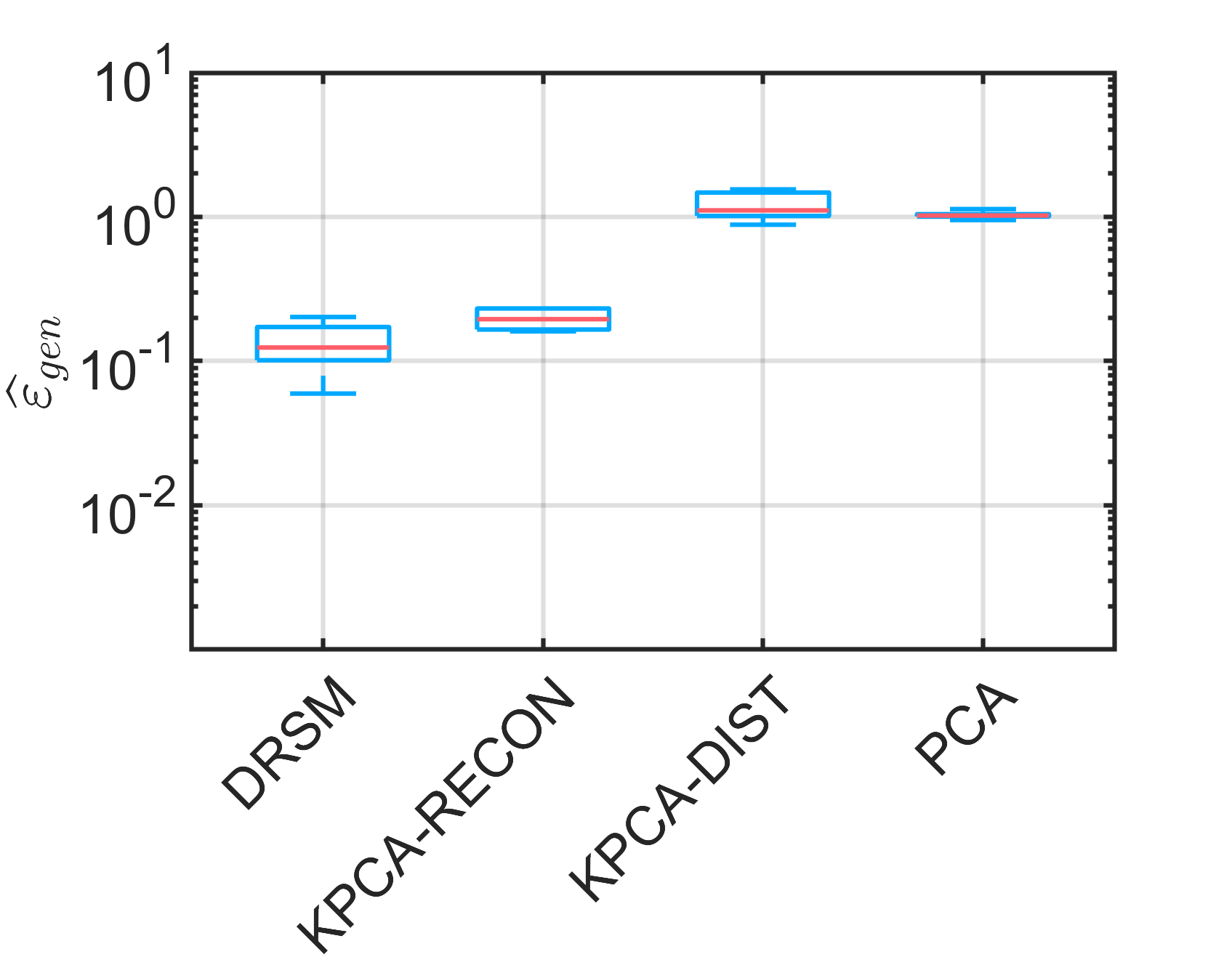}
	}
	\subfigure[Polynomial chaos 
	expansions]{\label{fig:res_sobol_drsm_vs_others-pce}
		\includegraphics[width=.47\textwidth,trim={0.2cm 0 2.0cm 
		0},clip]{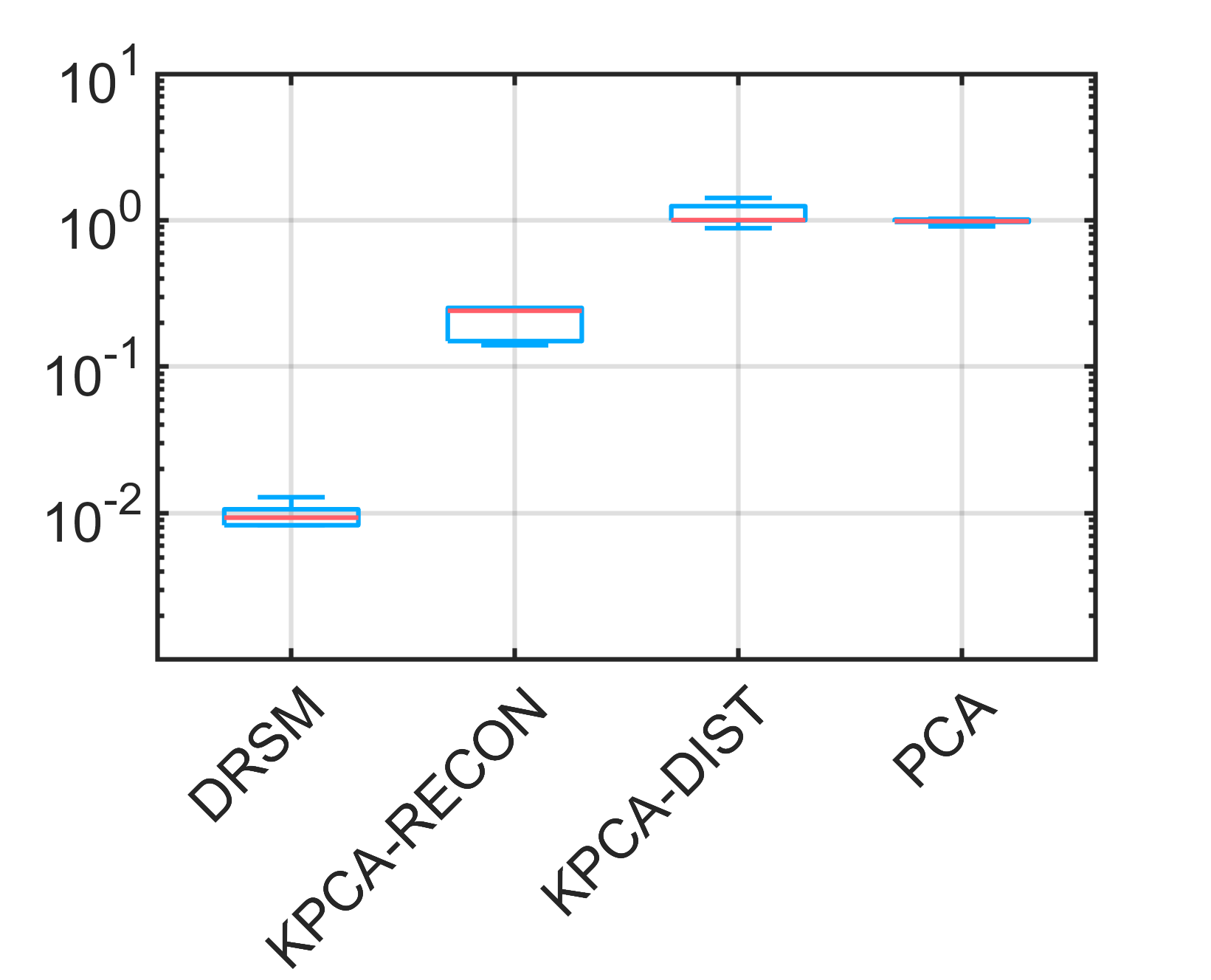}
	}
	\caption{Sobol' function: estimates of the generalisation error with 
		different DR approaches (see \tabref{tbl:validation_configs} for 
		their description).
		The reduced dimension identified by DRSM (see 		
		\tabref{tab:res_sobol_drsm_mstar}) is used for all cases.}
	\label{fig:res_sobol_drsm_vs_others}
\end{figure}

Subsequently, the performance of DRSM is compared against an unsupervised 
approach, in which dimensionality reduction is carried out first, before 
applying surrogate modelling. 
To facilitate a meaningful comparison between the various methods, the reduced 
dimension and the optimal KPCA kernel as determined by the first analysis (see 
\tabref{tab:res_sobol_drsm_mstar}) is used.
The results are summarised in \figref{fig:res_sobol_drsm_vs_others}, while the 
corresponding list of tested configurations for both DRSM and the sequential 
DR-SM is given in \tabref{tbl:validation_configs}.

\begin{table}
	\centering
	\caption{Different setups considered for evaluating the final surrogate 
	model performance after using each of them for dimensionality reduction.}
	\label{tbl:validation_configs}
	\begin{tabularx}{\textwidth}{l X r}
		\hline
		\textbf{Dim. reduction} & \textbf{Parameter tuning objective} & 
		\textbf{Abbreviation} \\
		\hline
		Kernel PCA & $\varepsilon_{LOO}$ of Kriging (KG) or PCE surrogate 
		(\eqref{eq:DRSM_outer_loop}) & DRSM \\
		Kernel PCA & Reconstruction error (\eqref{eq:KPCA_Jrecon}) & KPCA-RECON 
		\\
		Kernel PCA & Pairwise distance preservation (\eqref{eq:KPCA_Jdist})& 
		KPCA-DIST\\
		PCA & - & PCA\\
		\hline
	\end{tabularx}
	
\end{table}

The experimental design consists of $800$ samples. 
The performance of each method is evaluated in terms of the generalisation 
error of the final surrogate $\widehat{\cm}(\bfz)$ evaluated on a validation 
set $\acc{\cx_v, \bfy_v=\cm(\cx_v)}$ with $10^5$ samples. 
To evaluate the robustness of the results, this process is repeated $10$ times, 
each corresponding to a different set $\cx$, drawn at random using the Latin 
Hypercube sampling method. 
On the left (resp. right) panel, a Kriging (resp. PCE) surrogate is calculated 
using one of the methods in \tabref{tbl:validation_configs}.
Each box plot in \figref{fig:res_sobol_drsm_vs_others} provides summary 
statistics of the generalisation error that was achieved by each configuration 
over the $10$ repetitions. 
The central mark indicates the median, and the bottom and top edges of the box 
indicate the $25^{\text{th}}$ and $75^{\text{th}}$ percentiles, respectively. 
The whiskers extend to the most extreme data points up to $1.5$ times the 
inter-quartile range above or below the box edges. Any sample beyond that range 
is considered an outlier and plotted as a single point.

The DRSM approach consistently shows superior performance compared to the 
unsupervised approaches. 
This performance improvement becomes more apparent in the case of PCE surrogate 
modelling, where the average validation error over the $10$ repetitions  is 
reduced by almost two orders of magnitude compared to the other methods. 
Further investigation on the significant performance difference between 
Kriging- and PCE- based DRSM shows that it 
stems from a combination of two different factors. On the one hand PCE models 
tend to achieve a lower generalization
error than the corresponding Kriging models, regardless on the specific KPCA 
parameters on this particular function. 
On the other hand, the choice of a Kriging proxy surrogate with an isotropic 
covariance function causes the optimization in \eqref{eq:DRSM_inner_loop} to 
identify a sub-optimal configuration.

\begin{figure}[!t]
	\centering
	% trim={0.05cm 0 0.1cm 0},clip
	\subfigure[Total Sobol' indices]{\label{fig:res_sobol_drsm_features_sobol} 
		\hspace{0.8cm}			\includegraphics[height=6.5cm,,trim={0cm 0 
		1.6cm 0},clip]{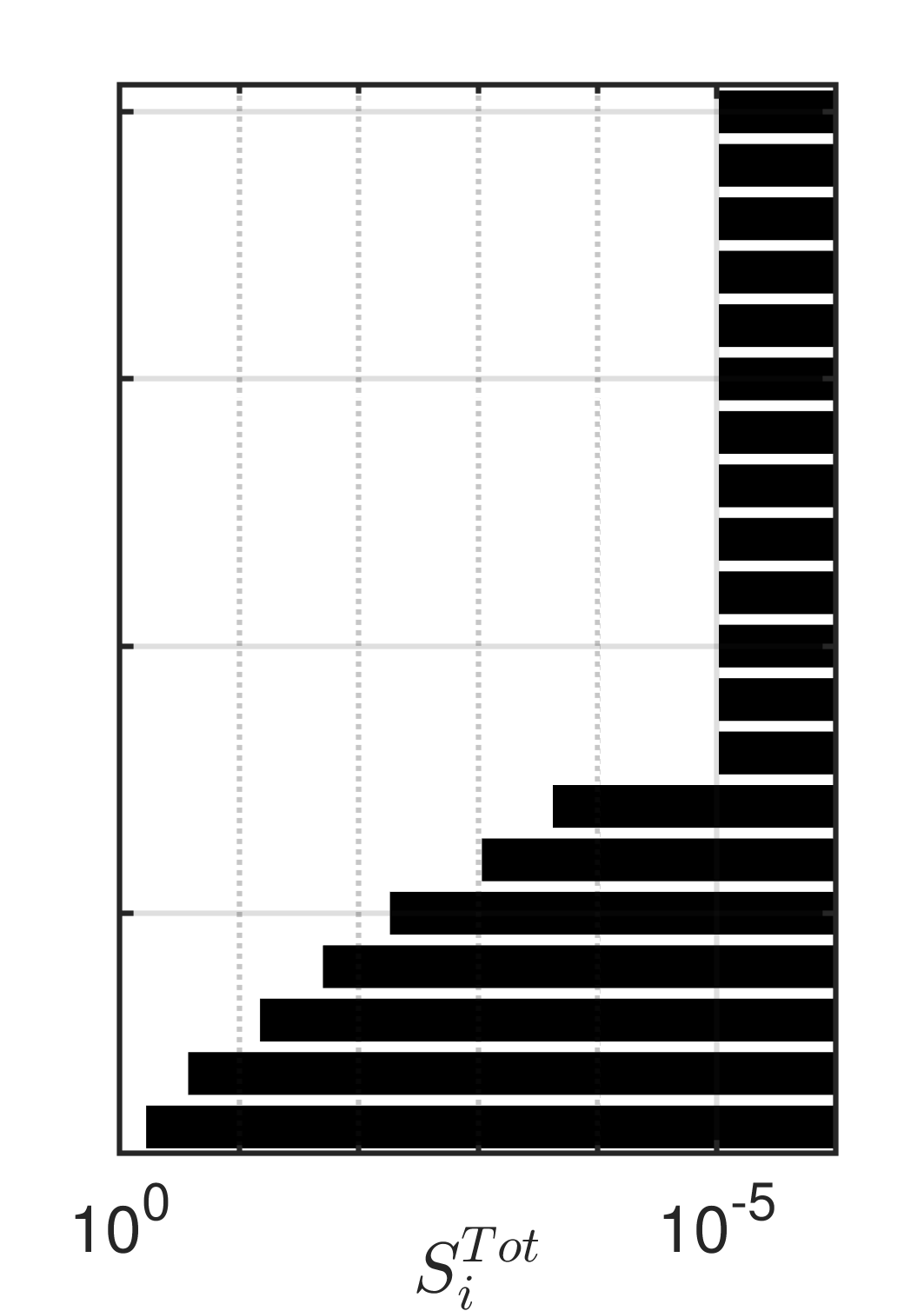}
	}
	\subfigure[{$m=3$}]{\label{fig:res_sobol_drsm_features_corr-1}
		\includegraphics[height=6.5cm,trim={1.4cm 0 8.0cm 
		0},clip]{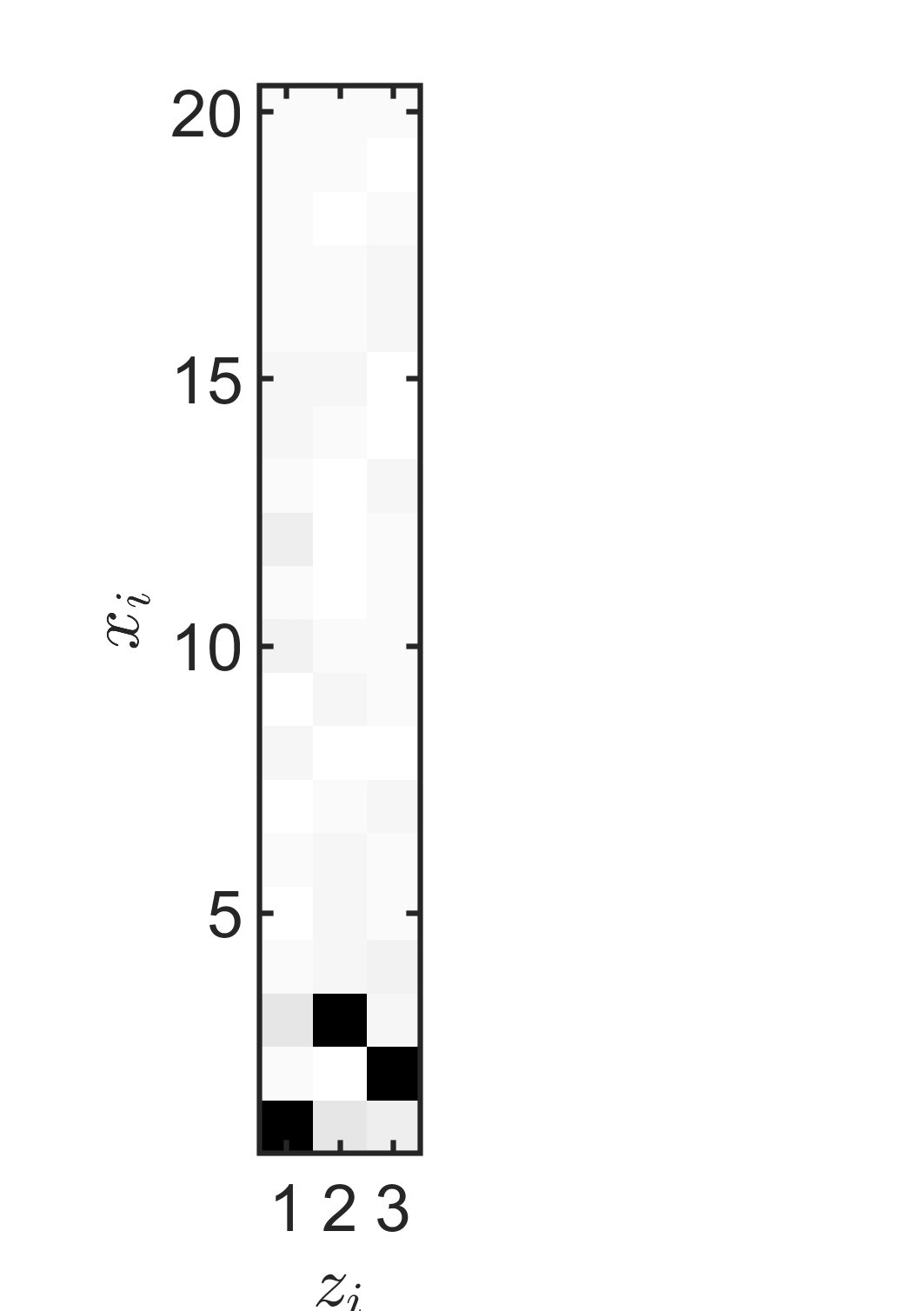}
	}
	\subfigure[{$m=4$}]{\label{fig:res_sobol_drsm_features_corr-2}
		\includegraphics[height=6.5cm,trim={1.4cm 0 8.0cm 
		0},clip]{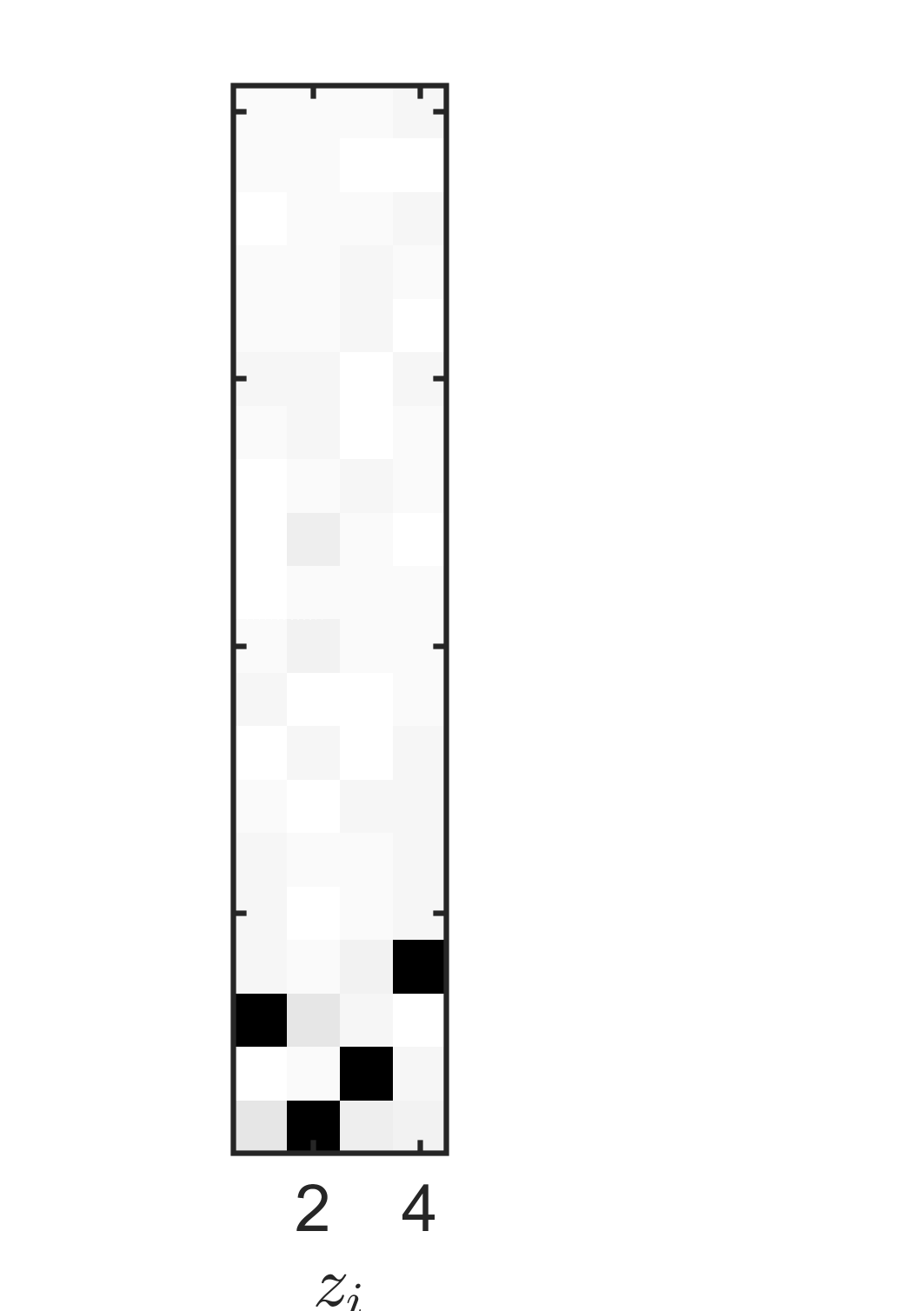}
	}
	\subfigure[{$m=5$}]{\label{fig:res_sobol_drsm_features_corr-3}
		\includegraphics[height=6.5cm,trim={1.4cm 0 8.0cm 
		0},clip]{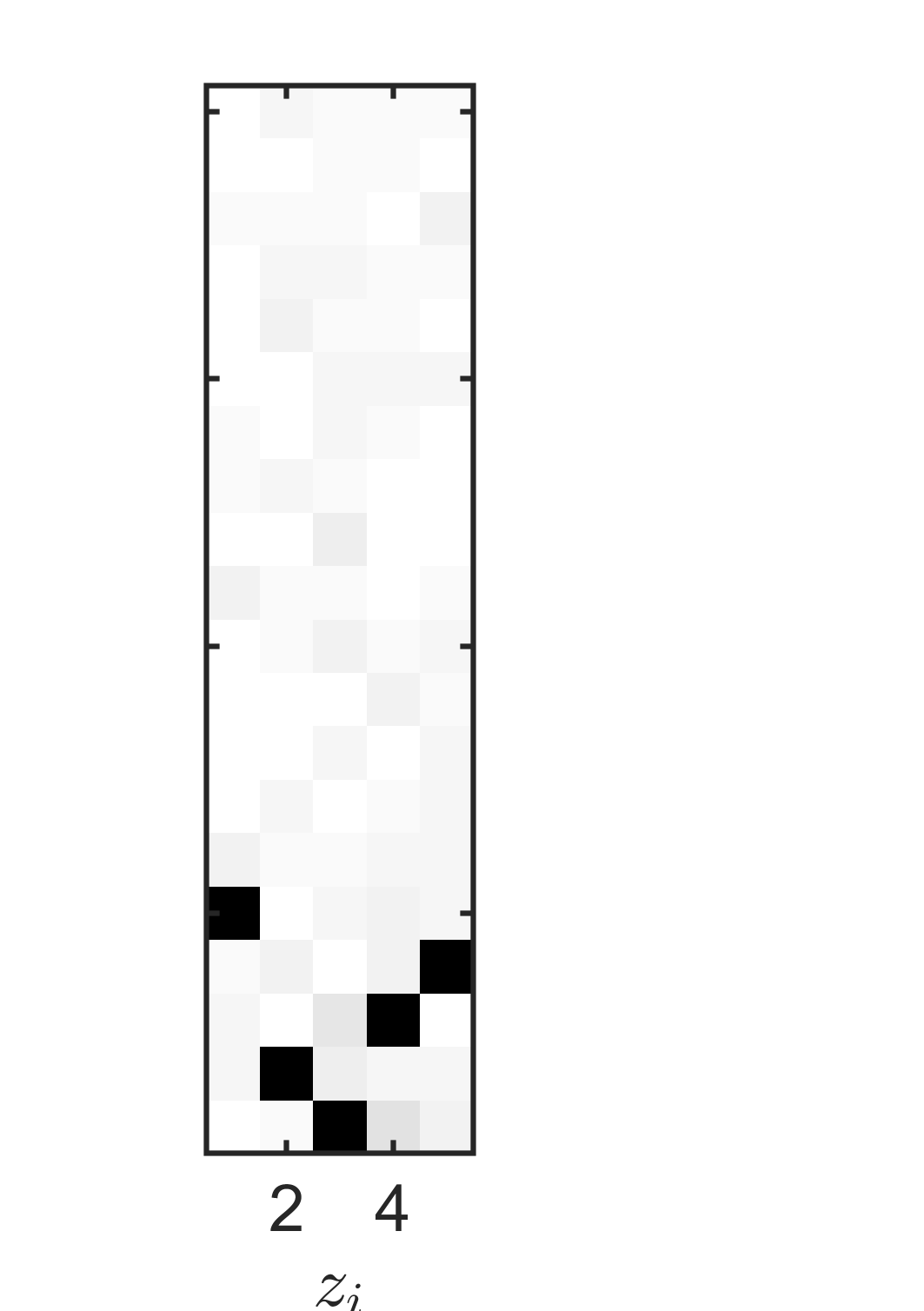}
	}
	\subfigure[{$m=6$}]{\label{fig:res_sobol_drsm_features_corr-4}
		\includegraphics[height=6.5cm,trim={1.4cm 0 5.0cm 
		0},clip]{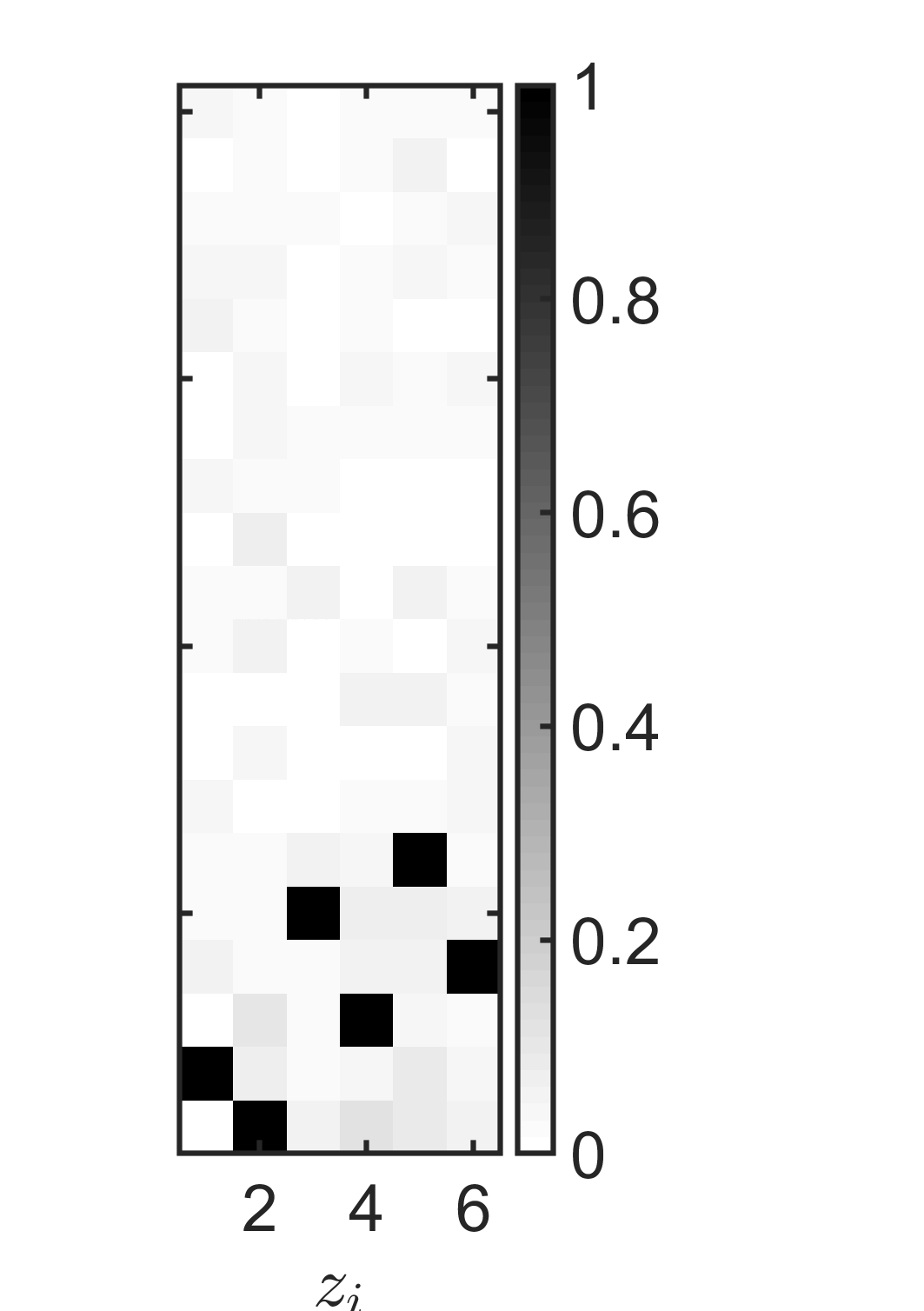}
	}

	%	\caption{Sobol' function: visualisation of the sample-based Spearman 
	%correlation coefficient (absolute value) between the model inputs $\bfX$ 
	%and the reduced space inputs $\bfZ$. }
	\caption{Sobol' function: total Sobol' indices (in logarithmic scale) and 
	visualisation of the sample-based Spearman correlation coefficient 
	(absolute value) between the model inputs $\bfX$ and the reduced space 
	inputs $\bfZ$ for varying reduced space dimension. }
	\label{fig:res_sobol_drsm_features_corr}
\end{figure}

Due to the analytical nature of the model under consideration, we further 
evaluate the DRSM-based input compression by means of how the most important 
input variables are mapped to the reduced space. 
We adopt the total Sobol' sensitivity indices as a rigorous measure of the 
importance of each input variable. 
Sobol' sensitivity analysis is a form of global sensitivity analysis based on 
decomposing the variance of the model output into contributions that can be 
directly attributed to inputs or sets of inputs \citep{Sobol1993}. 
The total Sobol' sensitivity index  of an input variable $X_i$, denoted by 
$S_i^{Tot} \in [0,1]$, quantifies the total effect of $X_i$ on the variance of 
$Y$. 
In this particular example, the total Sobol' indices can be analytically 
derived \citep{Saltelli2000}. 
Their values are shown for reference in 
\figref{fig:res_sobol_drsm_features_sobol}.

It is clear from \eqref{eq:app_Sobol} and \eqref{eq:app_Sobol_constants} that 
all $20$ input variables contribute to the output variability, \ie{} the 
intrinsic dimension of the problem is $20$. 
However, the contribution of each input component quickly diminishes with 
larger values of $c_i$ (see \figref{fig:res_sobol_drsm_features_sobol} in which 
the values of the $20$ total Sobol' indices are plotted, in logarithmic scale, 
as horizontal bars). 
Compressing the inputs in this problem is expected to lead to a mapping where 
those first few input components have the largest contribution.

In \figref{fig:res_sobol_drsm_features_corr} the features in the reduced space 
$\bfZ$ are compared against the original inputs $\bfX$. 
The rationale behind this heuristic analysis is simple: if the features 
obtained by DRSM are correctly identified, they should depend mostly on the 
same variables identified as important in the Sobol' analysis in 
\figref{fig:res_sobol_drsm_features_sobol}. 
A simple measure of dependence between the reduced space components $\acc{z_i 
\, , \, i = 1 \enu m}$ and the initial input space components $\acc{x_i \, , \, 
i = 1 \enu M}$ is provided by the metric $\abs{\rho\prt{z_i,x_i}}$, where 
$\rho$ denotes the Spearman correlation coefficient. 
Figures \ref{fig:res_sobol_drsm_features_corr-1} - 
\ref{fig:res_sobol_drsm_features_corr-4}  represent graphically the quantity 
$\abs{\rho\prt{z_i,x_i}}$ for the best surrogate identified in 
\tabref{tab:res_sobol_drsm_mstar}, namely a PCE coupled with KPCA using an 
anisotropic Gaussian kernel, evaluated on the validation set 
$\acc{\cx_v,\bfy_v}$. 
Each figure corresponds to a different selection of reduced space dimension $m$.
\figref{fig:res_sobol_drsm_features_corr} clearly shows that (i) each $z_i$ 
correlates strongly with a specific $x_i$, (ii) the $z_i$'s correlate with the 
$m$ ``most important'' $x_i$'s, and, (iii) the larger $m$ value leads to the 
discovery of a new input $z_i$ that correlates with the next ``most important'' 
component of $\bfx$.

% Add the N/m investigation here 

\begin{figure}[!ht]
	\centering
	\subfigure[Kriging]{\label{fig:res_sobol_Nm_vs_epsilon-kg}
		\includegraphics[width=.47\textwidth,trim={0.2cm 0 0.5cm 
		0},clip]{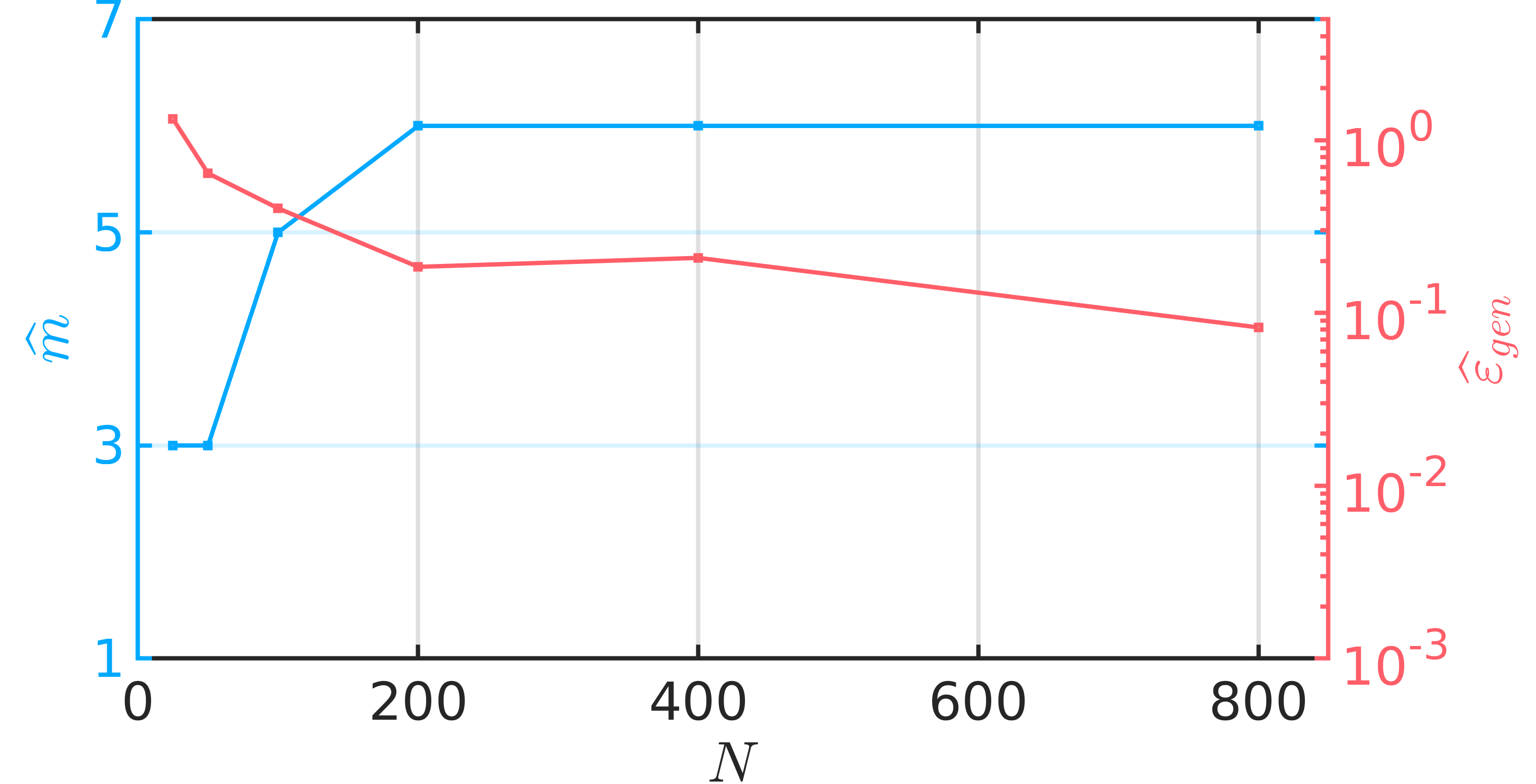}
	}
	\subfigure[Polynomial chaos 
	expansions]{\label{fig:res_sobol_Nm_vs_epsilon-pce}
		\includegraphics[width=.47\textwidth,trim={0.2cm 0 0.5cm 
		0},clip]{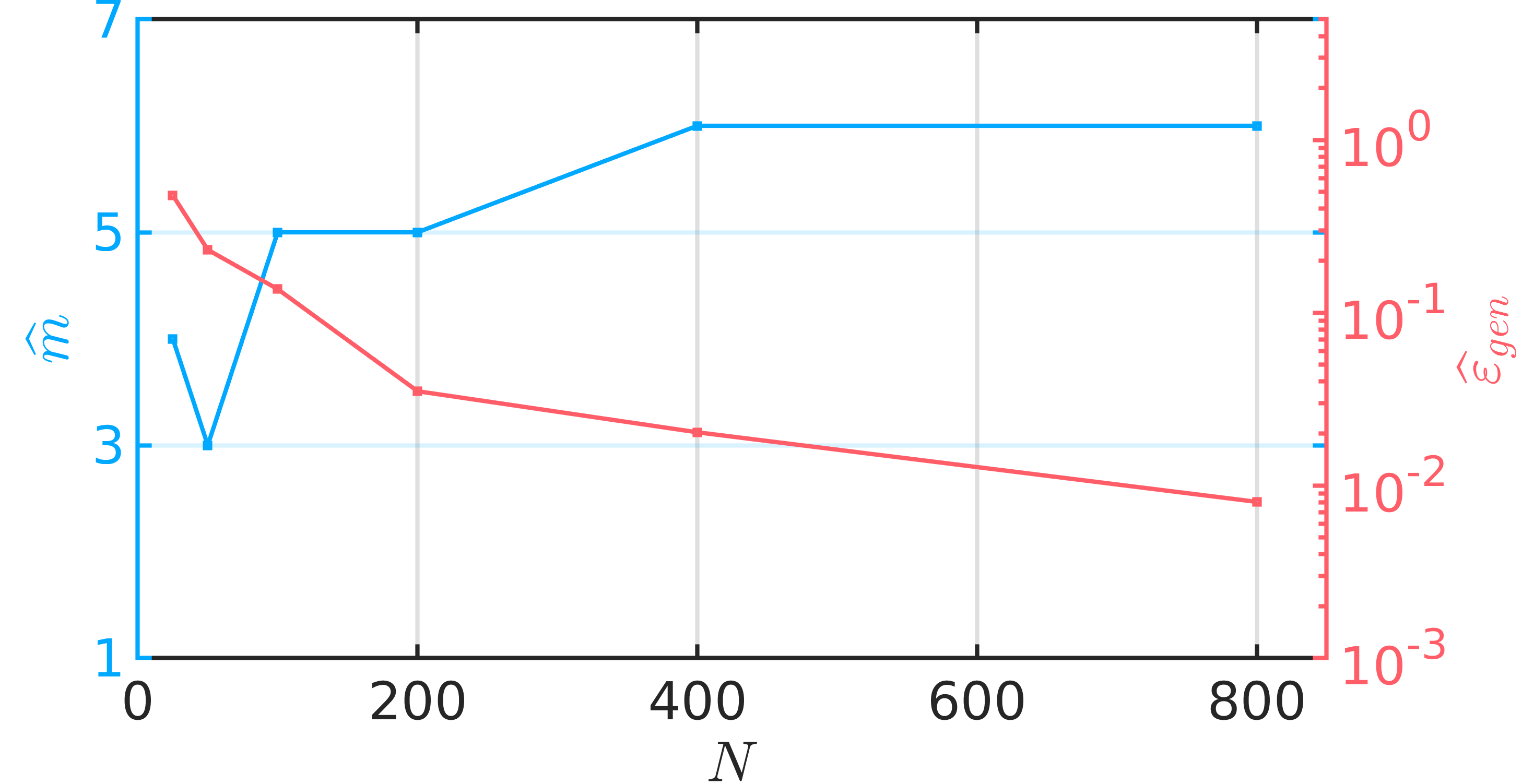}
	}
	\caption{Sobol' function: Optimal reduced space dimension (identified by 
	DRSM) and generalisation error of the surrogate using an experimental 
	design of varying size.}
	\label{fig:res_sobol_Nm_vs_epsilon}
\end{figure}

Finally, in \figref{fig:res_sobol_Nm_vs_epsilon} we investigate the convergence 
behaviour of the DRSM algorithm, with respect to the experimental design size 
$N$. 
Using the same validation set as before, we repeat the DRSM training process 
using a varying number of samples in the experimental design. 
\figref{fig:res_sobol_Nm_vs_epsilon-kg} (resp. 
\figref{fig:res_sobol_Nm_vs_epsilon-pce}) shows the resulting generalisation 
performance of the Kriging (resp. polynomial chaos expansion) surrogate and the 
optimal reduced space dimension $\widehat{m}$ in each case. 
At a first glance, the convergence behaviour is similar when a Kriging or a PCE 
surrogate is used, but the latter shows consistently better performance 
regardless of the number of experimental design samples available. 
%The optimal reduced space dimension increases with the number of training 
%samples, up to $\widehat{m} = 6$, with some fluctuation in the low sample 
%region in \figref{fig:res_sobol_Nm_vs_epsilon-pce}. 
As expected, the identified optimal reduced dimension increases with the number 
of available experimental design points, plateauing in both cases to 
$\widehat{m} = 6$.
This is consistent with the notion that identifying the inherent dimensionality 
of a system requires a sufficient amount of information, which in this case is 
directly related to the available number of samples. 

%%%%%
\subsection{Electrical resistor network} \label{sec:App:resistor_networks}
%%%%%

\begin{figure}[!ht] 
	\centering
	\includegraphics[width=.75\textwidth]{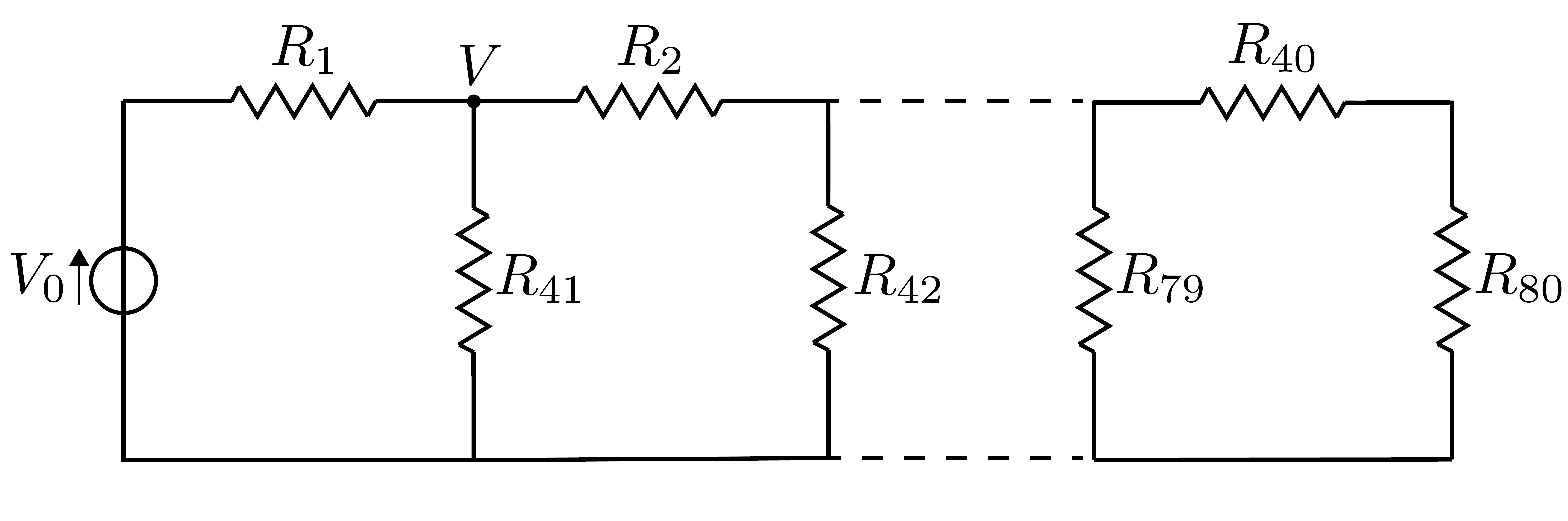}
	\caption{The resistor networks application example}
	\label{fig:app_resnets}
\end{figure}

The electrical resistor network in \figref{fig:app_resnets} \citep{Jakeman2015} 
is considered next. 
It contains $80$ resistances of uncertain ohmage (model inputs) and it is 
driven by a voltage source providing a known potential $V_0$. 
The output of interest is the voltage $V$ at the node shown in  
\figref{fig:app_resnets}. A single set of $1,000$ experimental design samples 
and model responses is available, courtesy of J. Jakeman.

As in the previous section, the goal of the first analysis is to determine the 
generalisation performance of the DRSM surrogate as a function of the reduced 
space dimension $m$ when KPCA is combined with either Kriging or PCE. 
In addition, the accuracy of the LOO error in \eqref{eq:epsilon_LOO} is 
compared to the validation error in \eqref{eq:epsilong_gen_estim}. 
The samples are randomly split into $500$ pairs $\acc{\cx, \bfy}$ used during 
the DRSM calibration and $500$ pairs $\acc{\cx_v, \bfy_v}$ used for validation. 

\begin{figure}[!t]
	\centering
	\subfigure[Kriging - LOO 
	error]{\label{fig:res_resnets_drsm_m_vs_rmse-kg-loo}
		\includegraphics[width=.47\textwidth,trim={0.2cm 0 2.0cm 
		0},clip]{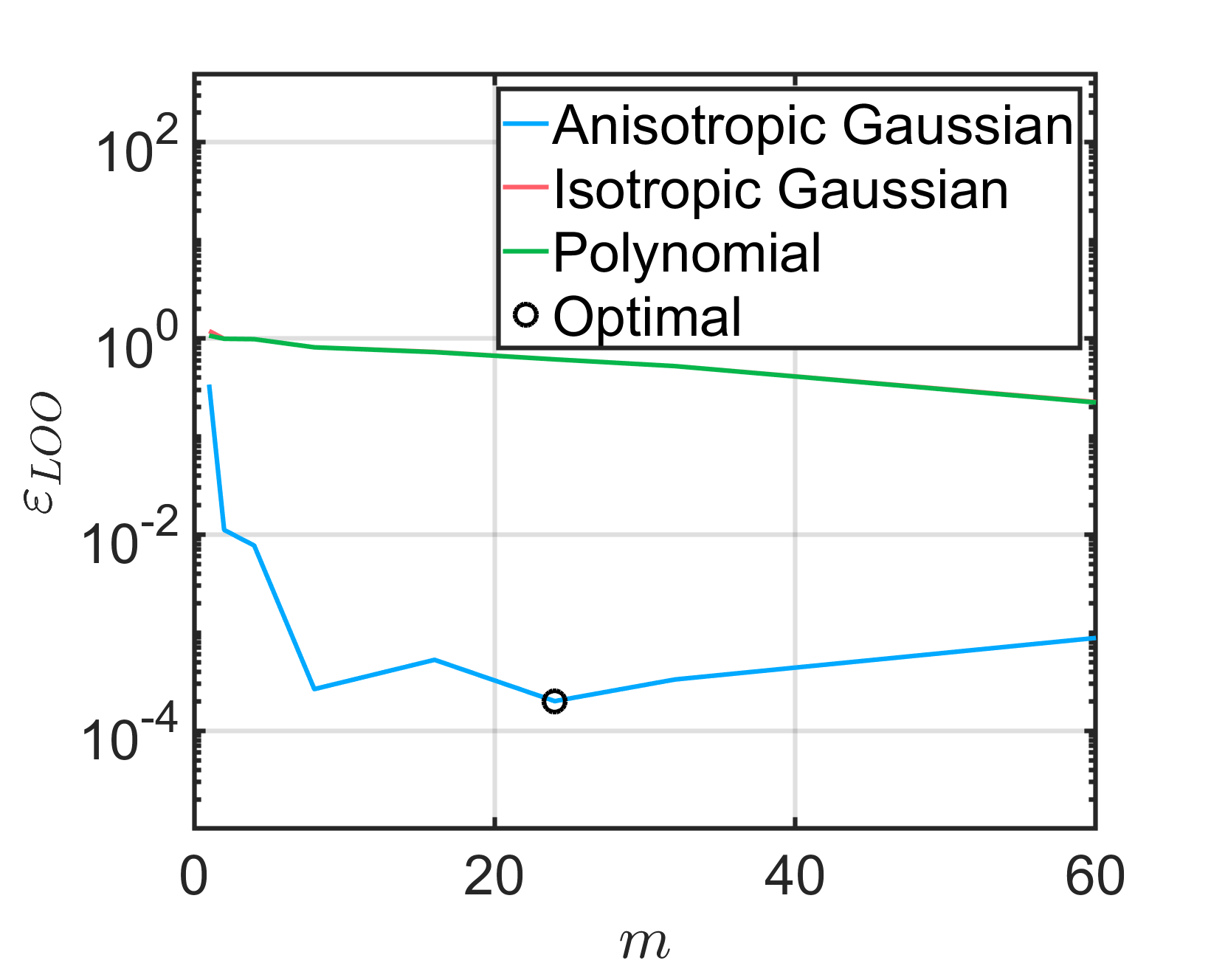}
	}
	\subfigure[PCE - LOO error]{\label{fig:res_resnets_drsm_m_vs_rmse-pce-loo}
		\includegraphics[width=.47\textwidth,trim={0.2cm 0 2.0cm 
		0},clip]{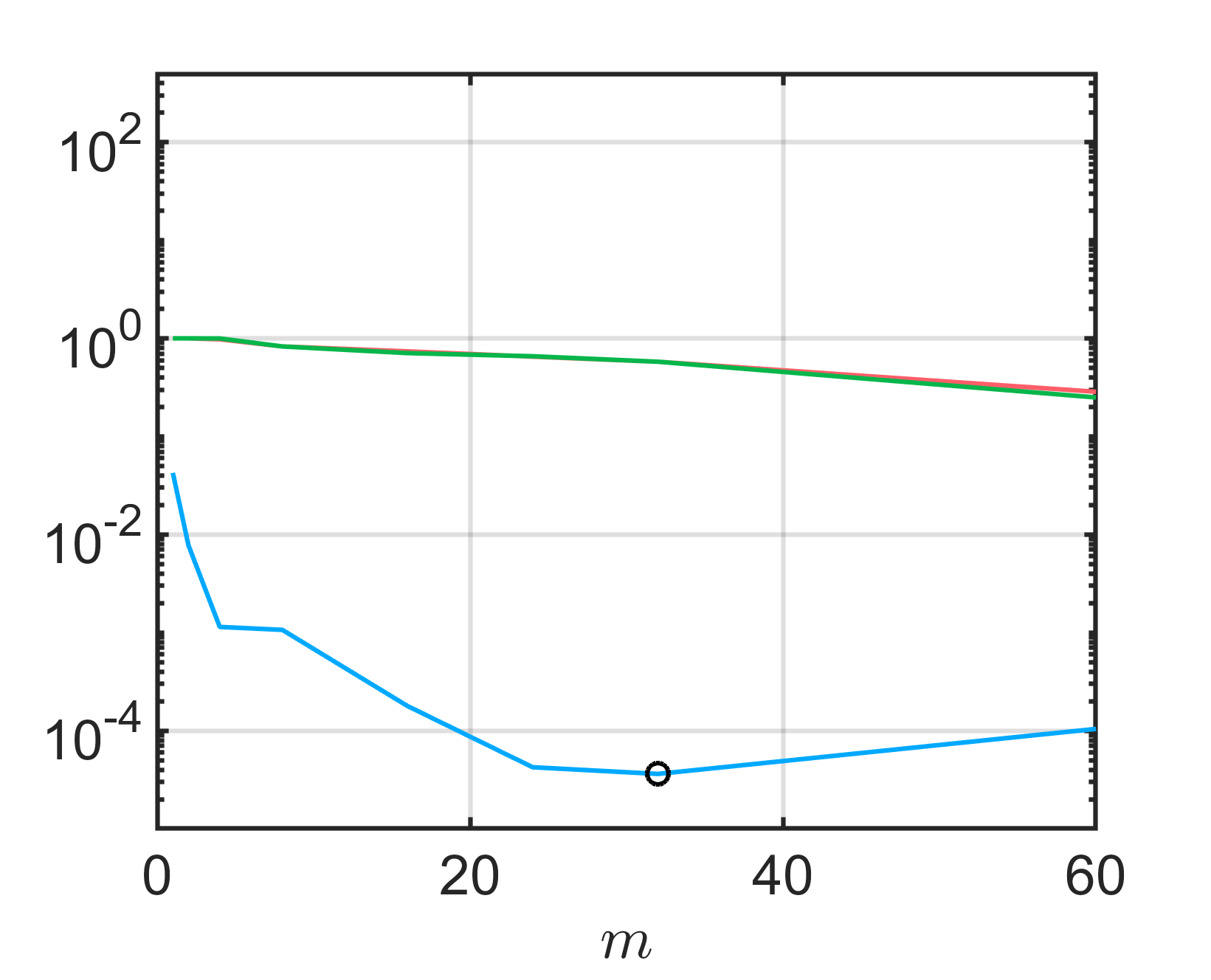}
	}
	
	\subfigure[Kriging - Validation 
	error]{\label{fig:res_resnets_drsm_m_vs_rmse-kg-rmse}
		\includegraphics[width=.47\textwidth,trim={0.2cm 0 2.0cm 
		0},clip]{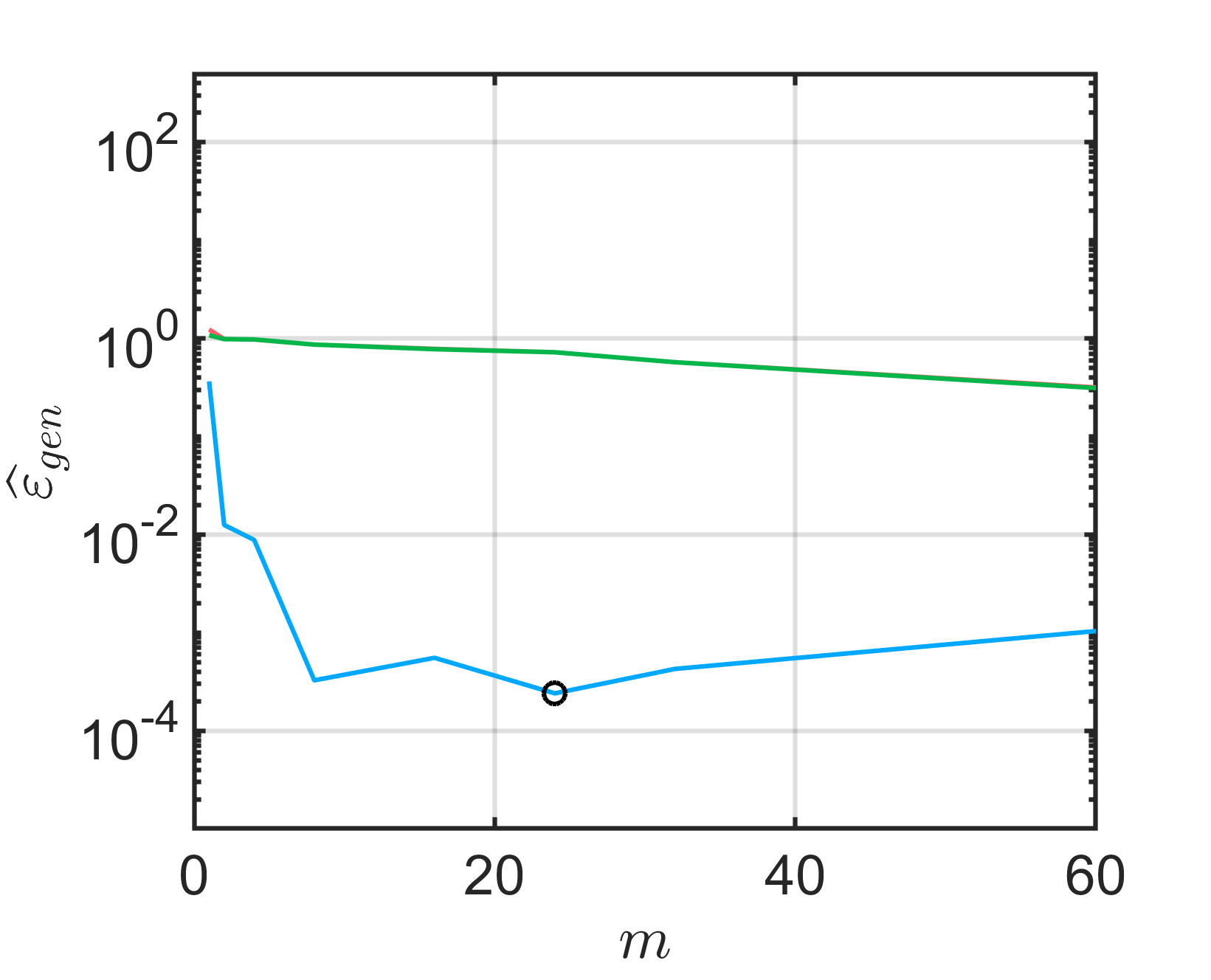}
	}
	\subfigure[PCE - Validation 
	error]{\label{fig:res_resnets_drsm_m_vs_rmse-pce-rmse}
		\includegraphics[width=.47\textwidth,trim={0.2cm 0 2.0cm 
		0},clip]{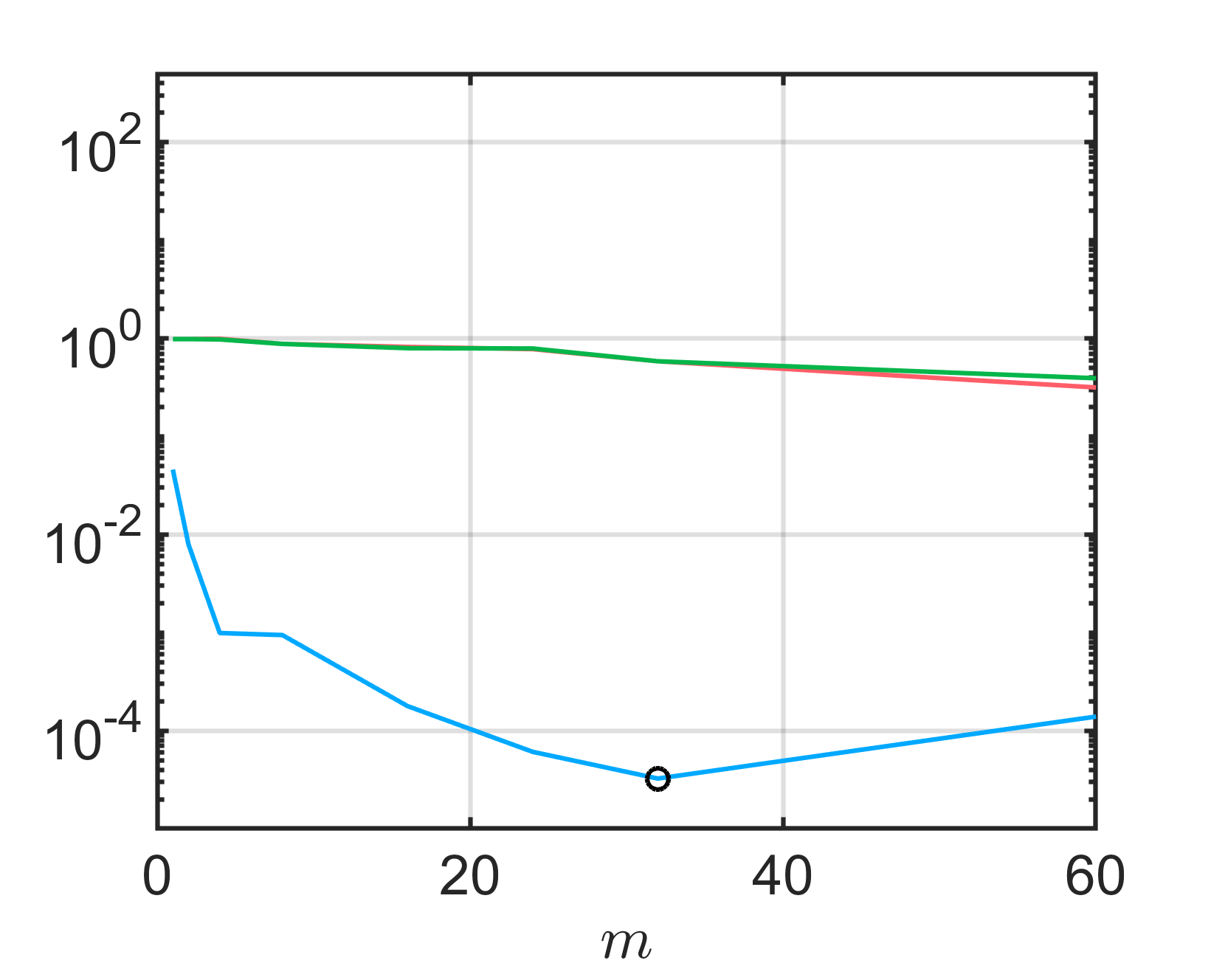}
	}
	
	\caption{Electrical resistor networks: 
		error estimates of the DRSM surrogate as a function of the reduced 
		dimension. Kernel PCA is used with anisotropic (resp. isotropic) 
		Gaussian as well as polynomial kernels. }
	\label{fig:res_resnets_drsm_m_vs_rmse}
\end{figure}

Figures~\ref{fig:res_resnets_drsm_m_vs_rmse-kg-loo} and  
\ref{fig:res_resnets_drsm_m_vs_rmse-pce-loo} show the LOO error estimator of 
the final surrogate model  (Kriging or PCE), evaluated on $\acc{\cx,\bfy}$, 
whereas
Figures~\ref{fig:res_resnets_drsm_m_vs_rmse-kg-rmse} and  
\ref{fig:res_resnets_drsm_m_vs_rmse-pce-rmse} show the validation error of the 
surrogate, evaluated on $\acc{\cx_v,\bfy_v}$.
In each panel, each curve corresponds to a different KPCA kernel, namely 
anisotropic or isotropic Gaussian, and polynomial. 
Finally, the optimal configuration for each SM method is illustrated by a black 
dot. 
Similarly to the Sobol' function, the use of an anisotropic kernel in KPCA 
results in significantly reduced generalisation error. 
Indeed this is expected from a physical standpoint. 
The effect of the resistors on the voltage $V$ will decay with distance (in 
terms of the number of preceding resistors) from $V$, which implies anisotropy 
in terms of the effect of each input variable to the output. 
As in the previous application example, the LOO error in 
Figures~\ref{fig:res_resnets_drsm_m_vs_rmse-kg-loo} and  
\ref{fig:res_resnets_drsm_m_vs_rmse-pce-loo} provides a reliable proxy of the 
generalisation error in Figures~\ref{fig:res_resnets_drsm_m_vs_rmse-kg-rmse} 
and  \ref{fig:res_resnets_drsm_m_vs_rmse-pce-rmse} and the same optimal 
parameters are identified w.r.t. the two error measures.
The optimal DRSM configuration for each surrogate model is given in 
\tabref{tab:res_resnets_drsm_mstar}. 
\begin{table}[!ht]
	\centering
	\caption{Resistor networks: optimal DRSM configurations for Kriging and PCE 
	surrogate models}
	\label{tab:res_resnets_drsm_mstar}
	\begin{tabular}{l c c c  r}
		\hline
		SM method & KPCA kernel & $\widehat{m}$ & $\varepsilon_{LOO}$ & 
		$\widehat{\varepsilon}_{gen}$ \\ \hline
		Kriging &
		Anisotropic Gaussian &
		$24$ &
		2.000e-04 &
		2.402e-04 \\
		PCE &
		Anisotropic Gaussian &
		$32$ &
		3.621e-05 &
		3.249e-05 \\
		\hline
	\end{tabular}
	
\end{table}

\begin{figure}
	\centering
	\subfigure[Kriging]{\label{fig:res_resnets_drsm_vs_others-kg}
		\includegraphics[width=.47\textwidth,trim={0.2cm 0 2.0cm 
		0},clip]{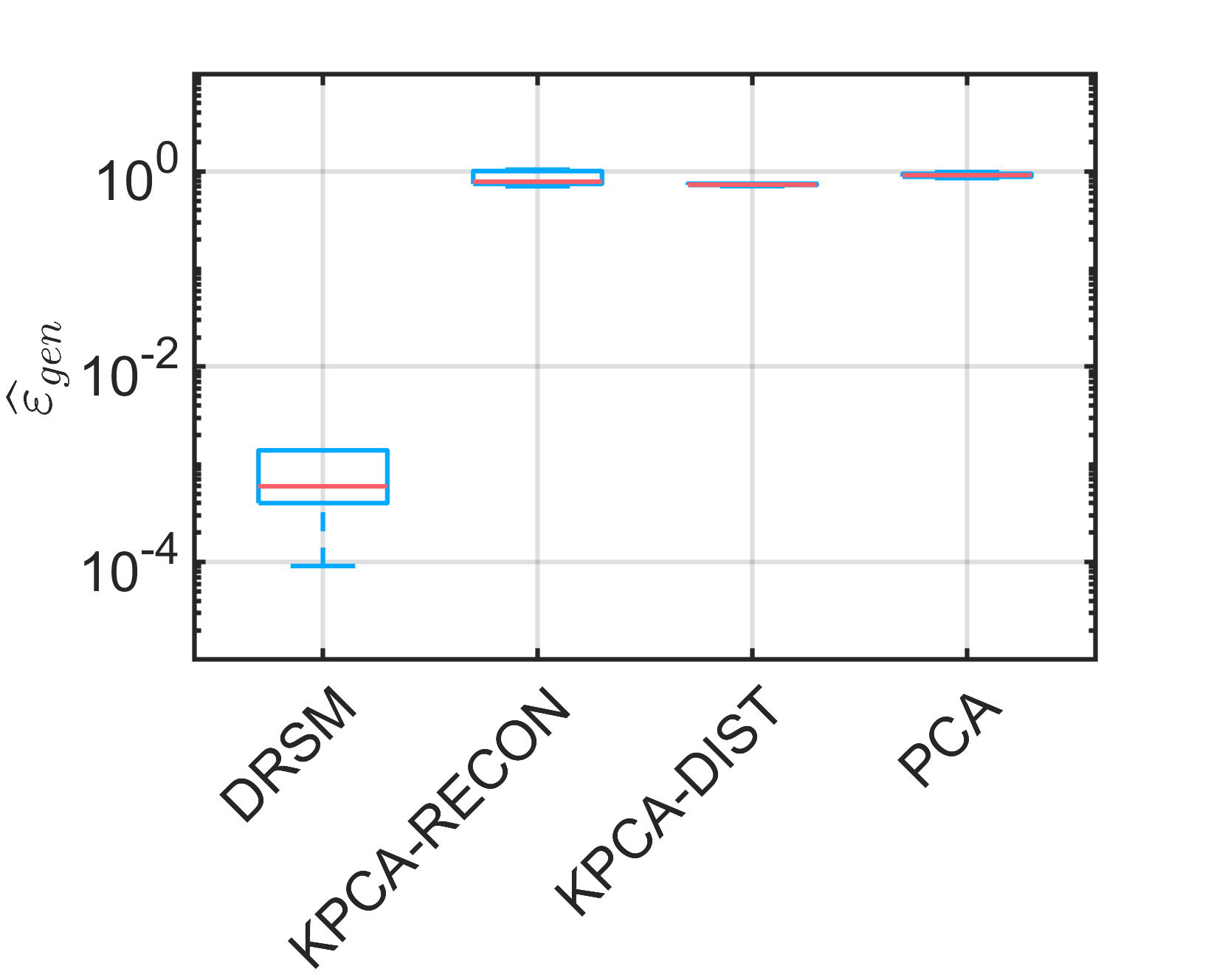}
	}
	\subfigure[Polynomial chaos 
	expansions]{\label{fig:res_resnets_drsm_vs_others-pce}
		\includegraphics[width=.47\textwidth,trim={0.2cm 0 2.0cm 
		0},clip]{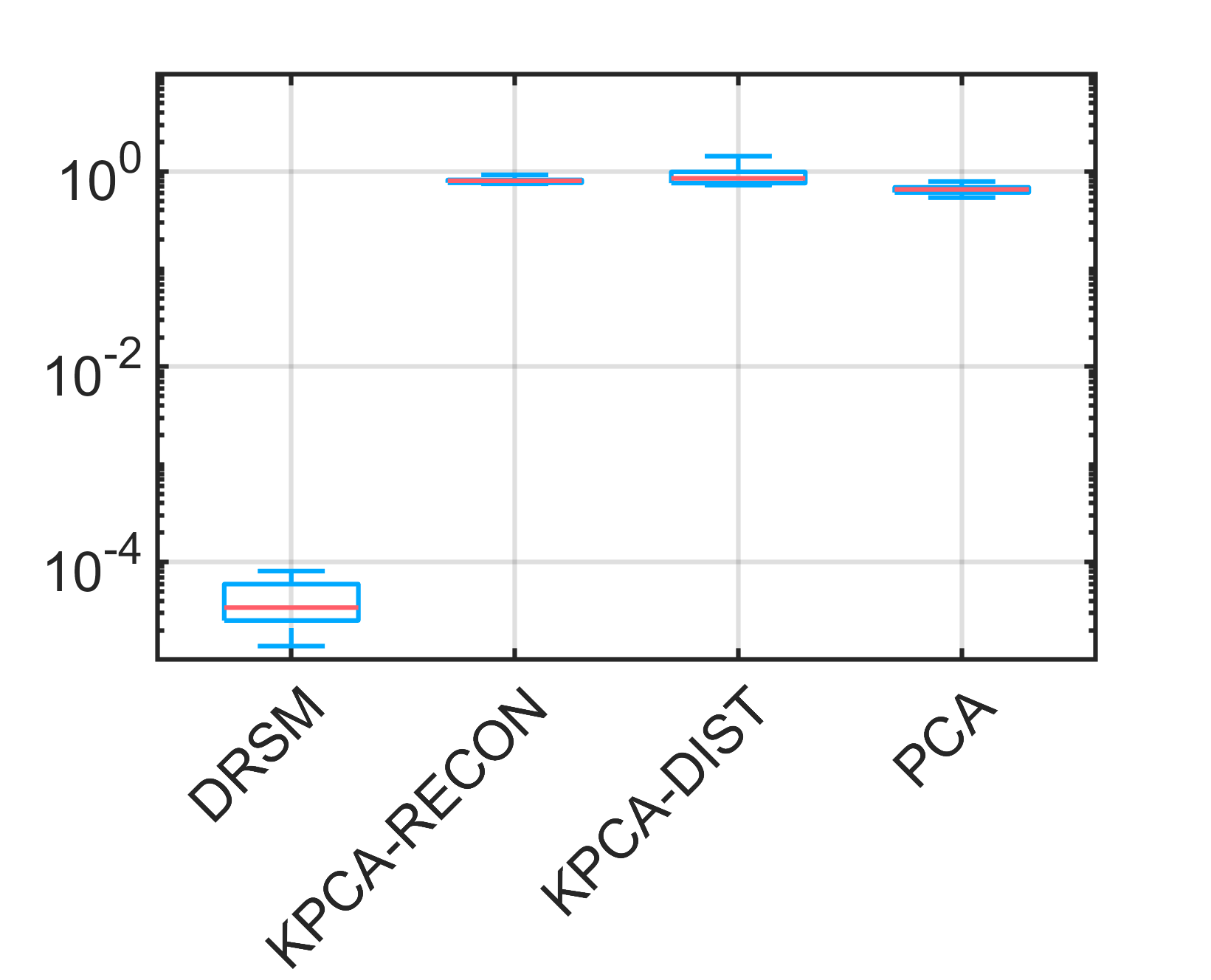}
	}
	
	\caption{Electrical resistor networks: estimates of the generalisation 
		error with different DR approaches (see \tabref{tbl:validation_configs} 
		for 
		their description). The optimal dimensionality identified by DRSM (see 
		\tabref{tab:res_resnets_drsm_mstar}) is used for all configurations.} 
	\label{fig:res_resnets_drsm_vs_others}
\end{figure}

Next, the performance of DRSM is compared to unsupervised approaches 
considering the setups in \tabref{tbl:validation_configs}. 
The results of this comparative study are given in 
\figref{fig:res_resnets_drsm_vs_others} using box plots. 
They are obtained by the repeated random selection of $500$ samples from the 
available $1,000$, leading to $10$ separate surrogate models for each case. 
The performance of each method is determined by means of the 
$\widehat{\varepsilon}_{gen}$ of the final surrogate $\widehat{\cm}(\bfz)$ 
evaluated on the validation set $\acc{\cx_v, \bfy_v=\cm(\cx_v)}$, that 
corresponds to the remaining $500$ samples of each split. 
Hence, each box-plot provides summary statistics of the validation error over 
the different splits. 
Each of the setups is tested both for Kriging 
(\figref{fig:res_resnets_drsm_vs_others-kg}) and PCE surrogates 
(\figref{fig:res_resnets_drsm_vs_others-pce}). 
In this application example the DRSM-based surrogates outperform the others by 
several orders of magnitude in both cases (Kriging, PCE). 
This highlights the difference between the unsupervised and supervised 
compression: compressing the input using only the information in $\cx$ appears 
inefficient when followed by surrogate modelling.  

\begin{figure}[!t]
	\centering
	% trim={0.05cm 0 0.1cm 0},clip
	\subfigure[SRC coefficients]{\label{fig:res_resnets_drsm_features_src} 
		\hspace{0.8cm}	\includegraphics[height=6.4cm,,trim={0cm 0 0.2cm 
		0},clip]{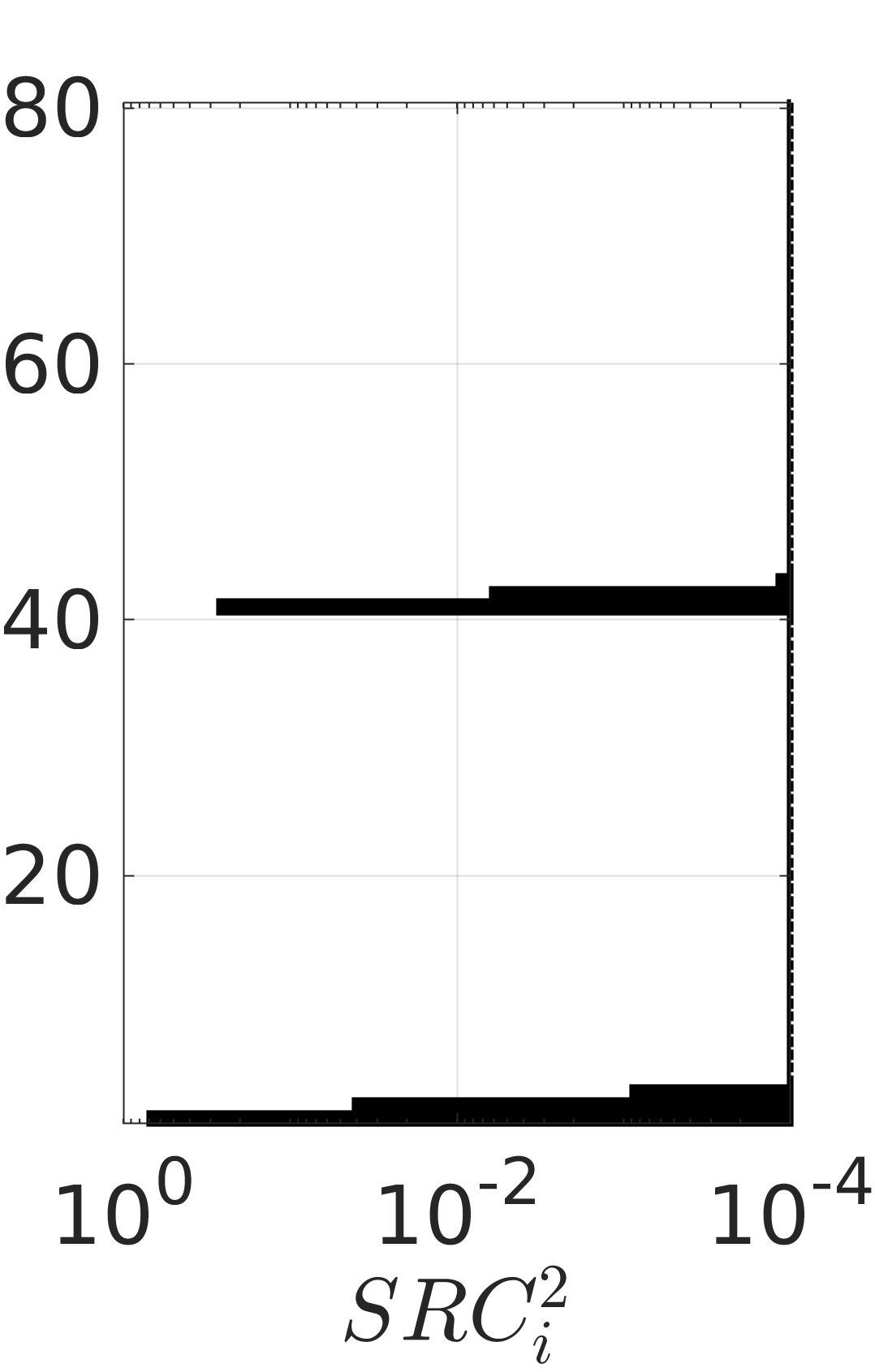}
	}
	\subfigure[$m=2$]{\label{fig:res_resnets_drsm_features_corr-1}
		\includegraphics[height=6.4cm,trim={1.4cm 0 9.7cm 
		0},clip]{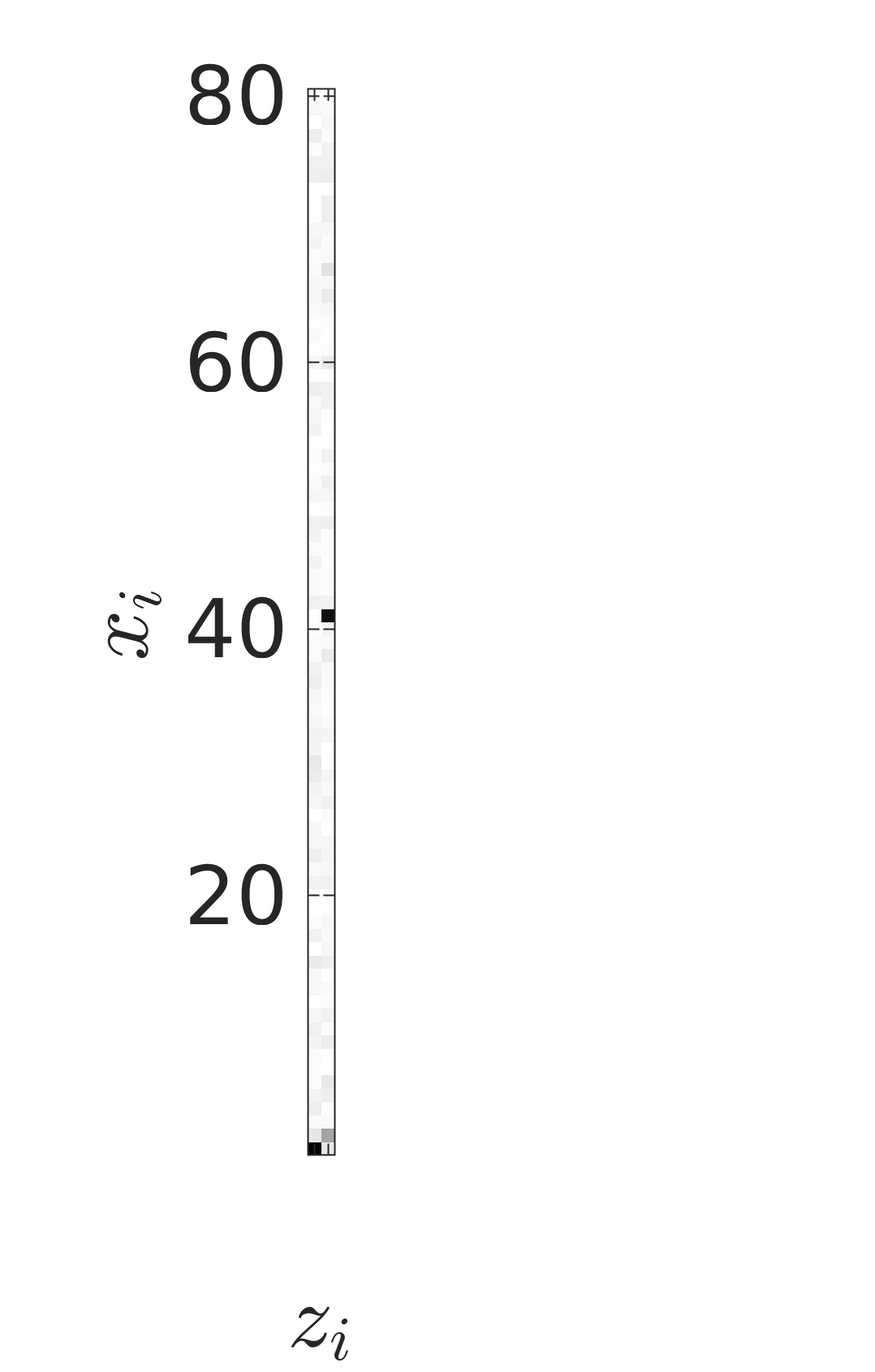}
	}
	\subfigure[$m=4$]{\label{fig:res_resnets_drsm_features_corr-2}
		\includegraphics[height=6.4cm,trim={5.4cm 0 8.8cm 
		0},clip]{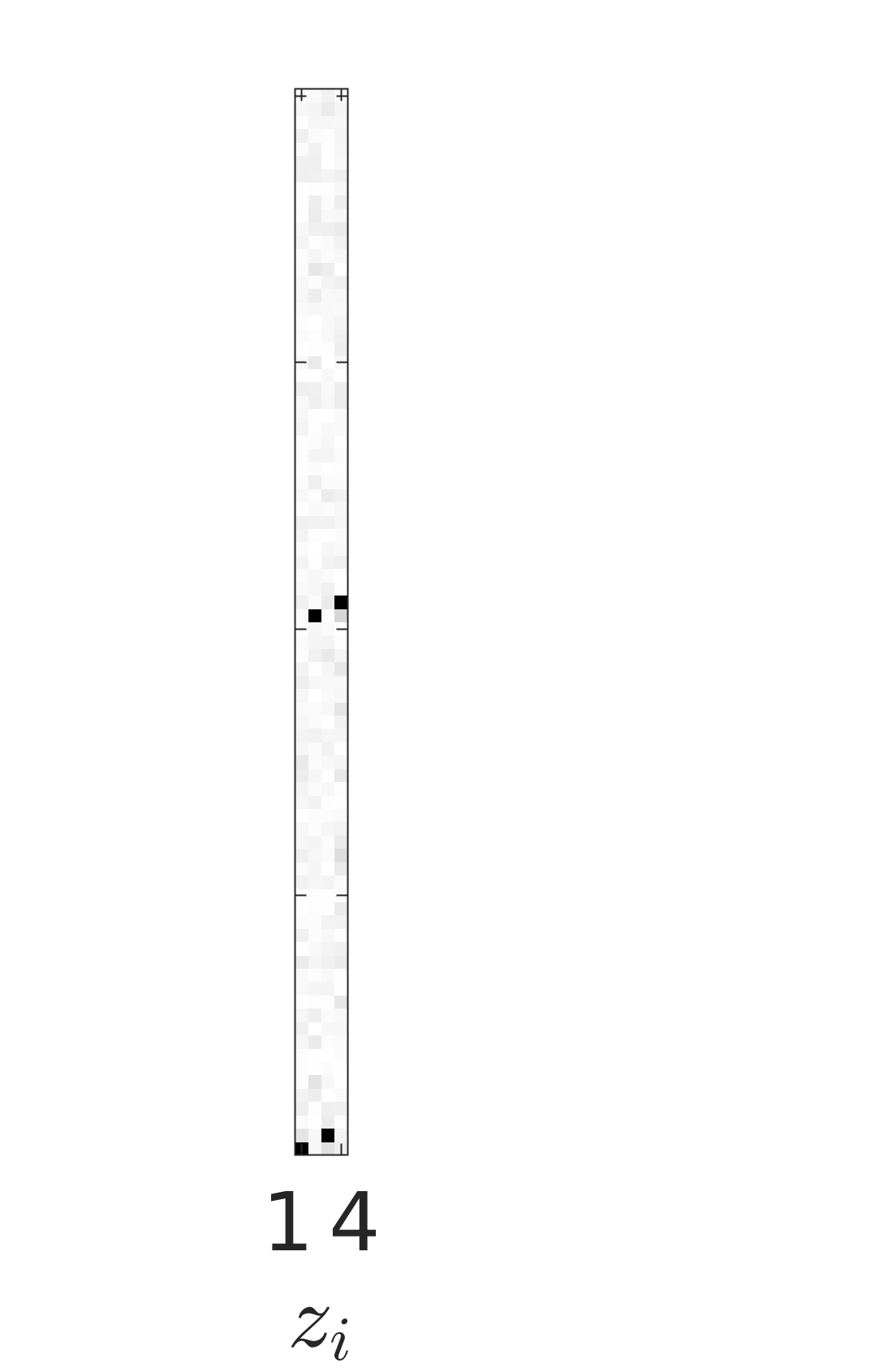}
	}
	\subfigure[$m=8$]{\label{fig:res_resnets_drsm_features_corr-3}
		\includegraphics[height=6.4cm,trim={5.4cm 0 8.5cm 
		0},clip]{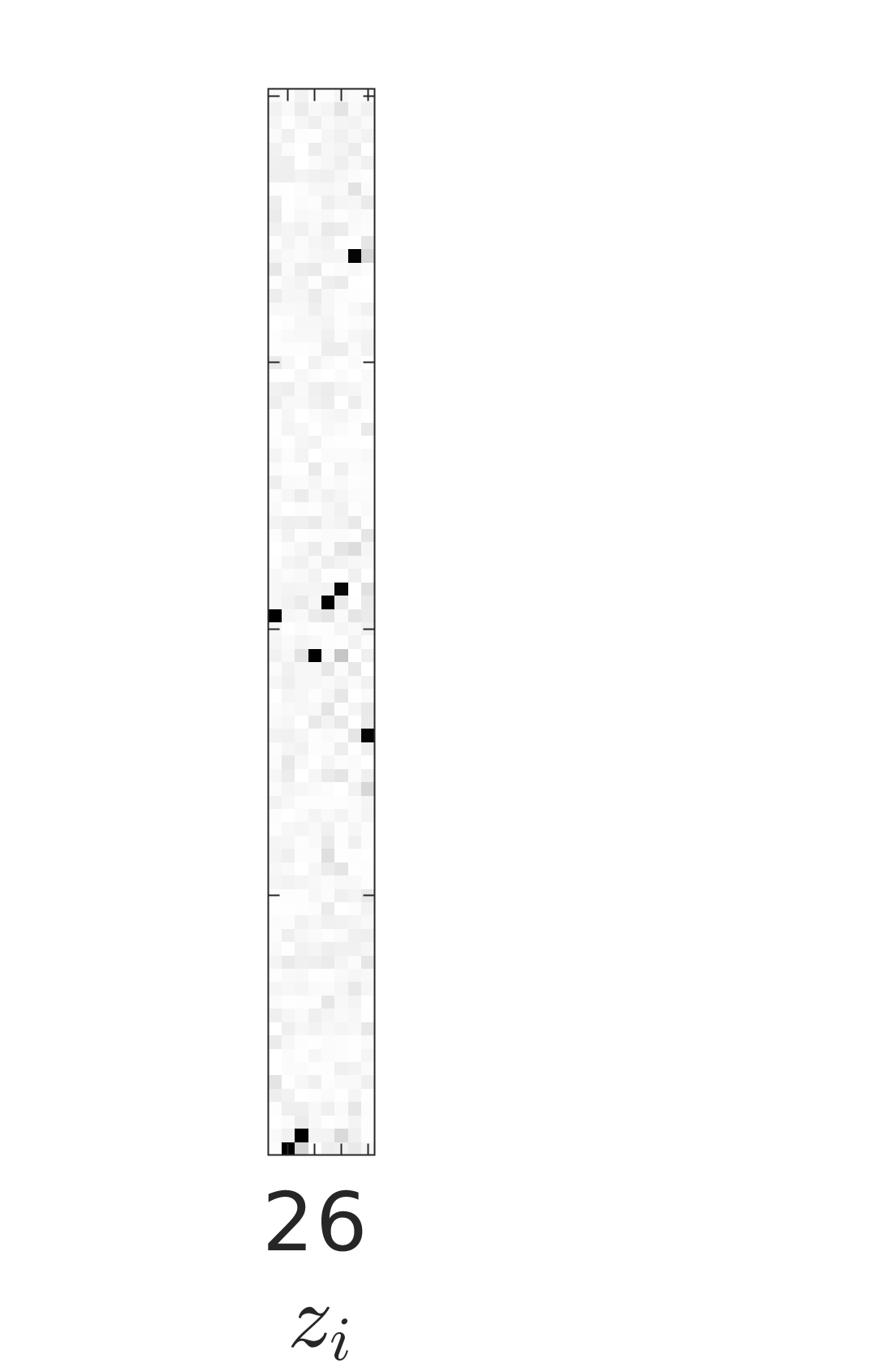}
	}
	\subfigure[$m=16$]{\label{fig:res_resnets_drsm_features_corr-4}
		\includegraphics[height=6.4cm,trim={1.2cm 0 6.0cm 
		0},clip]{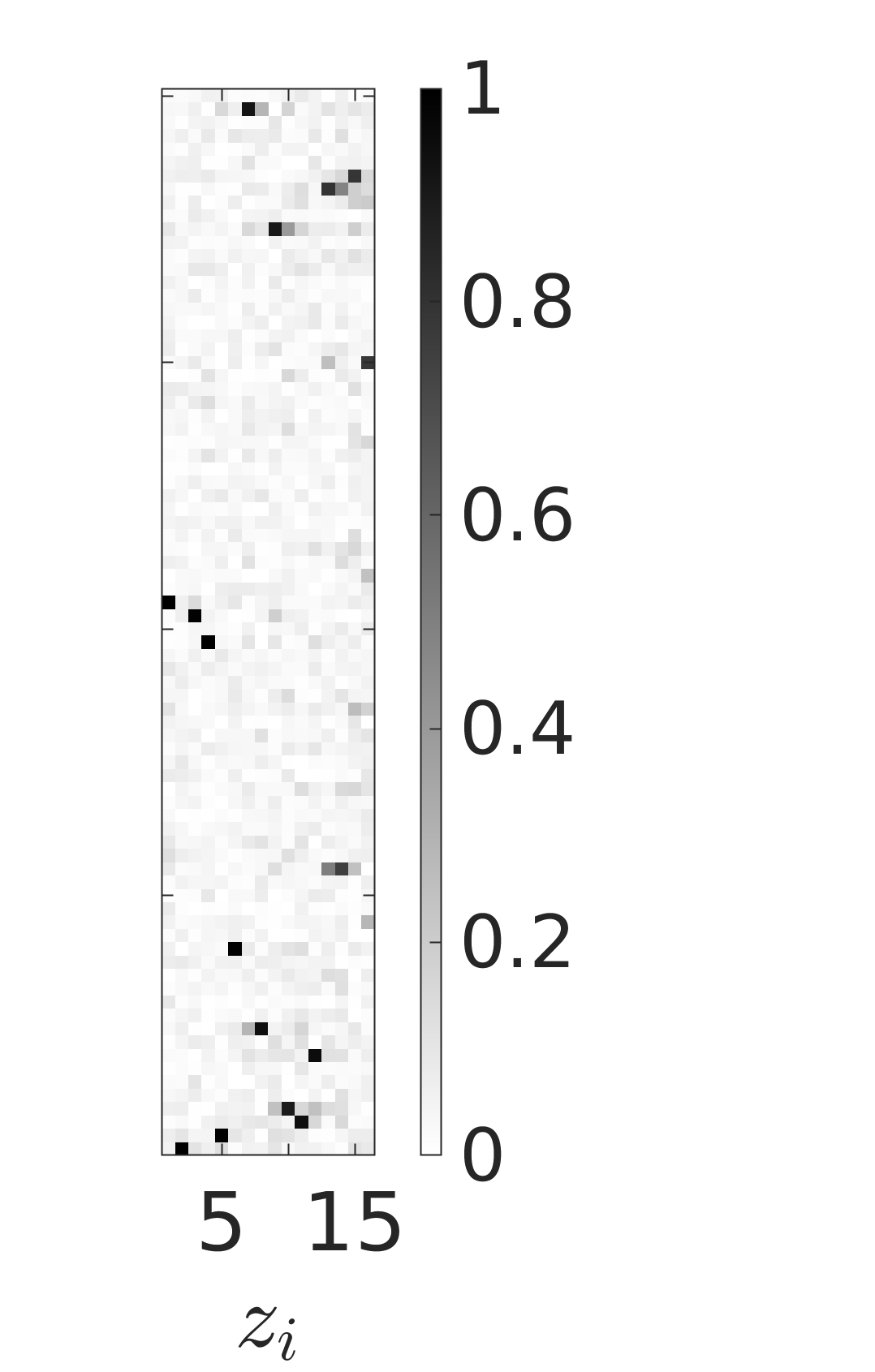}
	}

	%	\caption{Electrical resistor networks: visualisation of the 
	%sample-based Spearman correlation coefficient (absolute value) between the 
	%model inputs $\bfX$ and the reduced space inputs $\bfZ$. }
	\caption{Electrical resistor networks: total Sobol' indices (in logarithmic 
	scale) and visualisation of the sample-based Spearman correlation 
	coefficient (absolute value) between the model inputs $\bfX$ and the 
	reduced space inputs $\bfZ$ for varying reduced space dimension. }
	\label{fig:res_resnets_drsm_features_corr}
\end{figure}

Finally, in \figref{fig:res_resnets_drsm_features_corr} we investigate how the 
reduced variables $\bfZ$ correlate with the input variables $\bfX$.
In contrast to the previous example, no computational model was available for 
this application, but only a relatively large pre-calculated dataset.
To estimate the importance of each input variable in the physical space, we 
therefore chose to use a sensitivity measure called standardised regression 
coefficients (SRC,  \citet{Helton1985}, \citet{UQdoc_12_106}). 
Standard regression coefficients correspond to the variance decomposition 
(Sobol' indices) of a global linear model directly fitted on the available data:
\begin{equation}
\cm\prt{\bfX} \approx \beta_0 + \sum_{i=1}^M\beta_i X_i \, ,
\end{equation}
where the coefficients $\bm{\beta} = \acc{\beta_0 \enu \beta_M}$ are estimated 
by ordinary least-squares directly on the experimental design: 
\begin{equation}
\widehat{\bm{\beta}} = \prt{\cx^\top \cx}^{-1} \cx^\top \bfy \, .
\end{equation}

The SRC indices are then defined as:
\begin{equation}
SRC_i = \frac{\widehat{\bm{\beta}}\,\widehat{\sigma}_{X_i}}{\widehat{\sigma}_Y} 
~,~ i = 1 \enu M \,, 
\end{equation}
where $\widehat{\sigma}_i$ denotes the sample-based standard deviation of $X_i$ 
and ${\widehat{\sigma}_Y}$ the sample-based standard deviation of $Y$. 

Their computational efficiency and simple interpretation (especially for 
linear- or quasi-linear models) are attractive, and little or no information 
about the input distributions is available.
Because $\sum_{i=1}^{M}SRC_i^2 \approx 1$, in 
\figref{fig:res_resnets_drsm_features_src} we plot the $SRC_i^2$.
Indeed variables $X_1$, $X_2$, $X_{41}$ and $X_{42}$ are expected to have the 
most significant contribution to the output variance because the effect of each 
resistor decays as the distance from $V$ increases.  
Similarly to case of the Sobol' function, Figures 
\ref{fig:res_resnets_drsm_features_corr-1} - 
\ref{fig:res_resnets_drsm_features_corr-4} represent graphically the Spearman's 
correlation coefficient $|\rho\prt{z_i,x_i}|$, each panel corresponding to a 
reduced space of different dimension.
In each case, we used the DRSM algorithm to calculate the parameters of the 
KPCA (anisotropic Gaussian) kernel coupled with a PCE surrogate, because this 
setup performed best in \tabref{tab:res_resnets_drsm_mstar}.
Interestingly we observe that as the dimensionality of the reduced space, $m$, 
increases, the most important variables $X_i$ are always captured by at least 
one $Z_i$, while the additional $Z_i$'s either correlate with the next more 
important input variable or correlate with multiple $X_i$'s when $m=8$ or 
larger.

When including the PCE prediction error results in  
Figures~\ref{fig:res_resnets_drsm_m_vs_rmse-pce-loo} and 
\ref{fig:res_resnets_drsm_m_vs_rmse-pce-rmse} in the picture, it is clear that 
the steep decline in the PCE generalisation error for $m<4$ is due to the 
importance of each of the four $X_i$'s with $i \in \acc{1,2,41,42}$. 
For $4<m\leq 32$ the PCE accuracy still improves but at a reduced rate, because 
the additional $Z_i$'s still introduce further meaningful information about $Y$.
%Although the rest of the $X_i$'s with $i \notin \acc{1,2,41,42}$ are less 
%important in terms of their SRC indices, they still affect the value of the 
%model response. 

\subsection{2-dimensional heat diffusion} \label{sec:App:2d_diffusion}

\begin{figure}[!t]
	\centering
	\subfigure[Finite element 
	mesh]{\label{fig:res_diffusion2d_input_output-mesh}
		\includegraphics[width=.30\textwidth,trim={0.5cm 0.5cm 0.5cm 
		1.2cm},clip]{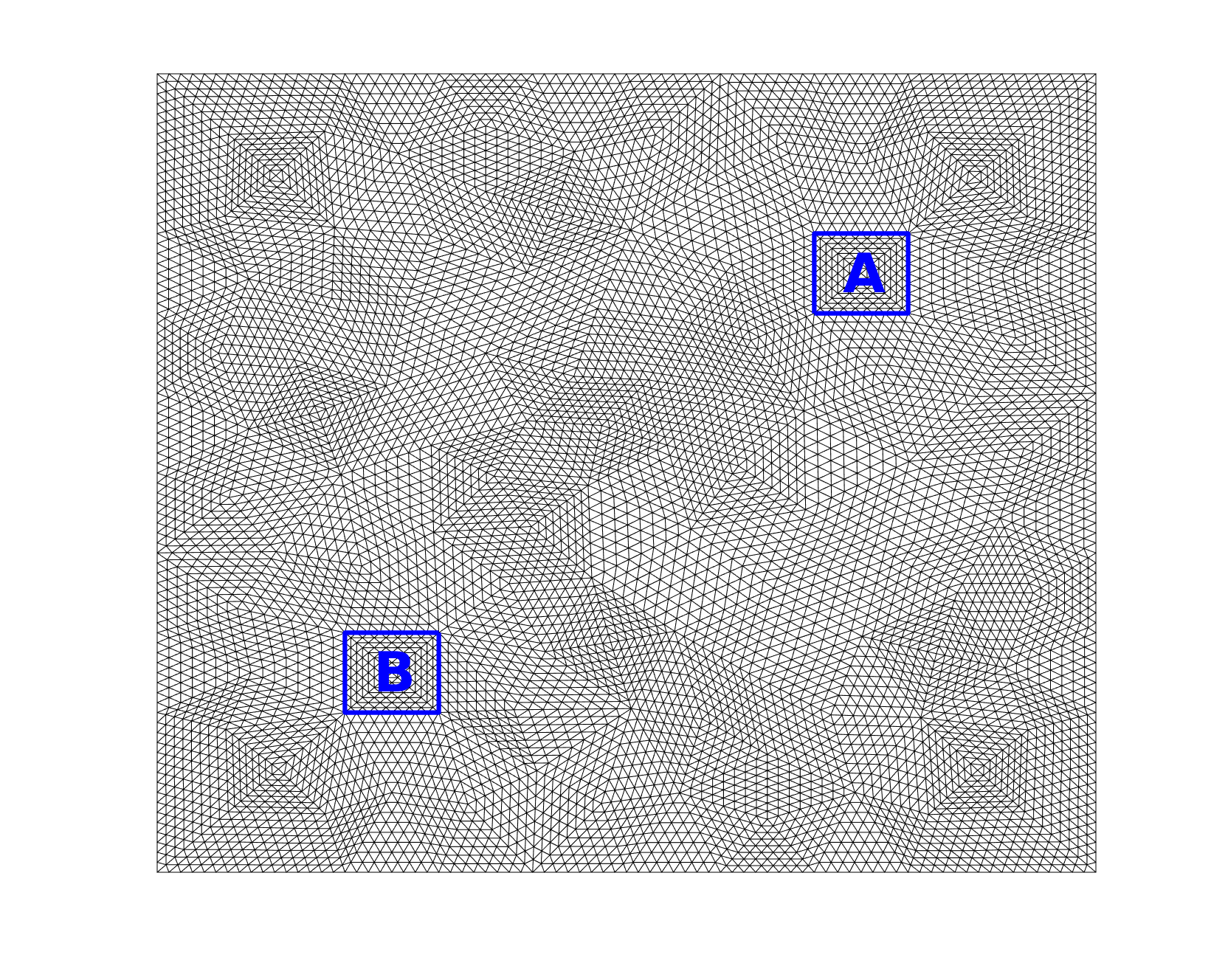}
	}
	\subfigure[Input random field 
	realisation]{\label{fig:res_diffusion2d_input_output-input}
		\includegraphics[width=.30\textwidth,trim={0.5cm 0.5cm 0.5cm 
		1.2cm},clip]{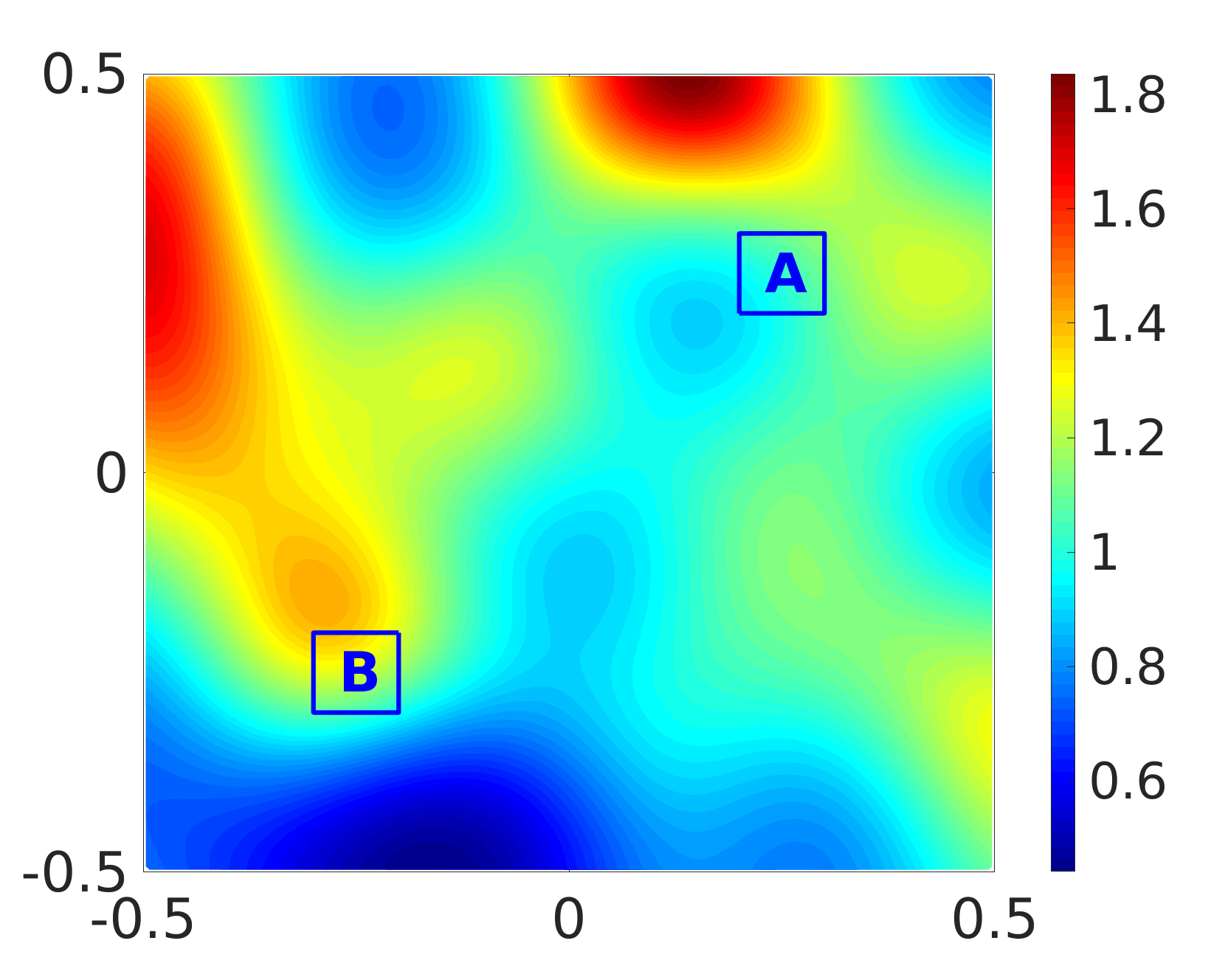}
	}
	\subfigure[Corresponding temperature 
	distribution]{\label{fig:res_diffusion2d_input_output-output}
		\includegraphics[width=.30\textwidth,trim={0.5cm 0.5cm 0.5cm 
		1.2cm},clip]{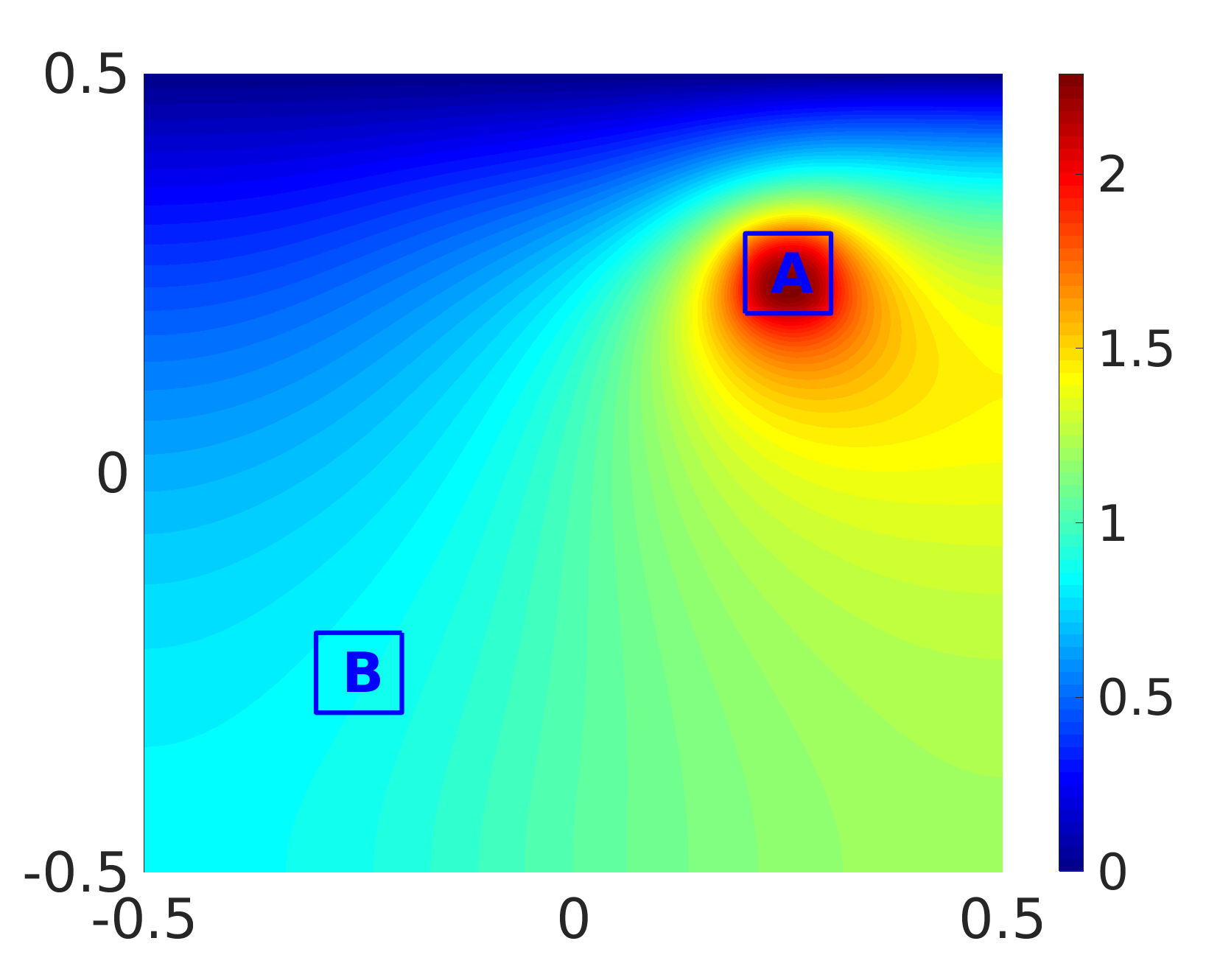}
	}
	\caption{2D heat diffusion problem: illustration of the model input and 
	output. }
	\label{fig:res_diffusion2d_input_output}
\end{figure}

As a last application, we consider a 2-dimensional stationary heat diffusion 
problem. 
The problem is defined in a square domain, $D=[-0.5,0.5]\times[-0.5,0.5]$, 
where the temperature field $T(\bfv),\, \bfv \in D$ is the solution of the 
elliptic partial differential equation:
\begin{equation}
\label{eq:heat_pde}
- \nabla \cdot
\prt{d(\bfv) 
	\nabla{} 
	T(\bfv)} = 500 \, 
I_A(\bfv),
\end{equation}
with boundary conditions $T=0$ on the top boundary and $\nabla T\cdot \bm{n}=0$ 
on the left, right and bottom boundaries, where $\bm{n}$ denotes the vector 
normal to the boundary. In \eqref{eq:heat_pde}, $A$ corresponds to a square 
domain (see \figref{fig:res_diffusion2d_input_output}) and $I_A$ is the 
indicator function equal to 1 if $\bfv \in A$ and $0$ otherwise. The high 
dimensional input is given by the uncertain diffusion coefficient $d(\bfv)$.
To generate samples of the diffusion coefficient, we choose to model it as a 
lognormal random field, defined by:
\begin{equation}
\label{eq:heat_diffcoeff}
d(\bfv) = \exp \prt{a_d + b_d \,g(\bfv) },
\end{equation}
where $g(\bfv)$ is a Gaussian random field and the parameters $a_d$, $b_d$ are 
such that the mean and standard deviation of $d$ are $\mu_d = 1$ and $\sigma_d 
= 0.3$ respectively. The random field is characterised by a Gaussian 
correlation function $R(\bfv,\bfv') = \exp \prt{- \norm{\bfv - 
\bfv'}^2/\ell^2}$, with $\ell=0.2$. The output of interest is the average 
temperature in the square domain $B$ within $D$ (see 
\figref{fig:res_diffusion2d_input_output}). 

% Consider a grid in $D$ with nodes $\acc{\bm{v}_1 \enu \bm{v}_n}$. By 
%retaining the first $p$ terms in the EOLE series, $g(\bfv)$ is approximated by:
% \begin{equation}
%     \label{eq:heat_EOLE}
%     \widehat{g}(\bfv) = \sum_{i=1}^{p} 
%\frac{\xi_i}{\sqrt{l^{(i)}}}\prt{\bm{\phi}^{(i)}}^\top\mat{C}_{\bfv\bm{v}}(\bfv),
% \end{equation}

% where $\acc{\xi_1 \enu \xi_p}$ are independent standard normal random 
%variables, $\mat{C}_{\bfv\bm{v}}$ is a vector with elements 
%$C_{\bfv\bm{v}}^{(k)}=R(\bfv,\bm{v}_k)$ for $k=1 \enu n$ and $\{ \prt{l^{(i)}, 
%\allowbreak \bm{\phi}^{(i)}}, \allowbreak \, i=1 \enu n\}$ are the eigenvalues 
%and eigenvectors of the correlation matrix $\mat{C}_{\bm{v}\bm{v}}$ with 
%elements $C_{\bm{v}\bm{v}}^{(i,j)} = R(\bm{v}_i,\bm{v}_j)$ for $i,j= 1 \enu 
%n$. 
%In the following analysis the Gaussian random field realisations are computed 
%using $p=30$ terms in the EOLE series in \eqref{eq:heat_EOLE}, which allows to 
%represent $93.69\%$ of the variance of the original field. 

The underlying deterministic problem is solved with an in-house finite-element 
analysis code developed in \matlab{}. 
The mesh shown in \figref{fig:res_diffusion2d_input_output-mesh} consists of 
$16,000$ triangular T3 elements.
\figref{fig:res_diffusion2d_input_output-input} shows a realisation of the 
diffusion coefficient random field which corresponds to the input of the model. 
The corresponding model output, shown in  
\figref{fig:res_diffusion2d_input_output-output}, is the mean temperature in 
the highlighted square region $B$. 
Each realisation of the diffusion coefficient random field is discretised over 
the mesh in \figref{fig:res_diffusion2d_input_output-mesh}. 
In the following analysis, the system is treated as a black-box, with the 
discretised heat diffusion coefficient as a high-dimensional input ($M=16,000$) 
and the average temperature in square B as the scalar model output. 
Realizations of the Gaussian random field $g(\bfv)$ are obtained on the 
modelling mesh with the expansion optimal linear estimation (EOLE) method 
\citep{DerKiureghian1993}, by retaining $p = 30$ modes in the numerical 
expansion.
A single set of $500$ experimental design samples and model responses is 
generated and made available for the analysis.
This example mimics a realistic scenario in which various maps of spatially 
varying parameters measured on a regular grid are input to a computational 
model that analyses some performance of the system. 

\begin{figure}
	\centering
	\subfigure[Kriging - LOO error]{\label{fig:res_diffusion2d_m_vs_rmse-KG-LOO}
		\includegraphics[width=.47\textwidth,trim={0.2cm 0 2.0cm 
		0},clip]{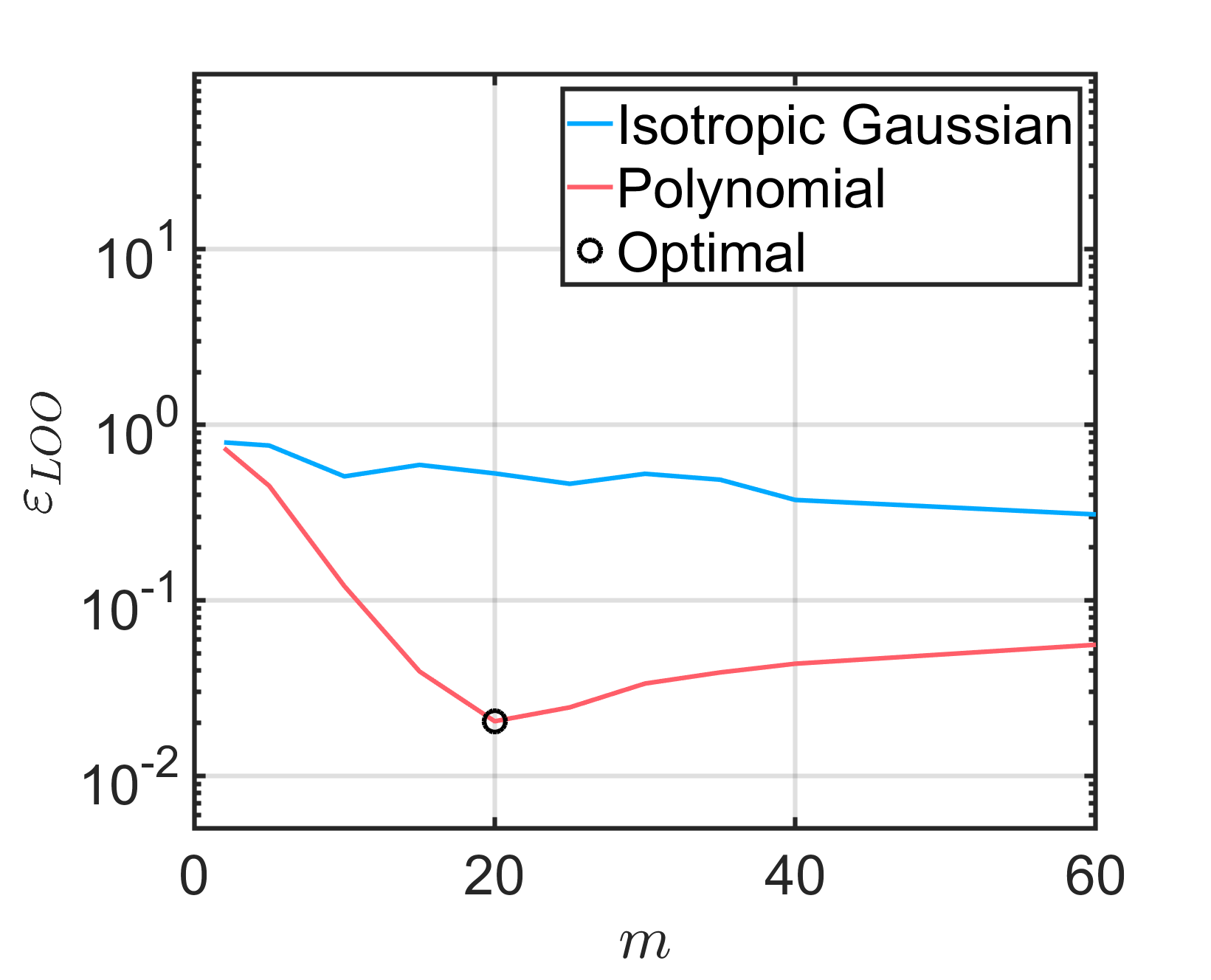}
	}
	\subfigure[PCE - LOO error]{\label{fig:res_diffusion2d_m_vs_rmse-PCE-LOO}
		\includegraphics[width=.47\textwidth,trim={0.2cm 0 2.0cm 
		0},clip]{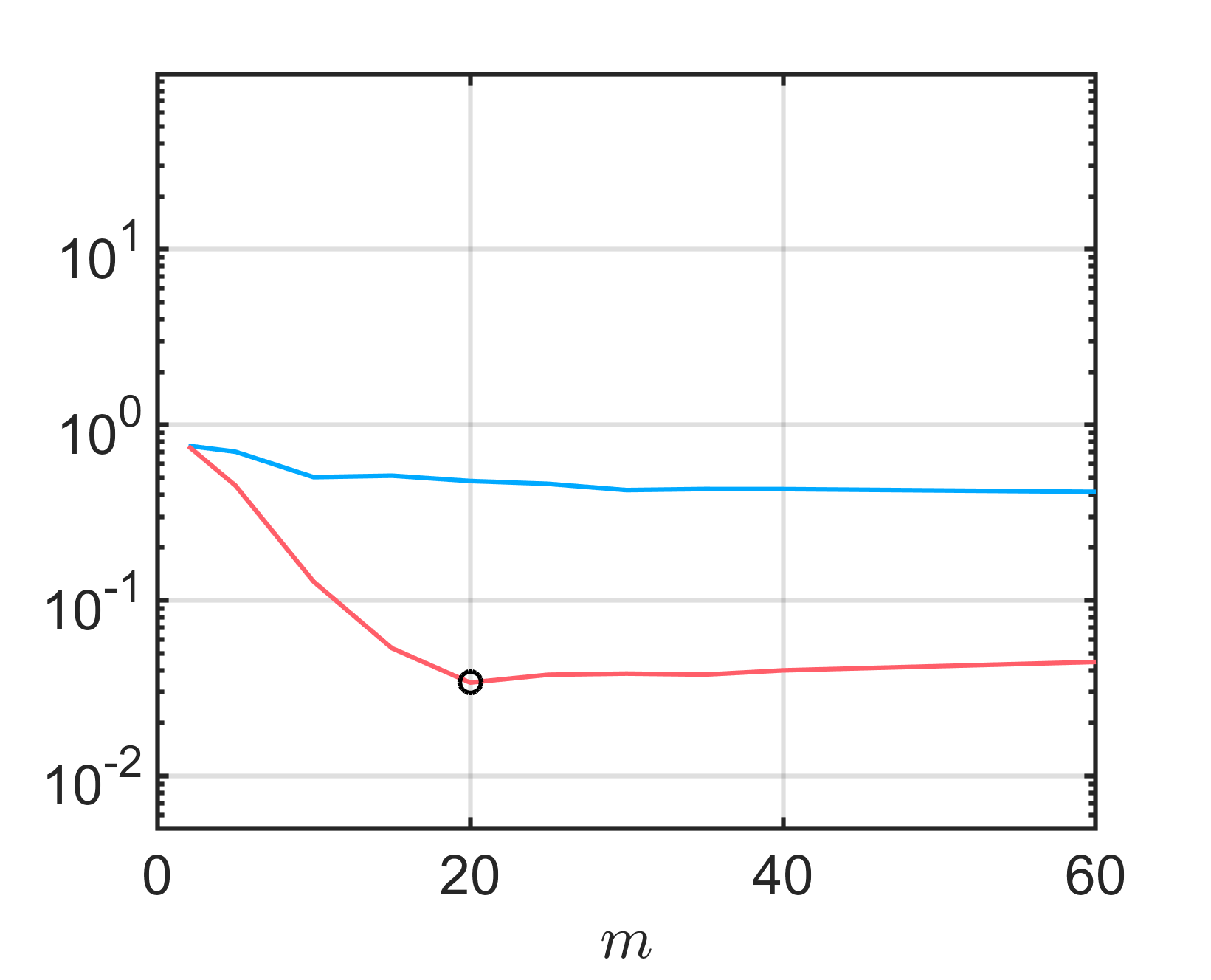}
	}
	
	\subfigure[Kriging - Validation 
	error]{\label{fig:res_diffusion2d_m_vs_rmse-KG-RMSE}
		\includegraphics[width=.47\textwidth,trim={0.2cm 0 2.0cm 
		0},clip]{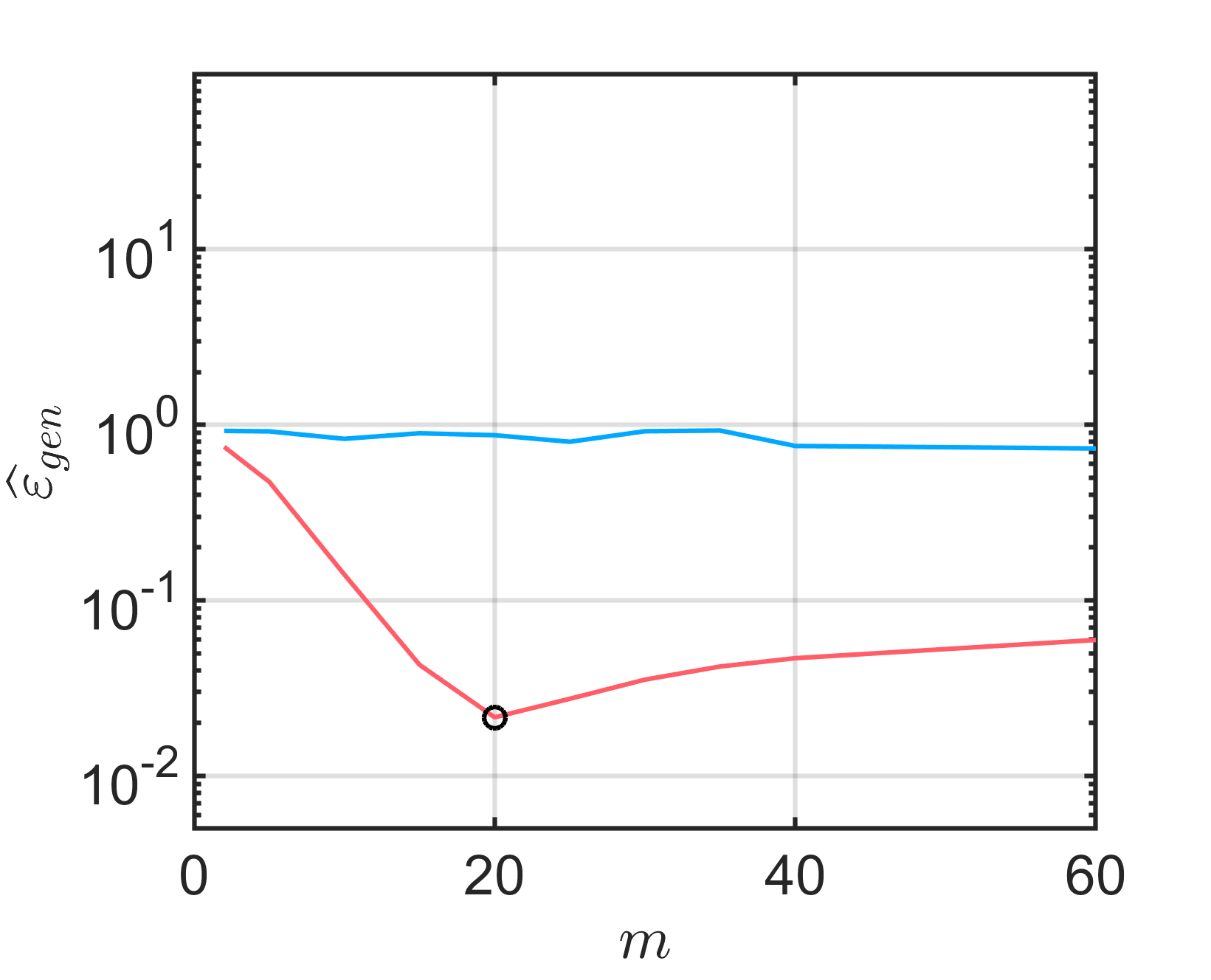}
	}
	\subfigure[PCE - Validation 
	error]{\label{fig:res_diffusion2d_m_vs_rmse-PCE-RMSE}
		\includegraphics[width=.47\textwidth,trim={0.2cm 0 2.0cm 
		0},clip]{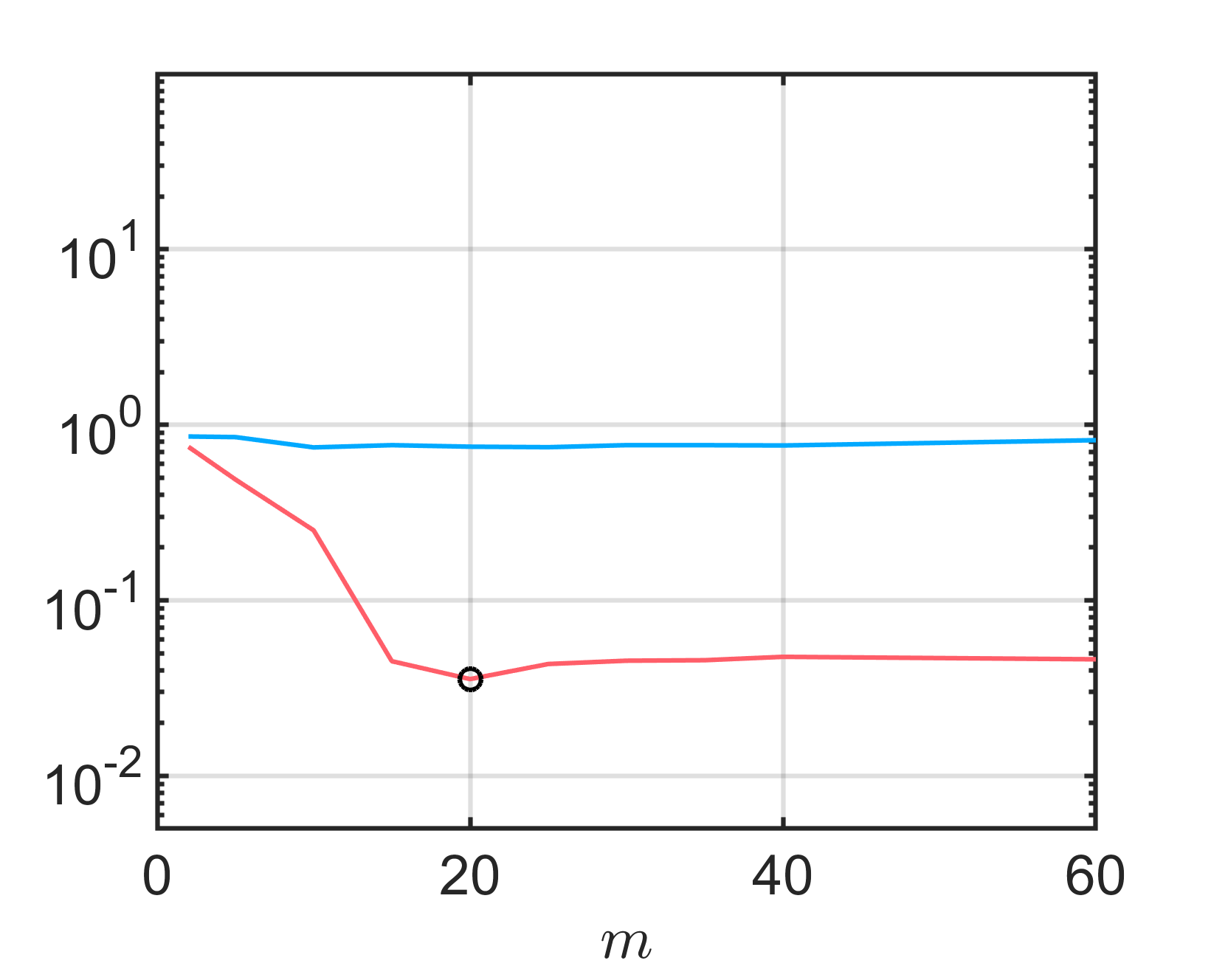}
	}
	\caption{2D heat diffusion problem: Error estimates of the DRSM surrogate 
	as a function of the reduced space dimension. Kernel PCA is used with 
	isotropic Gaussian and polynomial kernels.}
	\label{fig:res_diffusion2d_m_vs_rmse}
\end{figure} 

\begin{table}[!ht]
	\centering
	
	\caption{2D diffusion: optimal DRSM configurations for Kriging- and 
	PCE-based surrogate models}
	\label{tbl:res_heat_drsm_mstar}
	
	\begin{tabular}{l c c c c c c r}
		\hline
		SM method & 
		KPCA kernel & 
		$\widehat{m}$ & 
		\multicolumn{3}{c}{$\bfwh$ (\eqref{eq:KPCA_kernel_poly})} & 
		$\varepsilon_{LOO}$ & 
		$\widehat{\varepsilon}_{gen}$ \\ 
		
		& 
		& 
		& 
		$\widehat{w}_1$ & 
		$\widehat{w}_2$ &
		$\widehat{w}_3$ &
		& 
		\\ 
		\hline
		
		Kriging &
		Polynomial &
		$20$ &
		$131.3681$ &       
		$112.0040$ &
		$1$ &
		$0.0205$ &
		$0.0216$ \\
		
		PCE &
		Polynomial &
		$20$ &
		$17.5225$ &        
		$15.1853$ &
		$1$ &
		$0.0340$ &
		$0.0356$ \\
		\hline
	\end{tabular}
	
\end{table}

As in the previous application examples, the goal of the first analysis is to 
determine the optimal DRSM configuration in terms of the KPCA kernel and the 
reduced space dimension, as well as test the effectiveness of the LOO error as 
a proxy of the validation error. 
In this analysis, the available samples are randomly split into $300$ pairs to 
be used during the DRSM optimisation and $200$  pairs to be used for 
validation. The results are shown in \figref{fig:res_diffusion2d_m_vs_rmse}. 
Figures~\ref{fig:res_diffusion2d_m_vs_rmse-KG-LOO} and 
\ref{fig:res_diffusion2d_m_vs_rmse-PCE-LOO} show the LOO error estimator of the 
final Kriging (resp. PCE) surrogate, evaluated on $\acc{\cx, \bfy}$, whereas 
Figures~\ref{fig:res_diffusion2d_m_vs_rmse-KG-RMSE} and 
\ref{fig:res_diffusion2d_m_vs_rmse-PCE-RMSE} show the validation error of the 
surrogate evaluated on $\acc{\cx_v, \bfy_v}$.
Each curve corresponds to a specific type of KPCA kernel, namely isotropic 
Gaussian and polynomial, and a specific surrogate, namely Kriging and PCE. 
We omitted the anisotropic Gaussian kernel for KPCA which is intractable due to 
the large input dimensionality.

A similar convergence behaviour is observed between Kriging- and PCE- based 
DRSM. 
The corresponding optimal parameter values are highlighted in 
\figref{fig:res_diffusion2d_m_vs_rmse} and their numerical values are reported 
in \tabref{tbl:res_heat_drsm_mstar}.
The linear polynomial kernel performs best in both cases and leads to the same 
reduced space dimension  
$\widehat{m}=20$. 
The convergence to a linear kernel is not surprising, as it corresponds to a 
re-scaled PCA (see \ref{sec:Appendix:PCA_vs_linKPCA}), which is in 
turn closely related to its KL expansion. 
This significantly low dimension has a twofold explanation: although $16,000$- 
dimensional, the heat diffusion 
coefficient is a non-linear combination of $p = 30$ independent standard normal 
random variables (the modes retained 
in the EOLE expansion during the sampling procedure). Additionally, the forward 
operator in \eqref{eq:heat_pde} 
has a smoothening effect that is captured by DRSM, further reducing the 
effective dimensionality of the problem.
Moreover, the LOO and validation error curves show similar behaviour both in 
terms of their trend and their absolute 
value. Hence, the LOO error served as a reliable proxy of the validation error, 
as was observed in the previous 
application examples too. 

\begin{figure}
	\centering
	\subfigure[Kriging]{\label{fig:res_diffusion2d_drsm_vs_others-KG}
		\includegraphics[width=.47\textwidth,trim={0.2cm 0 2.0cm 
		0},clip]{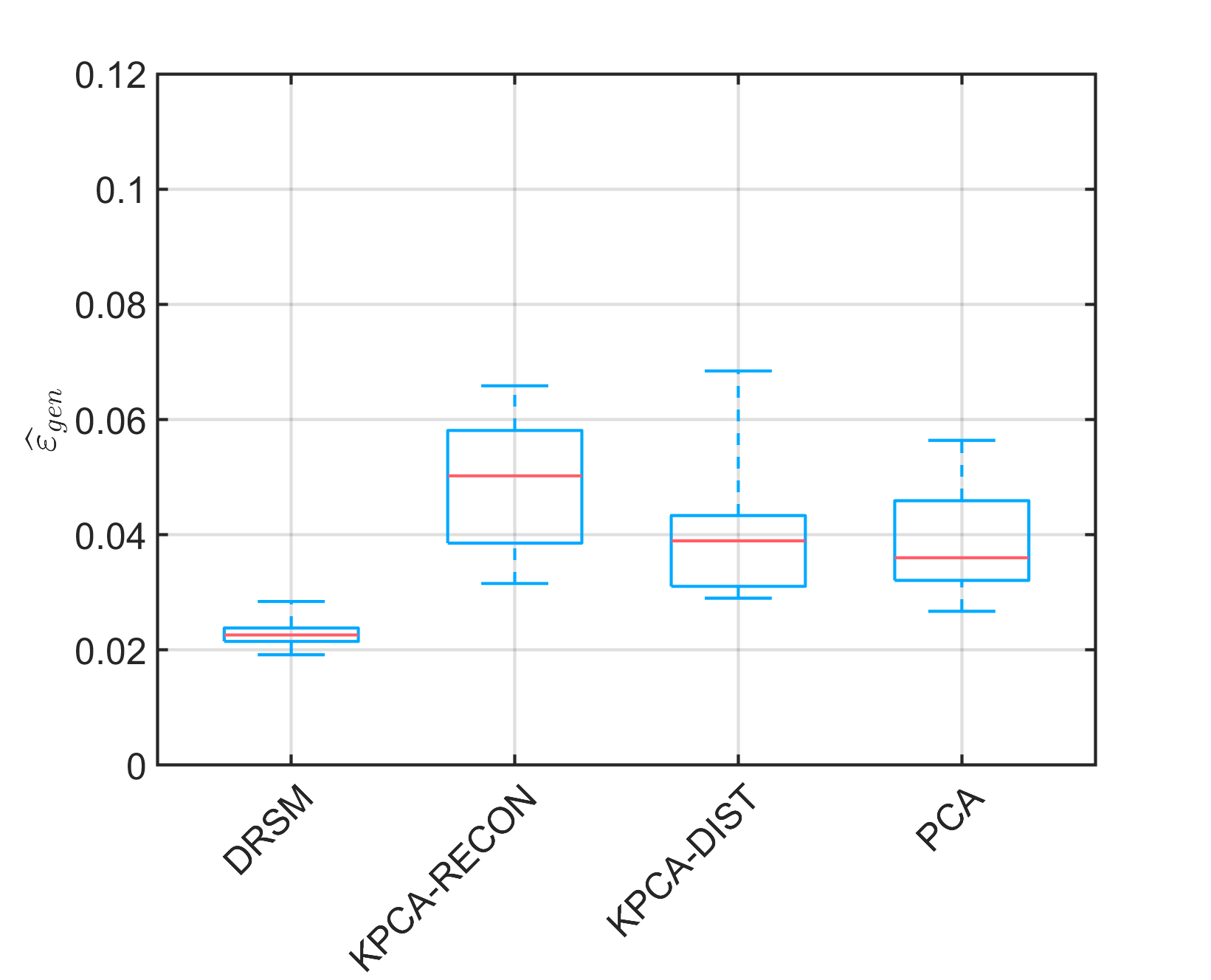}
	}
	\subfigure[Polynomial chaos 
	expansions]{\label{fig:res_diffusion2d_drsm_vs_others-PCE}
		\includegraphics[width=.47\textwidth,trim={0.2cm 0 2.0cm 
		0},clip]{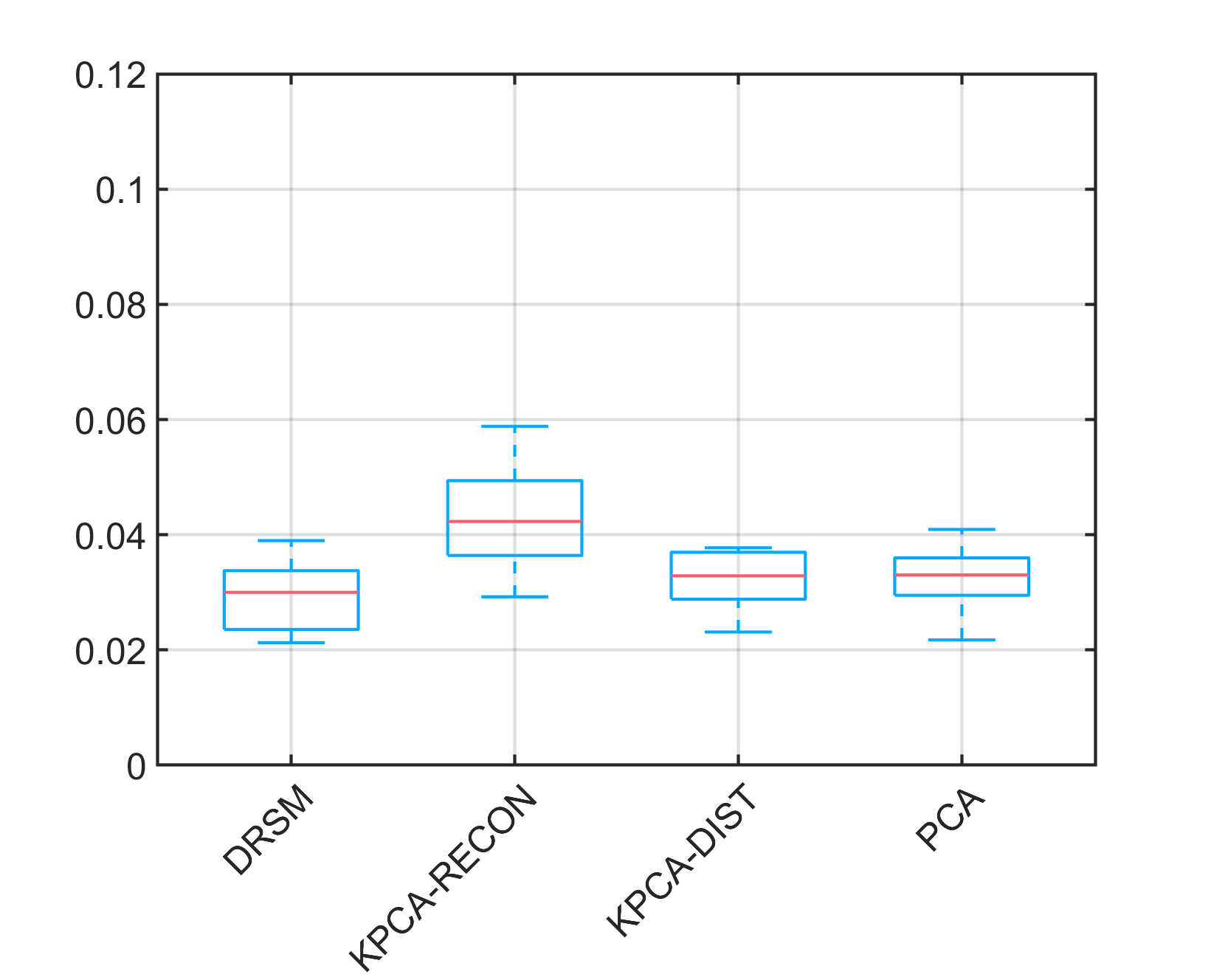}
	}
	
	\caption{2D heat diffusion problem: estimates of the generalisation error 
		with different DR approaches (see \tabref{tbl:validation_configs} for 	
		their description). The optimal reduced dimension identified by DRSM 
		(see 
		\tabref{tbl:res_heat_drsm_mstar}) is used in all cases.}
	\label{fig:res_diffusion2d_drsm_vs_others}
\end{figure} 

In the subsequent analysis we compare the performance of the DRSM approach 
against other sequential approaches listed in \tabref{tbl:validation_configs}. 
To test each setup, we repeat the calculation process $10$ times. 
In each case the $500$ available samples are split randomly into $300$ samples 
for calculating the surrogate and $200$ samples for validation. 
The optimal KPCA kernel that was determined by DRSM is used in all methods that 
involve KPCA. 
Also, for the sake of comparison, the same reduced space dimension 
$\widehat{m}=20$ is assumed for all methods.

The results of this comparative study are given in 
\figref{fig:res_diffusion2d_drsm_vs_others} using box plots to provide summary 
statistics of the validation error over the different splits of the samples.
In case of Kriging surrogate modelling, DRSM consistently provides  superior 
results compared to the other methods. 
%Notice that KPCA with linear kernel is equivalent to PCA on a scaled version 
%of the experimental design with scaling factor $\sqrt{w_1}$ (see 
%\ref{sec:Appendix:PCA_vs_linKPCA} for more details). 
The Kriging surrogates, in contrast to the PCE ones, are affected by this 
scaling. 
This also explains the performance improvement compared to the case of 
PCA-based DR. 
In case of PCE surrogate modelling, the performance improvement gained by DRSM 
is marginal compared to PCA and KPCA with distance preservation- based tuning 
of $\bfw$.

Overall, DRSM consistently provides more accurate or at least comparable 
results compared to the other approaches.  
The main difference with a standard UQ setting in which the thermal 
conductivity is supposed to be sampled from a random field with known 
properties, is that the proposed DRSM methodology is purely data-driven, \ie{} 
it would be applied identically in a case when the input maps are given without 
knowing the underlying random process.

% #######################	
\section{Summary and Conclusions}
% #######################

Surrogate modelling is a key ingredient of modern uncertainty quantification. 
Due to the detrimental effects of high input dimensionality on most recent 
surrogate modelling techniques, the input space needs to be compressed to make 
such problems tractable.
We proposed a novel approach for effectively combining dimensionality reduction 
with surrogate modelling, called DRSM. 
DRSM consists of three steps: (i) the DR and SM parameters are calculated by 
solving a nested optimisation problem, where only low-accuracy surrogates are 
considered to reduce the associated computational cost, (ii) the optimal 
configuration parameters, including the dimension of the reduced space, are 
empirically estimated based on the surrogate model performance, and, (iii) a 
final high-accuracy surrogate is calculated using the optimal values of all the 
aforementioned parameters.

The performance of DRSM was compared on three different benchmark problems of 
varying complexity against the classical approach of tuning the dimensionality 
reduction and surrogate modelling parameters sequentially. 
DRSM consistently showed superior performance compared to the others in all the 
benchmark applications. 

The novelty of the proposed methodology lies in its non-intrusive way of 
combining dimensionality reduction and surrogate modelling. This allows for the 
combination of various techniques without the need of altering the dedicated 
optimisation algorithms on which each of them capitalises.
A practical implication of the non-intrusiveness of DRSM is that off-the-shelf 
surrogate modelling methods (or even software) with sophisticated calibration 
algorithms can be directly used within this framework.

The focus was given to data-driven scenarios where only a limited set of 
observations and model responses is available. We demonstrated that the 
leave-one-out cross-validation error of the surrogate models can serve as a 
reliable proxy for estimating the generalisation error in order to  tune the DR 
parameters, but also to assess the overall accuracy of the resulting surrogate.

In application-driven scenarios where the goal is to obtain a surrogate with 
optimal performance (regardless of its type) for that specific problem, the 
proposed approach could be extended in a way that the surrogate type itself is 
included as one of the parameters that DRSM needs to optimise. 
However, special care would need to be given to the error metric used during 
the DRSM optimisation in this case, because the LOO error estimations by 
different surrogates may have widely varied levels of bias (see \eg{} 
\citet{Tibshirani2009}). 

In future extensions of this work, focus will be given to capitalising on 
available HPC resources to optimise for different combinations of surrogate 
models and dimensionality reduction methods. 
The choice of optimal low-accuracy surrogates needs also further study, as it 
may sometimes lead to relatively poor performance (e.g. as in the case of 
Kriging in Section \ref{sec:App:sobol}).
In addition, the cost of training surrogate models increases with the number of 
available experimental design samples.
Therefore, research efforts will also be directed towards dealing with large 
experimental designs, possibly within a \textit{big data} framework.

% #######################	
\section*{Acknowledgements}
% #######################

Dr John Jakeman (Sandia National Laboratories) is gratefully acknowledged for 
providing the data sets used in the electrical resistor networks application 
example (\secref{sec:App:resistor_networks}).

% #######################
% References
% #######################
%\section*{References}

\bibliography{IJUQ_31935_Bib}
\clearpage

% #######################	
\appendix
% #######################	

% #######################	
\section{Relationship between PCA and KPCA with linear kernel} 
\label{sec:Appendix:PCA_vs_linKPCA} 
% #######################	

Consider the PCA-based dimensionality reduction $\bfx \in \Rr^M \mapsto \bfz 
\in \Rr^m$. As discussed in \secref{sec:Meth:DR:PCA}, $\bfz$ is calculated as 
follows:

\begin{equation} \label{eq:App_pca_z}
\bfz = \bfx^\top \bm{V},
\end{equation}

where $\bm{V} \in \Rr^{M\times m}$ is the collection of the $m$ eigenvectors of 
$\bm{C} = \text{cov}\bra{\cx}$ and $\cx \in \Rr^{N\times M}$ is the 
experimental design. 

Next, consider the kernel PCA mapping  $\bfx \in \Rr^M \mapsto \bm{q} \in 
\Rr^m$ using the linear kernel function:

\begin{equation} \label{eq:app_linkernel}
\kappa\prt{\bm{x},\bm{x}^\prime} = a\,  \bm{x}^\top \bm{x}^\prime + b.
\end{equation}
It is straightforward to show that the following transformation is equivalent 
to the linear kernel in \eqref{eq:app_linkernel}:

\begin{equation} \label{eq:app_linkernel_Phi}
\Phi(\bfx) = \acc{\sqrt{b},\sqrt{a}\,x_1 \enu \sqrt{a}\,x_M}^\top, 
\end{equation}

because $\kappa\prt{\bm{x},\bm{x}^\prime} = \Phi(\bfx)^\top \Phi(\bfx^\prime)$.
A sample $\bm{q}$ in the reduced space is calculated as follows (see 
\secref{sec:Meth:DR:KPCA}):

\begin{equation} \label{eq:app_kpca_z}
\bm{q} = \Phi(\bfx)^\mathsf{T} \bm{V}_\ch,
\end{equation}

where $\bm{V}_\ch$ is the collection of the $m$ eigenvectors of $\bm{C}_{\ch} = 
\text{cov}\bra{\Phi(\cx)}$ with maximal eigenvalues. Notice that in case of 
$a=1$ and $b=0$, from Eqs.~(\ref{eq:App_pca_z}),~(\ref{eq:app_kpca_z}) follows 
that $\bfz = \bm{q}$. 

The covariance matrix $\bm{C}_{\ch}$ can be expressed as: 

\begin{equation}
\label{eq:app_Ch}
\bm{C}_{\ch} = \begin{bmatrix} 
0 & \ldots & 0 \\
\vdots & \multicolumn{2}{c}{\multirow{2}{*}{$a\, \bm{C}$}} \\
0 
\end{bmatrix}.
\end{equation}

Hence, excluding the eigenvector that corresponds to the zero eigenvalue, it is 
straightforward to show that 

\begin{equation}
\label{eq:app_Vh}
\bm{V}_{\ch} = \begin{bmatrix} 
0 & \ldots & 0 \\
\multicolumn{3}{c}{\bm{V}} \\
\end{bmatrix}.
\end{equation}

Based on Eqs.~(\ref{eq:app_linkernel_Phi}) and (\ref{eq:app_Vh}), 
\eqref{eq:app_kpca_z} can be written as follows:

\begin{alignat}{3}
\bm{q} & =  \begin{bmatrix}\sqrt{b} & \sqrt{a}\,\bfx^\top \end{bmatrix} 
\begin{bmatrix} 
0 & \ldots & 0 \\
\multicolumn{3}{c}{\bm{V}} \\
\end{bmatrix}\\
& =  \sqrt{a} \, \bfx^\top \bm{V} \\
& =  \sqrt{a} \, \bfz \quad \text{(from \eqref{eq:App_pca_z})}
\end{alignat}

Therefore, the dimensionality reduction using kernel PCA with a linear kernel 
provides a scaled version of standard PCA and the constant $b$ has no effect.

% #######################	
\section{Implementation details} \label{sec:Appendix:details}
% #######################	

This section provides an extensive list of the configuration parameter values 
that were used to produce the results in \secref{sec:Applications}. 
\tabref{tbl:DRSM_Kriging_config} (resp. \tabref{tbl:DRSM_PCE_config})lists the 
configuration parameters of Kriging (resp. polynomial chaos expansions) 
surrogate models. 
For each surrogate method a distinction is made, in terms of the parameters 
used, between the proxy (\ie{} low computational cost) surrogate and the 
high-accuracy one.
The proxy surrogates were used for solving the nested optimisation problem of 
DRSM in Eqs.~(\ref{eq:DRSM_outer_loop}),~(\ref{eq:DRSM_inner_loop}).
The same configuration was used to calculate the high-accuracy surrogates 
regardless of the input compression method (DRSM or disjoint PCA/KPCA).

\begin{table}[!ht]
	\centering
	
	\caption{The configuration of the Kriging surrogates that were calculated 
	during the various steps of DRSM for each application example. }
	\label{tbl:DRSM_Kriging_config}
	
	\begin{tabular}{p{0.25\textwidth}| p{0.21\textwidth} p{0.21\textwidth} 
	p{0.21\textwidth}}
		\hline
		% \multicolumn{3}{c}{Kriging-based DRSM configuration for optimal 
		%parameters search}\\
		Application & Sobol' function & Resistor networks & 2D diffusion \\
		\hline
		\multicolumn{4}{p{0.88\textwidth}}{
			1. Proxy surrogate configuration
		}\\\hline
		
		Trend &
		constant ($P=0$)&
		linear ($P=1$)&
		linear ($P=1$)\\
		
		Correlation family&
		\multicolumn{3}{p{0.66\textwidth}}{
			\textit{isotropic} Mat\'ern (\eqref{eq:Kriging_Matern_general}) 
			with $\nu = 5/2$
		}\\
		Estimation method&
		\multicolumn{3}{p{0.66\textwidth}}{
			Cross-validation (\eqref{eq:Kriging_thetaCV})
		}
		\\
		Optim. method &
		\multicolumn{3}{p{0.66\textwidth}}{
			Genetic algorithm (GA) with BFGS (gradient based) refinement of 
			final solution
		}\\
		Optim. constraints &
		\multicolumn{3}{p{0.66\textwidth}}{
			$\bfth \in [0.01, 100]$
		}\\
		Population size (GA) &
		\multicolumn{3}{p{0.66\textwidth}}{
			$10$
		} \\
		Max. iterations: &
		\multicolumn{3}{p{0.66\textwidth}}{
			$20$ for both GA and BFGS 
		}\\
		\hline
		\multicolumn{4}{p{0.88\textwidth}}{
			2. High-accuracy surrogate configuration. Only the parameters that 
			differ from the proxy surrogate configuration are listed 
		}\\\hline
		Correlation family&
		\multicolumn{3}{p{0.66\textwidth}}{
			\textit{anisotropic} Mat\'ern with $\nu = 5/2$
		}\\
		Population size (GA) &
		\multicolumn{3}{p{0.66\textwidth}}{
			$20$
		} \\
		Max. iterations: &
		\multicolumn{3}{p{0.66\textwidth}}{
			$50$ for both GA and BFGS 
		}\\
		\hline
	\end{tabular}
\end{table}

%%%%
\begin{table}[!ht]
	\centering
	\caption{The configuration of the PCE surrogates that were calculated 
	during the various steps of DRSM for each application example }
	\label{tbl:DRSM_PCE_config}
	
	\begin{longtable}{p{0.25\textwidth}| p{0.21\textwidth} p{0.21\textwidth} 
	p{0.21\textwidth}}
		\hline
		Application & Sobol' function & Resistor networks & 2D diffusion \\
		\hline
		\multicolumn{4}{p{0.88\textwidth}}{
			1. Proxy surrogate configuration
		}\\\hline
		
		Coeff. calculation method &
		\multicolumn{3}{p{0.66\textwidth}}{
			Ordinary least squares \citep{Berveiller2006} 
		}\\

		Univariate polynomials family&
		\multicolumn{3}{p{0.66\textwidth}}{
			Legendre
		}\\
		
		Hyperbolic truncation $q$  \citep{BlatmanPEM2010} &
		$0.75$&
		$0.50$&
		$0.65$\\
		
		Polynomial degree (adaptive search range) &
		$[1,10]$ &
		$[1,10]$ &
		$[1,5]$ \\
		
		\hline 
		\multicolumn{4}{p{0.88\textwidth}}{
			2. High-accuracy surrogate configuration. Only the parameters that 
			differ from the proxy surrogate configuration are listed 
		}\\\hline
		Coeff. calculation method &
		\multicolumn{3}{p{0.66\textwidth}}{
			Hybrid least angle regression \citep{BlatmanJCP2011}
		}\\
		Univariate polynomials family&
		\multicolumn{3}{p{0.66\textwidth}}{
			Orthogonal to the  probability density function of the input 
			variables that is estimated by kernel-smoothing, using the 
			Stieltjes procedure \citep{Gautschi2004}
		}\\
		Hyperbolic truncation $q$  \citep{BlatmanPEM2010}&
		\multicolumn{3}{p{0.66\textwidth}}{
			$0.75$
		}\\
		Polynomial degree (adaptive search range) &
		\multicolumn{3}{p{0.66\textwidth}}{
			$[1,15]$
		}\\
		\hline
	\end{longtable}
	
\end{table}

The parameters of the DRSM-based optimisation are listed in 
\tabref{tbl:DRSM_optim_config}. Note that the exact same optimisation algorithm 
and parameters were used for optimising $\bfw$ w.r.t. the KPCA reconstruction 
and point-wise distance error in the box-plots used to compare the various 
approaches. 
The optimisation constraints differ from the ones reported in 
\tabref{tbl:DRSM_optim_config} when a polynomial kernel is used in KPCA, as in 
\eqref{eq:KPCA_kernel_poly}, for improved numerical stability of the solver. 
On top of the bound constraints reported in the table, that still apply for 
$w_1$ and $w_2$, the variable $w_3$ (degree) is constrained to integer values 
$1\leq w_3 \leq 4$ instead. In addition, the following non-linear constraint is 
included:
\begin{equation}
w_1 \bfx^\mathsf{T}\bfx' + w_2 > 1.
\end{equation}

%%%%
\begin{table}[!ht]
	\centering
	\caption{Parameters of the DRSM optimisation algorithm}
	\label{tbl:DRSM_optim_config}
	
	\begin{longtable}{p{0.22\textwidth}| p{0.22\textwidth} p{0.22\textwidth} 
	p{0.22\textwidth}}
		\hline
		Application & Sobol' function & Resistor networks & 2D diffusion \\
		\hline
		
		Optim. method &
		\multicolumn{3}{p{0.66\textwidth}}{
			Genetic algorithm with BFGS (gradient based) refinement of final 
			solution
		}\\
		Optim. constraints &
		\multicolumn{3}{p{0.66\textwidth}}{
			$\bfw \in [0.1,300]$
		}\\
		Population size(GA): &
		$20$ for isotropic KPCA kernels, $80$ for anisotropic  &
		$20$ for isotropic KPCA kernels, $100$ for anisotropic &
		$20$ (only isotropic KPCA kernels were considered)\\
		Max. iterations: &
		$80$ for both GA and BFGS &
		$150$ for both GA and BFGS &
		$80$ for both GA and BFGS \\
		\hline
	\end{longtable}
	\vspace{0.1cm}
	
\end{table}

\end{document}